\pdfoutput=1

\documentclass[11pt]{article}

\usepackage[preprint]{acl}

\usepackage{times}
\usepackage{latexsym}

\usepackage[T1]{fontenc}

\usepackage[utf8]{inputenc}

\usepackage{microtype}

\usepackage{inconsolata}

\usepackage{graphicx}

\usepackage{booktabs}
\usepackage{multirow, multicol}
\usepackage{diagbox}
\usepackage{listings}
\usepackage{listings-ext}
\usepackage{longtable}
\usepackage{adjustbox}
\usepackage{xcolor}
\usepackage{algorithm}
\usepackage{algpseudocode}
\usepackage{ctable}
\usepackage{tabularx}
\usepackage{xspace}
\usepackage{amsmath}
\usepackage{amsthm}
\usepackage{float}
\usepackage{soul}
\usepackage{tablefootnote}
\usepackage{colortbl}
\usepackage{subcaption}
\usepackage{pifont}
\usepackage{amssymb}
\usepackage{enumitem}
\usepackage{listings}
\usepackage{listings-ext}
\usepackage{longtable}

\theoremstyle{definition}
\newtheorem{definition}{Definition}[section]

\definecolor{mediumgreen}{RGB}{60, 179, 113}
\lstdefinelanguage{Jinja2}{
  morekeywords={},
  sensitive=false,
  moredelim=[s][\color{blue}]{\{}{\}},
  moredelim=[s][\color{blue}]{\%}{\%},
  moredelim=[s][\color{mediumgreen}]{\{####}{####\}},
}

\definecolor{codegreen}{rgb}{0,0.6,0}
\definecolor{codegray}{rgb}{0.5,0.5,0.5}
\definecolor{codepurple}{rgb}{0.58,0,0.82}
\definecolor{backcolour}{rgb}{0.95,0.95,0.92}

\definecolor{comment-red}{rgb}{0.8,0,0}
\definecolor{lightgray}{gray}{0.7}

\newcommand{\D}[0]{\mathcal{D}\xspace}
\newcommand{\AEE}[0]{\text{AEE}\xspace}
\newcommand{\ED}[0]{\text{ED}\xspace}
\newcommand{\AEL}[0]{\text{AEL}\xspace}

\newcommand{\AEAE}[0]{\text{AEAE}\xspace}

\newcommand{\dataset}[0]{\textsc{Lemonade}\xspace}
\newcommand{\datasetEmoji}{\raisebox{-0.1em}{\includegraphics[trim=0 0 0 0, clip, height=1.1em]{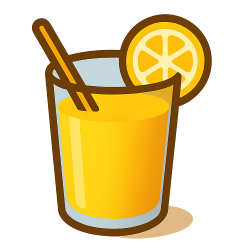}}
}

\newcommand{\datasetWithEmoji}{\texorpdfstring{\datasetEmoji \dataset}{\dataset}}

\newcommand{\system}[0]{\textsc{Zest}\xspace}

\makeatletter
\newcommand{\printfnsymbol}[1]{%
  \textsuperscript{\@fnsymbol{#1}}%
}

\title{\datasetWithEmoji: A Large Multilingual Expert-Annotated \\ Abstractive Event Dataset for the Real World}

\author{
\!Sina J. Semnani\textsuperscript{\textnormal{1}} \quad Pingyue Zhang\textsuperscript{\textnormal{2}}\quad  Wanyue Zhai\textsuperscript{\textnormal{1}}\quad  Haozhuo Li\textsuperscript{\textnormal{1}} 
\\
\textbf{Ryan Beauchamp}\textsuperscript{\textnormal{1}}\quad  \textbf{Trey Billing}\textsuperscript{\textnormal{3}}\quad  \textbf{Katayoun Kishi}\textsuperscript{\textnormal{3}}\quad \textbf{Manling Li}\textsuperscript{\textnormal{2}}\quad 
\textbf{Monica S. Lam}\textsuperscript{\textnormal{1}}
\\
{ \!\!\textsuperscript{1}Stanford University\ \
  \textsuperscript{2}Northwestern University \ \
  \textsuperscript{3}ACLED
  }
  \\
  {\small \texttt{\{sinaj, wzhai702, tommy01, rmb87, lam\}@cs.stanford.edu},} \\
  {\small \texttt{\{pingyue.zhang, manling.li\}@northwestern.edu}, } \\
  {\small \texttt{\{t.billing, k.kishi\}@acleddata.com}}
}

\begin{document}
\maketitle

\begin{abstract}
This paper presents \dataset, a large-scale conflict event dataset comprising 39,786 events across 20 languages and 171 countries, with extensive coverage of region-specific entities. \dataset is based on a partially reannotated subset of the Armed Conflict Location \& Event Data (ACLED), which has documented global conflict events for over a decade.

To address the challenge of aggregating multilingual sources for global event analysis, we introduce \emph{abstractive} event extraction (\AEE) and its subtask, \emph{abstractive} entity linking (\AEL). Unlike conventional span-based event extraction, our approach detects event arguments and entities through holistic document understanding and normalizes them across the multilingual dataset.
We evaluate various large language models (LLMs) on these tasks, adapt existing zero-shot event extraction systems, and benchmark supervised models. Additionally, we introduce \system, a novel zero-shot retrieval-based system for \AEL.

Our best zero-shot system achieves an end-to-end $F_1$ score of 58.3\%, with LLMs outperforming specialized event extraction models such as GoLLIE. For entity linking, \system achieves an $F_1$ score of 45.7\%, significantly surpassing OneNet, a state-of-the-art zero-shot baseline that achieves only 23.7\%. However, these zero-shot results lag behind the best supervised systems by 20.1\% and 37.0\% in the end-to-end and \AEL tasks, respectively, highlighting the need for further research.~\footnote{The dataset and code are available at \url{https://github.com/stanford-oval/Lemonade}.}

\end{abstract}

\section{Introduction}
Event Extraction (EE) involves extracting structured information about events and their arguments from unstructured text, such as news articles. This task is fundamental for understanding and analyzing real-world phenomena at scale.

This paper refines the EE task to better serve the study of global real-world phenomena. As a case study, we analyze data from the Armed Conflict Location \& Event Data (ACLED), a non-profit organization that has systematically documented violent conflict and protest events worldwide for over a decade~\cite{raleigh2010introducing}. ACLED's data supports critical humanitarian work by organizations including the United Nations' International Organization for Migration, the International Rescue Committee, and the European Commission, who use it to track forced displacement and evaluate humanitarian interventions~\citep{acled2023impact}.

Based on this analysis, we introduce \dataset, a cleaned and partially reannotated version of the ACLED dataset tailored for NLP research. \dataset addresses several critical gaps in existing event extraction resources:

\paragraph{Multilinguality and Geographic Diversity}
To provide a truly global perspective, event extraction must extend beyond the Global North to include perspectives from the Global South and international regions~\cite{10.1371/journal.pone.0048596}. Unlike existing EE datasets that focus primarily on English or Chinese, \dataset encompasses events across 171 countries reported in 20 languages.

\paragraph{Tail Entity Coverage}

Entity linking is essential for aggregating information about event participants. While general-purpose entity databases like Wikidata~\citep{wen-etal-2021-resin} and Wikipedia~\citep{li-etal-2019-multilingual, li-etal-2020-gaia} offer broad coverage, they often lack specialized domain entities. \dataset addresses this gap with a database of 10,707 entities, including:
\begin{itemize}[noitemsep]
    \item Generic terms (e.g., ``Students'')
    \item Specialized political entities (e.g., ``Liwa' Al Hashemiyoun'', a Syrian political militia active since 2023)
    \item Regional organizations (e.g., ``NNO: Nagorik Nari Oikya'', the women's wing of Nagarik Oikya in Bangladesh)
\end{itemize}
Many of these entities lack Wikipedia entries, creating unique challenges for entity linking systems. This is particularly significant because most large language models (LLMs) are pre-trained on Wikipedia and tend to memorize common entities.~\footnote{This limitation extends beyond event datasets. For instance, ~\citet{cao-etal-2022-kqa} found that models without entity-linking modules achieved 90\% accuracy on a Wikidata question-answering dataset, suggesting over-reliance on memorized entities.}

\paragraph{Expert Annotations}
High-quality annotations are essential when EE systems inform high-stakes policy decisions, such as international peacemaking efforts~\citep{intercoder-reliability2011}, where annotation errors can lead to biased conclusions. While prior work~\citep{raleigh2010introducing, caselli-huang-2012-sourcing} has emphasized the importance of domain expertise, most existing document-level EE datasets rely on crowdsourcing~\cite{ebner-etal-2020-multi, liu-etal-2024-docee, ren-etal-2024-deie, wang-etal-2020-maven}, student annotators~\cite{li-etal-2021-document}, or weakly supervised methods~\cite{li-etal-2023-glen}. In contrast, ACLED employs about 200 regional experts who conduct multiple review rounds, ensuring high annotation quality and consistency.

Our contributions are threefold:

\begin{itemize}
\item
\textbf{The \emph{Abstractive} Event Extraction Task.}
Recognizing that one of the primary applications of event data is trend discovery and aggregate reporting~\citep{li-etal-2019-multilingual, li-etal-2020-connecting, li2021future, reddy2023smartbook}, we introduce the \AEE task. This task extracts events from complete documents following a structured codebook, requiring all event arguments to be normalized to numerical values, categorical labels, or entities from a predefined database.

\item \textbf{The \dataset Dataset.}
We present \dataset (\textbf{L}arge \textbf{E}xpert-Annotated \textbf{M}ulti\-lingual \textbf{O}ntology-\textbf{N}ormalized \textbf{A}bstractive \textbf{D}ocument-Level \textbf{E}vent) dataset, for the multilingual \AEE task. \dataset is a high-quality document-level dataset based on human expert annotations on real-world conflicts comprising 39,786 events
across 20 languages and 171 countries.

\item \textbf{Models for the Multilingual \AEE Task.}
We adapt and evaluate diverse models from existing literature alongside several LLMs on \dataset in both zero-shot and supervised settings. Additionally, we introduce \system, a novel multilingual ZEro-ShoT entity linking system that achieves 45.7\% accuracy on the entity linking subtask, substantially outperforming all zero-shot baselines.

\end{itemize}

\begin{figure*}[ht]
    \centering
    \includegraphics[width=0.9\linewidth]{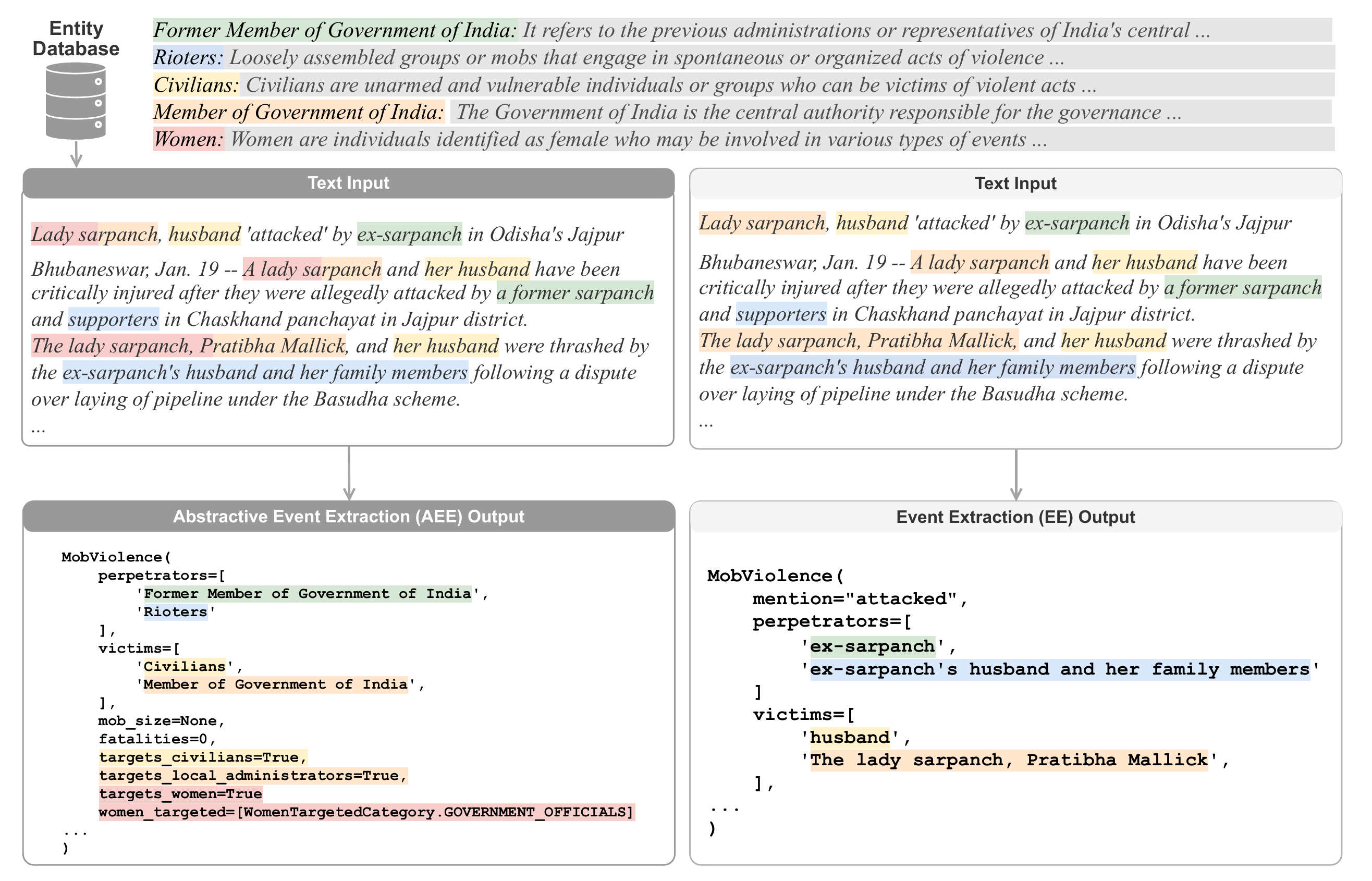}
    \caption{An example from \dataset showing abstractive event annotation. The input text and annotations are summarized for clarity. A hypothetical extractive annotation is included for comparison, illustrating the key differences between abstractive and extractive approaches.}
    
    \label{figure:lemonade-example}
\end{figure*}

\section{Related Work}

\paragraph{Event Extraction Task.}
The Message Understanding Conferences (MUC) pioneered the use of text spans as a unit for system outputs in information extraction. While the MUC-3 and MUC-4~\cite{sundheim-1992-overview, grishman-sundheim-1996-message} datasets originally included non-span event arguments, subsequent work has standardized evaluation on span-based arguments, as noted by \citet{gantt-etal-2024-multimuc} and \citet{chambers-jurafsky-2011-template}.

Contemporary EE research largely follows the ACE05 project's~\citep{ace05} task formulation, which decomposes event extraction into sentence-level subtasks using span-based intermediate annotations. Recent work has expanded this scope: \citet{li-etal-2021-document} extended EE to capture arguments from surrounding sentences and introduced the concept of the ``most informative span'' for argument selection. Building on this, \citet{tong-etal-2022-docee} introduced the DocEE dataset, where event arguments are, on average, scattered across 10 sentences in the document, establishing EE as a truly \emph{document-level} task.

\paragraph{Entity Detection and Linking.}
Entity Linking (EL) connects entity mentions in text to entries in a database~\cite{learningtolink, liu-etal-2024-spinach}. Traditional EL pipelines first detect entity mention spans, then disambiguate them against the target database. In contrast, our abstractive entity linking (\AEL) approach directly maps input text to a list of linked entities, without explicit span detection.

Zero-shot EL, which enables linking to new entity databases without direct supervision, is particularly relevant to our work. \citet{logeswaran2019zero} demonstrated the effectiveness of pre-trained models for zero-shot EL and introduced ZESHEL, now a popular benchmark. Recent advances include \citet{xu2023read}'s read-and-select framework using fine-tuned RoBERTa models for entity disambiguation, and OneNet~\cite{liu-etal-2024-onenet}, which achieves state-of-the-art performance through a three-module LLM-based pipeline.

\section{The Abstractive Event Extraction Task}
\label{section:task-definition}

The \AEE task shifts focus from surface text forms to grounding events in predefined ontologies and representing arguments as categorical, numerical, or string values.
\AEE removes two key constraints compared to traditional event extraction: arguments need not be text spans, and they need not be explicitly mentioned in the text.

\begin{definition}

An event extraction codebook $C = (T, \D, S)$ consists of:

\begin{itemize}[noitemsep]

\item $T$: the set of possible event types.

\item $\D$: a collection of domains, where each $D \in \D$  is a domain such as integers, strings, or a set of known entities.

 \item
    $S = [ (t_1, a_{1,1}, \ldots, a_{1, n_1}), \ldots,\\ (t_m, a_{m,1}, \ldots, a_{m, n_m})]$: a list of $m$ event signatures, where $n_i$ is the number of arguments for event type $t_i$, and each argument field $a_{i,j}$ in in domain $D_{i,j} \in \D$.

\end{itemize}
\end{definition}

\begin{definition}
The \textbf{Abstractive Event Extraction} (\AEE) task is: given codebook $C = (T, \D, S)$ and text $w$, extract abstractive event(s) of the form $(t_i, v_1, \ldots, v_{n_i})$ from $w$, where $t_i \in T$ is an event type, each $v_j \in D_{i,j}$ is a value for argument $a_j$, and $n_i$ is the number of arguments for event type $t_i$.
\end{definition}

Figure~\ref{figure:lemonade-example} demonstrates how \AEE can, for example, facilitate studying violence against women globally. For event type $t_i=\texttt{MobViolence} \in T$, its first two arguments, \texttt{perpetrators} and \texttt{victims} represent the two sides in the violence, with $D_{i, 1}, D_{i, 2}$ being the set of all subsets of possible entities from the event database. The seventh argument 
\texttt{targets\_women} is a boolean, thus $D_{i,7}$ is $\{\text{True}, \text{False}\}$.
The \AEE annotation captures critical information not explicitly stated as text spans in the input:
\begin{enumerate}[noitemsep]
\item Women were specifically targeted.
\item The targeted women were government officials.
\item No fatalities occurred.
\item The mob size was unspecified (\texttt{mob\_size=None}).
\end{enumerate}

These abstract arguments—represented as boolean, enum, and numerical fields—enable straightforward aggregation for analytical queries such as ``How many casualties resulted from violence against female government officials in 2024?''

\AEE eliminates the need for intermediate annotations such as event triggers and entity mentions~\cite{huang-etal-2024-textee}, avoiding the common practice of annotating multiple spans for the same argument. Instead, \AEE directly annotates events and their linked entities, enabling direct evaluation against gold annotations. This streamlined approach reduces annotation complexity, produces cleaner labels, and allows simple exact-match evaluation, addressing limitations of existing EE metrics~\cite{lu2024exact}.
The \AEE framework comprises three core subtasks:

\begin{itemize}[noitemsep]

\item \textbf{Event Detection (ED)}: Identify from codebook $C$ the event type(s) in text $w$. 

\begin{center}
$\text{ED}(w, C) = \{t_1, \ldots\} \subseteq T$
\end{center}

\item \textbf{Abstractive Event Argument Extraction (AEAE)}: Given gold event type $t$ in codebook $C$, extract non-entity arguments from $w$.

\begin{center}
$\text{AEAE}(w, C, t) = [v_1, \ldots]$
\end{center}

where $a_i$ are non-entity arguments.

\item \textbf{Abstractive Entity Linking (AEL)}: Given text $w$ and a gold event type $t$ in codebook $C$, identify relevant entities from the database and assign them to appropriate event arguments.

\begin{center}
$\text{AEL}(w, C, t) = [v_1, \ldots]$
\end{center}

where $a_i$ are entity arguments.
\end{itemize}

An end-to-end \AEE system first performs ED, then uses predicted event types (rather than gold types) to separately perform AEAE and AEL.

\section{\datasetWithEmoji: An \AEE Dataset}
\dataset spans 20 typologically diverse languages~\citep{clark-etal-2020-tydi}, ordered by the number of examples in \dataset from most to least:
English, Spanish, Arabic, French, Italian, Russian, German, Turkish, Burmese, Indonesian, Ukrainian, Korean, Portuguese, Dutch, Somali, Nepali, Chinese, Persian, Hebrew, and Japanese.

It surpasses existing event datasets in linguistic coverage and is the first event extraction dataset that includes Burmese, Indonesian, Hebrew, Somali, and Nepali. Table~\ref{table:dataset-comparison} compares \dataset with other document-level event datasets.

We construct \dataset by extending expert annotations from ACLED. To ensure compatibility with NLP systems, we reannotated several event argument types, transformed event-centric annotations into a document-centric format, and generated descriptions for 10,707 entities to facilitate retrieval-based entity linking.

Each example in \dataset consists of a news article with its primary event annotated, following the document-level single-event configuration established in DocEE~\cite{tong-etal-2022-docee, liu-etal-2024-docee}. The dataset includes 25 event types within the socio-political domain, ranging from peaceful protests to chemical weapons deployment. Annotations comprise the event type and associated entity and non-entity arguments with their corresponding roles. Appendix~\ref{appendix:dataset-schema} details all event types and their arguments.

\subsection{Dataset Construction}
\label{section:dataset-construction}
ACLED's \textit{original} annotations operate at the event level, with individual events potentially spanning multiple articles. These annotations integrate multiple sources, including maps and images, to determine event locations and participants. The primary challenge in developing \dataset involves ensuring document-level annotations contain only information extractable from individual documents. We summarize the construction process below; see Appendix~\ref{appendix:dataset-creation} for more details.

We utilize ACLED data spanning January 2024 to January 2025 (13 months), comprising 344,116 events.
Each event links to one or more source URLs, and has one corresponding event annotation. We filter URLs lacking substantive event information in text form (e.g., image-heavy social media posts) and keep news articles.
We obtain the full text from the provided URLs and clean the texts by removing advertisements and other extraneous content.
We then use GPT-4o for language detection.

\paragraph{Location Argument Reannotation} ACLED's original location annotations derive from multiple sources, including external maps and field reports. Since this information may not appear in article text, text-based extraction systems cannot reliably identify locations. We address this by reannotating all location arguments through automated tools and manual verification by the authors. Location arguments---from country to city block level---follow the guideline: ``\textit{The location argument is the most specific place supported by the text}.'' Therefore, EE systems are expected to extract location entirely from the text. During evaluation, we normalize locations using OpenStreetMaps, eliminating the need for EE systems to have detailed geographical knowledge of remote regions.

\paragraph{Schema Standardization} We refine the event schema and convert annotations to Python classes following~\citet{wang-etal-2023-code4struct}. This format enables structured decoding~\cite{dong2024xgrammar}, substantially improving performance of generative EE models.

\paragraph{Entity Database}
ACLED annotates the entities involved in each event, yielding a total of 10,707 unique entities. Note that this database is a superset of the 4,305 entities included in \dataset and the 2,648 entities in its development and test splits. Consequently, entity linking systems evaluated on this dataset must be proficient at distinguishing relevant entities from distracting ones.

Specialized domains require domain-specific knowledge for effective entity linking. We provide one-paragraph descriptions for each entity, supplying context and domain knowledge essential for understanding specialized entities, particularly long-tail instances~\citep{mallen-etal-2023-trust}. This approach parallels the Zeshel entity linking dataset design~\cite{logeswaran-etal-2019-zero}. Entity linking systems utilize these descriptions to identify the entities relevant to the input text. Appendix~\ref{appendix:entity-examples} includes sample entities and their descriptions.

\paragraph{Data Splits} We implement temporal splits: training data comprises events from January -- March 2024, while validation and test sets contain events from April 2024 -- January 2025. This design reflects real-world scenarios where event and entity distributions evolve temporally. Notably, 44.3\% of validation and test entities are absent from the training data. Validation and test sets are randomly divided. Appendix~\ref{appendix:dataset-stats} details the language, event type, and geographical distributions within \dataset.

\section{\system: A Novel Abstractive Entity Linker}

\system employs a multi-stage approach to linking entities: first, it leverages information retrieval techniques to narrow down candidate entities; second, it filters these candidates based on their relevance; and finally, it assigns each entity to the appropriate event argument. For instance, in Figure~\ref{figure:lemonade-example}, the entity ``Member of Government of India'' is assigned to the event argument \texttt{victims}.

\textbf{Stage 1: Entity Retrieval.} In the first stage, we construct a vector database by embedding all entities along with their descriptions using an embedding model. Given an input document, \system utilizes the underlying LLM to generate multiple queries for searching the entity database. These queries aim to closely match the descriptions of the gold entities, thus increasing the likelihood of retrieving relevant candidates. At test time, since the model does not have access to the gold entity descriptions, the LLM approximates these descriptions based solely on the information available in the input document.

For example, given the document shown in Figure~\ref{figure:lemonade-example}, the system generates multiple queries for possible entities, including:
``The former sarpanch of Chaskhand panchayat, involved in a political rivalry with the current sarpanch, Pratibha Mallick,'' and 
``The state government of Odisha, India, responsible for implementing development schemes and maintaining law and order in the region.''
The union of all entities retrieved by these queries is then passed to the next stage.

\textbf{Stage 2: Entity Filtering.} In the second stage, each candidate entity (along with its description) retrieved from Stage 1 is evaluated using a dedicated prompt (see Table~\ref{prompt:edl2}). This prompt helps determine whether there is sufficient evidence in the document to support the entity's involvement in the event. Entities lacking supporting evidence are removed from the candidate set.

\textbf{Stage 3: Entity Assignment.} In the final stage, the remaining entities are matched to their correct event arguments. To accomplish this, we employ another prompt (see Table~\ref{prompt:edl3}), which takes as input the list of filtered entities and the available event argument roles, and outputs the appropriate mapping between them.

\section{Experiments}
\subsection{Baselines for \AEE}

For \AEE and its various subtasks, we experiment with adapted versions of prior state-of-the-art solutions, as discussed below. Further details can be found in Appendix~\ref{appendix:experiment-details}. 

\textbf{All \AEE Tasks: Supervised LMs.} We fine-tune LLMs to autoregressively generate the complete structured output from the input document. The generated output begins with the event type, followed by event arguments and entities, all formatted in JSON. We experiment with several multilingual LLMs of varying sizes: the base versions of LLaMA-3.2 with 1B and 3B parameters, as well as LLaMA-3.1 with 8B parameters. Additionally, we evaluate Aya Expanse~\cite{dang2024ayaexpansecombiningresearch}, an 8B-parameter model specifically optimized for multilingual performance in 16 of the 20 languages covered by \dataset.

\textbf{All \AEE Tasks: GoLLIE.} For the \ED and \AEAE subtasks, we employ GoLLIE~\cite{sainz2024gollie}, a model specifically instruction-tuned from CodeLLaMA~\cite{rozire2023code} for information extraction tasks. We also use it as an entity span detection model in the \AEL subtask.

\textbf{\ED: XLM-R-RetroMAE (XLM-RRM).} For the \ED subtask, we fine-tune the XLM-R model~\cite{conneau-etal-2020-unsupervised}, whose context length was extended to 8,192 tokens by \citet{chen-etal-2024-m3} and further pre-trained using RetroMAE~\cite{xiao-etal-2022-retromae}. We select this model because it has been pre-trained on 100 languages and provides sufficient context length for \dataset. We refer to this model as XLM-RRM.

\textbf{\ED: In-Context Learning with LLMs.} Given that the set of event types ($T$) is relatively small (25 event types in \dataset), event detection can be naturally formulated as a zero-shot in-context learning task. We design a prompt (Table~\ref{prompt:ed}) that includes the input text $w$ and a list of event types with their descriptions. The model is tasked with returning the most likely event type $t$. We experiment with GPT-4o, GPT-4o mini, and the 8B-parameter instruction-tuned LLaMA-3.1~\cite{dubey2024llama}.

\textbf{\AEAE: Abstractive Code4Struct (AC4S).} For the \AEAE subtask, we develop Abstractive Code4Struct (AC4S) by modifying the instructions of Code4Struct~\cite{wang-etal-2023-code4struct} to adapt it to the document-level abstractive setting. Specifically, we instruct the LLM to directly output event arguments and their roles from the input article, rather than performing sentence-level span extraction as in the original paper. This is achieved using a prompt (Table~\ref{prompt:eae}) that, given the input text $w$ and the event type signature for $t$, outputs all non-entity argument values.

\textbf{\AEAE: Zero-shot Question Answering with LLMs.} Models that rely on question answering for event argument extraction~\cite{li-etal-2020-event, liu-etal-2020-event, choudhary-du-2024-qaevent, lu-etal-2023-event} can be naturally extended to our abstractive setting. We adopt the zero-shot LLM-based question generation method proposed by \citet{uddin-etal-2024-generating} as a baseline for \AEAE.

\textbf{\AEL: OneNet.} For the \AEL subtask, we adopt OneNet~\cite{liu-etal-2024-onenet}, a state-of-the-art few-shot entity linking model. OneNet leverages retrieval and entity descriptions to identify the best match for a given entity mention. We experiment with applying OneNet to entity mentions extracted by GoLLIE and GPT-4o.

We use greedy constrained decoding~\cite{shin-van-durme-2022-shot} for all models. For entity retrieval, we employ the mGTE embedding model~\cite{zhang-etal-2024-mgte}. To ensure a fair comparison, we modify OneNet and the QA baseline to use GPT-4o, and we further adapt OneNet to use mGTE for entity retrieval, just like \system. Additional details on the baselines are provided in Appendix~\ref{appendix:baselines}.

\subsection{Evaluation Metrics}
\label{section:metrics}
To evaluate a predicted event against a gold-standard event from \dataset, we first normalize location arguments by performing a lookup in the OpenStreetMap geographic database. We then use exact string matching to calculate precision, recall, and micro-averaged $F_1$ scores~\citep{intro-to-ir-2008}.

For event detection (\ED), we compare the predicted event type with the gold-standard event type and report the micro-averaged \textbf{ED $F_1$}. For abstractive event argument extraction (\AEAE), the model generates event arguments and their values ${(a'_1, v'_1), \dots}$. We treat this set as the predicted result and compute precision, recall, and $F_1$ scores against the gold set ${(a_1, v_1), \dots}$. We report this metric as \textbf{\AEAE $F_1$}. Two arguments are considered equal only if both their argument roles and values exactly match.

For entity linking, we report \textbf{\AEL $F_1$}, computed by comparing the predicted entity IDs with the gold-standard entity IDs. This calculation is similar to the \AEAE $F_1$ but considers only entity arguments.

Finally, in the end-to-end (E2E) setting, the system first predicts the event type and then uses that prediction to extract event arguments and entities. In this scenario, an incorrect event type prediction results in false positives for all predicted event arguments and entities, and false negatives for all gold-standard event arguments and entities.

\section{Results}
\label{section:results}

\subsection{Event Detection}

Table~\ref{tab:lemonade-ed-results} summarizes the results for the \ED subtask. In the zero-shot setting, GPT-4o achieves the highest performance across all languages, with an average $F_1$ score of 79.6. GPT-4o mini trails by 9.8 points, achieving an $F_1$ score of 69.8, while the 8B-parameter Llama 3.1 lags further behind by an additional 10.3 points. Llama 3.1 8B performs particularly poorly for Indonesian (id) and Somali (so). The variation in performance across different languages increases from GPT-4o to GPT-4o mini to Llama 3.1 8B, with the differences in performance between the best and worst languages being 18.7, 40.2, and 61.2 points, respectively. Interestingly, the performance ranking is consistent across the three models for almost all languages, with the exception that GPT-4o mini is significantly worse in Chinese (zh) than Llama 3.1 8B.

GoLLIE 7B performs significantly worse than all other models, even compared to the similarly sized Llama model in English, a language on which GoLLIE was specifically instruction-tuned. This suggests that instruction-tuning on extractive event datasets does not readily transfer to our abstractive setting.

In the supervised setting, all models, from the smallest 0.5B-parameter XLM-RRM to the larger 8B-parameter models, achieve similar performance. Aya Expanse slightly outperforms the other models on average, achieving an $F_1$ score of 87.5. XLM-RRM performs particularly well on Burmese (my), Indonesian (id), Somali (so), and Japanese (ja), surpassing other models. This advantage is likely due to XLM-RRM being the only model in this group pre-trained on Burmese and Somali.

Overall, there is a 7.9 percentage-point performance gap between the best zero-shot model (GPT-4o) and the best supervised model (Aya Expanse). Remarkably, GPT-4o surpasses supervised models on Burmese, Somali, and Hebrew, which are among the lowest-resource languages in our dataset.

The variance in performance across languages can be partially explained by the differences in event types present in each language (see Table~\ref{table:dataset-diversity}). Since the ACLED data aims to reflect real-world events, the distribution of event types in each language is heavily influenced by the political stability of countries where that language is spoken. For instance, nearly all events in Korean and Japanese are Peaceful Protest events. To further analyze this phenomenon, we include a simple baseline, \emph{majority class}~\cite{mohammad-etal-2016-semeval}, which always predicts the most frequent event type for each language. We observe extremely high scores for all supervised models on Korean and Japanese. Italian (it), German (de), Indonesian (id), and Chinese (zh) follow a similar pattern, though to a lesser extent.

\begin{table*}[!ht]
\centering
\begin{adjustbox}{center, width=0.9\linewidth}
\setlength{\tabcolsep}{2pt} 
\small
\begin{tabular}{l*{20}{r}r}
\toprule
\textbf{Model} &All &en &es &ar &fr &it &ru &de &tr &my &id &uk &ko &pt &nl &so &ne &zh &fa &he &ja \\
\midrule
\multicolumn{22}{c}{\textbf{Zero-shot}} \\[3pt]
GPT-4o&\textbf{79.6}&\textbf{72.2}&\textbf{76.0}&\textbf{73.8}&\textbf{73.0}&\textbf{89.0}&\textbf{76.6}&\textbf{88.8}&\textbf{84.8}&\textbf{71.4}&\textbf{78.0}&\textbf{75.6}&\textbf{85.6}&\textbf{74.2}&\textbf{90.1}&\textbf{80.4}&\textbf{77.4}&\textbf{85.0}&\textbf{83.8}&\textbf{76.2}&\textbf{86.0}\\ 

GPT-4o mini&  69.8&65.0&66.8&65.2&64.4&85.4&68.8&83.6&77.0&51.6&74.2&68.8&71.8&72.4&86.6&58.1&64.5&46.4&81.6&66.9&85.7\\
Llama 3.1 8B&  59.5&55.0&55.0&54.0&51.8&85.2&45.4&79.4&65.0&37.8&76.4&57.6&51.2&56.0&76.4&24.0&62.6&66.8&66.8&50.9&75.4\\
GoLLIE 7B&23.6&35.8&36.8&6.2&34.6&49.6&29.0&61.2&7.8&0.2&11.2&41.0&1.4&46.2&36.6&0.0&0.2&38.6&6.2&5.4&7.4\\
\midrule
\multicolumn{22}{c}{\textbf{Trained on \dataset}} \\[3pt]
XLM-RRM&85.0&76.8&83.0&76.0&73.0&95.0&74.8&93.2&94.8&\textbf{69.2}&\textbf{94.2}&75.6&98.6&88.2&92.6&\textbf{74.6}&89.1&94.6&88.0&72.3&\textbf{98.5}\\
    
Llama 3.2 1B&85.0&79.6&85.4&80.4&74.8&97.0&81.6&93.0&94.2&60.6&92.0&76.0&99.2&89.6&\textbf{95.4}&65.1&88.4&95.4&88.2&65.1&98.2\\ 
Llama 3.2 3B&86.6&81.4&86.6&\textbf{82.8}&77.2&97.0&82.8&\textbf{94.8}&95.4&68.8&93.2&77.4&99.2&89.8&94.4&66.2&88.6&96.2&89.2&70.8&97.8\\
Llama 3.1 8B&86.2&\textbf{82.0}&\textbf{87.2}&80.6&77.0&97.4&\textbf{83.4}&93.8&94.0&63.8&92.2&77.0&99.0&90.8&94.0&69.8&87.7&96.4&88.8&69.9&97.8\\
Aya Expanse 8B&\textbf{87.5}&80.4&87.0&82.6&\textbf{79.6}&\textbf{97.6}&83.2&\textbf{94.8}&\textbf{96.0}&66.2&92.8&\textbf{80.2}&\textbf{99.6}&\textbf{91.8}&\textbf{95.4}&70.9&\textbf{89.3}&\textbf{96.6}&\textbf{91.4}&\textbf{75.9}&98.2\\
\midrule
Majority Class &50.4&31.0&33.0&15.8&29.8&91.6&23.2&86.2&66.0&19.0&86.0&40.4&98.8&42.6&82.4&49.4&77.0&89.4&63.2&40.1&98.5\\
\bottomrule
\end{tabular}
\end{adjustbox}
\caption{ED $F_1$ results on the \dataset test set. The best result in each setting is highlighted in bold.}
\label{tab:lemonade-ed-results}
\end{table*}

\subsection{Abstractive Event Argument Extraction}
\begin{table*}[!ht]
\centering
\begin{adjustbox}{center, width=0.9\linewidth}
\setlength{\tabcolsep}{2pt} 
\small
\begin{tabular}{l*{20}{r}r}
\toprule
\textbf{Model} &
All&en&es&ar&fr&it&ru&de&tr&my&id&uk&ko&pt&nl&so&ne&zh&fa&he&ja \\
\midrule
\multicolumn{22}{c}{\textbf{Zero-shot}} \\[3pt]
AC4S (GPT-4o)&\textbf{84.6}&\textbf{85.2}&\textbf{91.0}&\textbf{73.1}&\textbf{85.0}&94.1&\textbf{82.9}&\textbf{90.9}&\textbf{89.0}&\textbf{70.9}&\textbf{93.2}&\textbf{74.4}&\textbf{90.2}&\textbf{91.1}&\textbf{92.2}&\textbf{72.1}&\textbf{87.9}&\textbf{89.3}&\textbf{83.6}&\textbf{64.4}&\textbf{94.4}\\ 
AC4S (GPT-4o mini)&81.0&83.1&87.1&71.1&84.1&\textbf{94.3}&80.1&89.7&86.4&60.0&91.5&73.5&82.2&81.7&89.0&68.3&82.5&83.2&82.9&63.4&90.9\\
AC4S (Llama 3.1 8B)&49.0&58.9&56.4&15.5&57.5&55.0&57.1&68.1&50.5&41.7&47.4&36.5&51.4&49.2&65.5&38.1&47.5&36.6&42.8&50.9&50.2\\
QA (GPT-4o)&75.3&74.0&79.7&56.3&73.3&88.5&57.2&86.7&83.0&61.7&87.7&61.2&79.2&82.3&91.2&64.5&80.8&79.0&79.7&65.0&84.2\\
GoLLIE 7B&40.0&47.5&45.9&21.9&46.8&54.1&42.7&59.9&27.5&1.3&49.1&39.3&30.7&49.7&54.4&0.7&12.6&58.3&31.7&22.0&49.2\\
\midrule
\multicolumn{22}{c}{\textbf{Trained on \dataset}} \\[3pt]
Llama 3.2 1B&
    85.4&
    87.1&
    85.9&
    78.6&
    81.2&
    94.3&
    78.8&
    91.6&
    90.7&
    71.4&
    93.8&
    84.6&
    94.5&
    95.1&
    94.4&
    71.6&
    88.0&
    86.1&
    77.1&
    72.5&
    86.0\\ 
Llama 3.2 3B&
    87.7&
    \textbf{89.0}&
    88.4&
    79.7&
    86.3&
    95.5&
    83.5&
    93.5&
    93.2&
    \textbf{77.3}&
    95.0&
    85.4&
    96.1&
    95.7&
    95.0&
    75.6&
    90.2&
    87.3&
    79.8&
    75.4&
    87.1\\
Llama 3.1 8B&
    87.6&
    88.5&
    89.7&
    80.4&
    87.2&
    96.2&
    83.8&
    94.1&
    92.1&
    76.4&
    94.5&
    85.1&
    95.8&
    95.7&
    94.7&
    \textbf{76.6}&
    \textbf{91.0}&
    89.2&
    78.3&
    71.9&
    85.8\\
Aya Expanse 8B&
    \textbf{89.0}&
    88.9&
    \textbf{90.5}&
    \textbf{81.4}&
    \textbf{88.3}&
    \textbf{97.7}&
    \textbf{85.2}&
    \textbf{94.2}&
    \textbf{93.5}&
    76.3&
    \textbf{96.3}&
    \textbf{87.5}&
    \textbf{97.2}&
    \textbf{96.1}&
    \textbf{95.7}&
    75.4&
    90.3&
    \textbf{91.6}&
    \textbf{82.8}&
    \textbf{77.8}&
    \textbf{92.6}\\
\bottomrule
\end{tabular}
\end{adjustbox}
\caption{AEAE $F_1$ results on the \dataset test set. The best result in each setting is highlighted in bold.}
\label{tab:lemonade-aeae-results}
\end{table*}

Table~\ref{tab:lemonade-aeae-results} presents the results for the \AEAE subtask. 
In the supervised setting, Aya Expanse achieves the best overall performance, reaching an average $F_1$ score of 89\%, ranging from 76\% for Burmese (my) to 97\% for Italian (it). It consistently ranks either first or within 1.3 percentage points of the top-performing model across all languages. The Llama models are within 4.6\% on average compared to the Aya Expanse model.

In the zero-shot setting, Abstractive Code4Struct with GPT-4o outperforms all other models, achieving performance within 4.4 percentage points of the best supervised model. Notably, it even surpasses the best supervised model by 0.5 to 1.8 percentage points on Spanish (es), Farsi (fa), and Japanese (ja). 

The QA-based model performs, on average, 9.3 percentage points worse, indicating that directly generating event arguments is more effective than formulating the task as question answering. Similar to the event detection results, GoLLIE performs worse than all other models, even in English, and completely fails on Burmese and Somali. This poor performance can be attributed to the limited multilingual capabilities of its underlying base model, CodeLLaMA.

\subsection{Abstractive Entity Linking}

\begin{table*}[!ht]
\centering
\begin{adjustbox}{center, width=0.85\linewidth}
\setlength{\tabcolsep}{2pt} 
\small
\begin{tabular}{l*{20}{r}r}
\toprule
\textbf{Model} &All&en&es&ar&fr&it&ru&de&tr&my&id&uk&ko&pt&nl&so&ne&zh&fa&he&ja \\
    
\midrule
\multicolumn{22}{c}{\textbf{Zero-shot}} \\[3pt]

\system (GPT-4o)&
    \textbf{45.7}&
    \textbf{49.7}&
    \textbf{46.6}&
    \textbf{46.0}&
    \textbf{52.2}&
    \textbf{44.8}&
    \textbf{42.3}&
    \textbf{41.8}&
    \textbf{43.7}&
    \textbf{44.8}&
    \textbf{45.1}&
    \textbf{51.2}&
    \textbf{37.7}&
    \textbf{50.2}&
    \textbf{52.8}&
    \textbf{55.2}&
    \textbf{46.4}&
    \textbf{55.2}&
    \textbf{56.4}&
    \textbf{33.3}&
    \textbf{22.4}\\
\system (GPT-4o mini)&
    27.2&
    34.1&
    28.7&
    31.8&
    36.8&
    20.2&
    28.5&
    19.7&
    24.4&
    28.8&
    19.2&
    50.5&
    15.6&
    31.3&
    26.6&
    39.7&
    22.1&
    26.2&
    26.6&
    31.2&
    11.2\\
Span (GoLLIE 7B) + OneNet&
    11.1&
    18.7&
    13.0&
    7.8&
    19.3&
    13.4&
    12.4&
    21.1&
    8.7&
    0.0&
    6.9&
    24.9&
    4.3&
    11.8&
    18.7&
    0.0&
    0.4&
    5.2&
    11.1&
    1.7&
    4.6\\
Span (GPT-4o) + OneNet &
    23.7&
    26.0&
    20.8&
    30.7&
    31.1&
    28.4&
    16.5&
    28.9&
    28.5&
    10.9&
    16.5&
    25.6&
    18.8&
    18.2&
    30.1&
    41.1&
    18.7&
    9.8&
    20.9&
    22.0&
    19.1\\
    
\midrule
\multicolumn{22}{c}{\textbf{Trained on \dataset}} \\[3pt]

Llama 3.2 1B&
    81.9&
    79.2&
    \textbf{81.7}&
    79.1&
    72.7&
    85.1&
    \textbf{81.7}&
    82.0&
    \textbf{87.9}&
    67.9&
    89.7&
    90.0&
    87.5&
    86.2&
    84.8&
    59.4&
    82.9&
    90.7&
    84.9&
    78.5&
    80.7\\
Llama 3.2 3B&
    82.1&
    79.6&
    81.0&
    80.5&
    72.7&
    85.2&
    81.2&
    80.7&
    86.4&
    \textbf{70.0}&
    89.8&
    \textbf{90.2}&
    88.1&
    86.9&
    85.0&
    \textbf{62.7}&
    85.7&
    91.0&
    84.5&
    78.4&
    79.3\\
Llama 3.1 8B&
    80.0&
    78.9&
    78.8&
    80.1&
    68.0&
    82.8&
    80.6&
    79.4&
    85.0&
    66.6&
    88.5&
    88.5&
    85.4&
    84.4&
    84.3&
    57.6&
    82.1&
    89.5&
    83.3&
    76.6&
    78.8\\
Aya Expanse 8B&
    \textbf{82.7}&
    \textbf{79.8}&
    80.5&
    \textbf{81.2}&
    \textbf{74.3}&
    \textbf{86.2}&
    \textbf{81.7}&
    \textbf{82.1}&
    87.5&
    69.6&
    \textbf{90.5}&
    89.4&
    \textbf{88.7}&
    \textbf{87.1}&
    \textbf{85.1}&
    60.9&
    \textbf{86.1}&
    \textbf{91.3}&
    \textbf{85.1}&
    \textbf{83.0}&
    \textbf{81.2}\\
\bottomrule
\end{tabular}
\end{adjustbox}
\caption{\AEL $F_1$ results on the \dataset test set. The best result in each setting is highlighted in bold.}
\label{tab:lemonade-ael-languages-results}
\end{table*}

Table~\ref{tab:lemonade-ael-languages-results} presents the complete results for the \AEL subtask.
In the zero-shot setting, our proposed method, \system (GPT-4o), achieves an $F_1$ score of 45.7\%, substantially outperforming all baselines, including the state-of-the-art OneNet model (also using GPT-4o), by 20.0 percentage points.
When using GoLLIE to extract entity spans, OneNet's performance drops significantly, achieving only an 11.1\% $F_1$ score. 
These results demonstrate that span detection is a critical limiting factor for entity linking performance on \dataset, as the dataset contains many abstractive entities. Appendix~\ref{appendix:ael-system-outputs} provides side-by-side examples of system outputs.

All supervised models significantly outperform zero-shot models, with Aya Expanse achieving the best average performance.
To better understand this performance gap, we further analyze entity linking performance across several subsets of entities. Table~\ref{tab:lemonade-ael-subsets-results} compares results for entities appearing in the \dataset training set (\emph{Seen}) versus those not appearing (\emph{Unseen}), as well as for {\em Generic entities} (e.g., ``Student'') versus {\em Specific} entities (e.g., ``Government of Panama''). 
We observe that supervised methods lag behind zero-shot methods (\system and OneNet) in the \emph{Unseen} category. Additionally, while supervised models exhibit a notable performance drop for {\em Specific} entities, the decline is much smaller for \system (7.0\% compared to 17.4\%), with OneNet performing even better in this regard.

While models generally perform well on the \emph{seen} entities, all significantly struggle with new entities, with the best model achieving only 30.4\%.

\begin{table*}[!ht]
\centering
\begin{adjustbox}{center, width=0.9\linewidth}
\setlength{\tabcolsep}{2pt} 
\small
\begin{tabular}{lrrrrr}
\toprule
&All&Seen&Unseen&Generic&Specific\\
    
\midrule
\multicolumn{6}{c}{\textbf{Zero-shot}} \\[3pt]

\system (GPT-4o)&45.7&48.9&20.0&49.6&42.6\\
\system (GPT-4o mini)&27.2&31.5&8.0&31.1&25.0\\
Span (GoLLIE 7B) + OneNet (GPT-4o)&11.1&10.9&14.7&7.2&15.9\\

Span (GPT-4o) + OneNet &23.7&23.2&30.4&10.5&37.2\\
    
\midrule
\multicolumn{6}{c}{\textbf{Trained on \dataset}} \\[3pt]

Llama 3.2 1B&    81.9&    83.7&    8.8&    89.4&    70.9\\

Llama 3.2 3B&82.1&83.9&10.6&89.2&71.8\\

Llama 3.1 8B&80.0 &81.8&9.4&88.3&68.0\\

Aya Expanse 8B&82.7&84.5&12.6&89.8&72.4\\
\bottomrule
\end{tabular}
\end{adjustbox}
\caption{\AEL $F_1$ results on the \dataset test set, grouped by entity categories.}
\label{tab:lemonade-ael-subsets-results}
\end{table*}

\subsection{End-to-End Results}
\begin{table*}[!ht]
\centering
\begin{adjustbox}{center, width=0.9\linewidth}
\small
\begin{tabular}{llllrr}
\toprule
\textbf{Training Data} & \textbf{\ED} & \textbf{\AEAE} & \textbf{\AEL} & \textbf{All} & \textbf{English} \\
\midrule

- & GPT-4o & AC4S (GPT-4o) & Zest (GPT-4o)        & \textbf{58.3} & \textbf{55.9}  \\ 
- & GPT-4o & AC4S (GPT-4o) & Span (GPT-4o) + OneNet   & 54.6 & 51.0  \\ 
- & Llama 3.1 8B   & AC4S (Llama 3.1 8B) & Zest (Llama 3.1 8B) & 20.6 & 21.2  \\
- & GoLLIE 7B   & GoLLIE 7B & Span (GoLLIE 7B) + OneNet    & 14.2 & 18.3  \\

\midrule
\dataset (all of train set) & \multicolumn{3}{c}{Aya Expanse 8B} & \textbf{78.4} & \textbf{71.6} \\

\dataset (10\% of train set) & \multicolumn{3}{c}{Aya Expanse 8B} & 68.2 & 65.0 \\

\dataset (5\% of train set) & \multicolumn{3}{c}{Aya Expanse 8B} & 65.5 & 59.2 \\
\dataset (1\% of train set) & \multicolumn{3}{c}{Aya Expanse 8B} & 57.9 & 48.9 \\
\dataset (English subset of train set) & \multicolumn{3}{c}{Aya-Expanse 8B} & 64.0 & 71.3 \\

\bottomrule
\end{tabular}
\end{adjustbox}
\caption{End-to-end $F_1$ results on the \dataset test set. The best result in each setting is highlighted in bold. Supervised experiments include training on the entire training set of \dataset, training on randomly sampled subsets of it, and only on its English subset.}
\label{tab:lemonade-end-to-end-results}
\end{table*}

Table~\ref{tab:lemonade-end-to-end-results} summarizes the end-to-end (E2E) results for selected combinations of subtask systems, evaluated across all languages and specifically on English. 

Among the zero-shot systems, the pipeline combining GPT-4o and \system achieves the highest performance, with an $F_1$ score of 58.3\%. In contrast, the best supervised model achieves an $F_1$ score of 78.4\%, representing a 20.1\% improvement over the best zero-shot system.
We also see that the quality of the underlying LLM is important, as for example, switching from GPT-4o to Llama 3.1 8B reduces the overall score by 37.7\%. As expected, GoLLIE performs worse than its similarly-sized model in all settings.

We also investigate the impact of training data availability in the supervised setting. When fine-tuning Aya Expanse solely on the English subset, overall performance drops by 14.4 percentage points, although the performance on English remains nearly unchanged. Reducing the overall amount of training data negatively impacts performance on both English and non-English languages. Notably, we observe that the best zero-shot model performs comparably to a supervised model trained on 1 -- 5\% (214 -- 1,007 examples) of the training data.

\subsection{Discussion}
From the results on the \ED and \AEAE subtasks, we can conclude that for many languages, the best models perform reasonably well, and can perhaps be used in practice to augment (but not replace) manual news monitoring efforts in this domain. Notable exceptions are Somali and Burmese languages where even the best models lag behind.
The \AEL subtask, however, paints a different picture as all models struggle with unseen entities.

As such, we believe future work on \dataset can especially focus on 1) entity linking for unseen entities, and 2) closing the performance gap between supervised and zero-shot models on all subtasks.

\section{Conclusion}

In this paper, we introduced the task of abstractive event extraction (\AEE), a formulation that better aligns with the requirements of real-world event extraction applications. To support research in this direction, we created a large-scale, high-quality dataset for \AEE in 20 languages, derived from expert-annotated data provided by ACLED.

Our experiments demonstrate that existing span-based models, such as GoLLIE and OneNet, are inherently unsuitable for the abstractive setting, consistently performing worse than models based on in-context learning.

Additionally, we proposed a novel zero-shot entity linking system, \system, which significantly narrows the performance gap in the abstractive entity linking (\AEL) subtask. Despite this improvement, a substantial gap remains between zero-shot and fully supervised models. We hope that the release of \dataset will inspire further research, ultimately expanding the capabilities of future zero-shot event extraction models.

\section*{Limitations}
This paper focuses on document-level event extraction and does not address event coreference resolution across multiple documents~\cite{eirew-etal-2022-cross}, which is essential for aggregate event analysis.
Existing event coreference methods, such as those proposed by \citet{gao-etal-2024-enhancing}, could potentially be adapted to the abstractive event extraction (\AEE) setting. We leave this promising direction for future research.

Additionally, \dataset currently excludes other common information extraction tasks, such as relation extraction, and provides annotations only for event and entity extraction.

Finally, the domain of \dataset is limited to violent conflict and protest events, emphasizing subtle distinctions between closely related event types. For instance, a peaceful protest met with excessive force is treated as a distinct event type from one without such force. In contrast, datasets like GLEN~\cite{li-etal-2023-glen} offer broader topical coverage, encompassing events ranging from conflicts to sports and other domains.

\section*{Ethics Statement}

No human subjects were involved in this study, and no crowdsourced annotations were performed. All annotation tasks were conducted either by expert annotators from ACLED or by the authors of this paper.
The dataset is derived exclusively from publicly accessible news articles and does not include personally identifiable information (such as names, addresses, or mobile device identifiers) of private individuals.

\section*{Acknowledgment}
We thank Professor Clionadh Raleigh, the CEO of ACLED, for her assistance and insights on this project. We also thank the reviewers for their valuable comments and suggestions.
This work is supported in part by 
the Verdant Foundation and Microsoft Azure AI credits.

\bibliography{anthology, custom}

\begin{thebibliography}{91}
\providecommand{\natexlab}[1]{#1}

\bibitem[{{ACLED}(2020)}]{acled2020coding}
{ACLED}. 2020.
\newblock \href {https://acleddata.com/acleddatanew/wp-content/uploads/2021/11/ACLED_Coding-Review-Process_v2_September-2020.pdf} {Coding review process}.
\newblock Accessed: 2024-09-30.

\bibitem[{{ACLED}(2023)}]{acled2023impact}
{ACLED}. 2023.
\newblock \href {https://acleddata.com/acleddatanew/wp-content/uploads/2024/07/2023-Impact-Report-Final-July-17.pdf} {2023 impact report}.
\newblock Accessed: 2024-09-28.

\bibitem[{ACLED(2023)}]{acled_codebook_2023}
ACLED. 2023.
\newblock Armed conflict location \& event data project (acled) codebook.
\newblock https://acleddata.com/knowledge-base/codebook/.
\newblock Last updated: 27 September 2024.

\bibitem[{Andrea~Ruggeri and Dorussen(2011)}]{intercoder-reliability2011}
Theodora-Ismene~Gizelis Andrea~Ruggeri and Han Dorussen. 2011.
\newblock \href {https://doi.org/10.1080/03050629.2011.596028} {Events data as bismarck's sausages? intercoder reliability, coders' selection, and data quality}.
\newblock \emph{International Interactions}, 37(3):340--361.

\bibitem[{Balali et~al.(2022)Balali, Asadpour, and Jafari}]{Balali_2022}
Ali Balali, Masoud Asadpour, and Seyed~Hossein Jafari. 2022.
\newblock \href {https://doi.org/10.2139/ssrn.4117538} {Cofee: A comprehensive ontology for event extraction from text}.
\newblock \emph{SSRN Electronic Journal}.

\bibitem[{Bentivogli et~al.(2010)Bentivogli, Forner, Giuliano, Marchetti, Pianta, and Tymoshenko}]{bentivogli-etal-2010-extending}
Luisa Bentivogli, Pamela Forner, Claudio Giuliano, Alessandro Marchetti, Emanuele Pianta, and Kateryna Tymoshenko. 2010.
\newblock \href {https://aclanthology.org/W10-3503/} {Extending {E}nglish {ACE} 2005 corpus annotation with ground-truth links to {W}ikipedia}.
\newblock In \emph{Proceedings of the 2nd Workshop on {T}he {P}eople`s {W}eb {M}eets {NLP}: {C}ollaboratively {C}onstructed {S}emantic {R}esources}, pages 19--27, Beijing, China. Coling 2010 Organizing Committee.

\bibitem[{Braha(2012)}]{10.1371/journal.pone.0048596}
Dan Braha. 2012.
\newblock \href {https://doi.org/10.1371/journal.pone.0048596} {Global civil unrest: Contagion, self-organization, and prediction}.
\newblock \emph{PLOS ONE}, 7(10):1--9.

\bibitem[{Cao et~al.(2022)Cao, Shi, Pan, Nie, Xiang, Hou, Li, He, and Zhang}]{cao-etal-2022-kqa}
Shulin Cao, Jiaxin Shi, Liangming Pan, Lunyiu Nie, Yutong Xiang, Lei Hou, Juanzi Li, Bin He, and Hanwang Zhang. 2022.
\newblock \href {https://doi.org/10.18653/v1/2022.acl-long.422} {{KQA} pro: A dataset with explicit compositional programs for complex question answering over knowledge base}.
\newblock In \emph{Proceedings of the 60th Annual Meeting of the Association for Computational Linguistics (Volume 1: Long Papers)}, pages 6101--6119, Dublin, Ireland. Association for Computational Linguistics.

\bibitem[{Caselli and Huang(2012)}]{caselli-huang-2012-sourcing}
Tommaso Caselli and Chu-Ren Huang. 2012.
\newblock \href {https://aclanthology.org/C12-2121/} {Sourcing the crowd for a few good ones: Event type detection}.
\newblock In \emph{Proceedings of {COLING} 2012: Posters}, pages 1239--1248, Mumbai, India. The COLING 2012 Organizing Committee.

\bibitem[{Chambers and Jurafsky(2011)}]{chambers-jurafsky-2011-template}
Nathanael Chambers and Dan Jurafsky. 2011.
\newblock \href {https://aclanthology.org/P11-1098/} {Template-based information extraction without the templates}.
\newblock In \emph{Proceedings of the 49th Annual Meeting of the Association for Computational Linguistics: Human Language Technologies}, pages 976--986, Portland, Oregon, USA. Association for Computational Linguistics.

\bibitem[{Chen et~al.(2024)Chen, Xiao, Zhang, Luo, Lian, and Liu}]{chen-etal-2024-m3}
Jianlyu Chen, Shitao Xiao, Peitian Zhang, Kun Luo, Defu Lian, and Zheng Liu. 2024.
\newblock \href {https://doi.org/10.18653/v1/2024.findings-acl.137} {{M}3-embedding: Multi-linguality, multi-functionality, multi-granularity text embeddings through self-knowledge distillation}.
\newblock In \emph{Findings of the Association for Computational Linguistics: ACL 2024}, pages 2318--2335, Bangkok, Thailand. Association for Computational Linguistics.

\bibitem[{Choudhary and Du(2024)}]{choudhary-du-2024-qaevent}
Milind Choudhary and Xinya Du. 2024.
\newblock \href {https://aclanthology.org/2024.findings-eacl.126/} {{QAEVENT}: Event extraction as question-answer pairs generation}.
\newblock In \emph{Findings of the Association for Computational Linguistics: EACL 2024}, pages 1860--1873, St. Julian{'}s, Malta. Association for Computational Linguistics.

\bibitem[{Clark et~al.(2020)Clark, Choi, Collins, Garrette, Kwiatkowski, Nikolaev, and Palomaki}]{clark-etal-2020-tydi}
Jonathan~H. Clark, Eunsol Choi, Michael Collins, Dan Garrette, Tom Kwiatkowski, Vitaly Nikolaev, and Jennimaria Palomaki. 2020.
\newblock \href {https://doi.org/10.1162/tacl_a_00317} {{T}y{D}i {QA}: A benchmark for information-seeking question answering in typologically diverse languages}.
\newblock \emph{Transactions of the Association for Computational Linguistics}, 8:454--470.

\bibitem[{Colruyt et~al.(2023)Colruyt, Clercq, Desot, and Hoste}]{dutch-ee}
Camiel Colruyt, Orphée~De Clercq, Thierry Desot, and Véronique Hoste. 2023.
\newblock \href {https://doi.org/10.1007/s10579-022-09623-2} {Eventdna: a dataset for dutch news event extraction as a basis for news diversification}.
\newblock \emph{Language Resources and Evaluation}, 57(1):189--221.

\bibitem[{Colvin et~al.(2024)Colvin, Jolibois, Ramezani, Badaracco, Dorsey, Montague, Matveenko, Trylesinski, Runkle, Hewitt, and Hall}]{pydantic}
Samuel Colvin, Eric Jolibois, Hasan Ramezani, Adrian~Garcia Badaracco, Terrence Dorsey, David Montague, Serge Matveenko, Marcelo Trylesinski, Sydney Runkle, David Hewitt, and Alex Hall. 2024.
\newblock \href {https://docs.pydantic.dev/latest/} {Pydantic}.

\bibitem[{Conneau et~al.(2020)Conneau, Khandelwal, Goyal, Chaudhary, Wenzek, Guzm{\'a}n, Grave, Ott, Zettlemoyer, and Stoyanov}]{conneau-etal-2020-unsupervised}
Alexis Conneau, Kartikay Khandelwal, Naman Goyal, Vishrav Chaudhary, Guillaume Wenzek, Francisco Guzm{\'a}n, Edouard Grave, Myle Ott, Luke Zettlemoyer, and Veselin Stoyanov. 2020.
\newblock \href {https://doi.org/10.18653/v1/2020.acl-main.747} {Unsupervised cross-lingual representation learning at scale}.
\newblock In \emph{Proceedings of the 58th Annual Meeting of the Association for Computational Linguistics}, pages 8440--8451, Online. Association for Computational Linguistics.

\bibitem[{Cunha et~al.(2024)Cunha, Silvano, Campos, and Jorge}]{ace05-pt}
Lu\'{\i}s~Filipe Cunha, Purifica\c{c}\~{a}o Silvano, Ricardo Campos, and Al\'{\i}pio Jorge. 2024.
\newblock \href {https://doi.org/10.1145/3626772.3657872} {Ace-2005-pt: Corpus for event extraction in portuguese}.
\newblock In \emph{Proceedings of the 47th International ACM SIGIR Conference on Research and Development in Information Retrieval}, SIGIR '24, page 661–666, New York, NY, USA. Association for Computing Machinery.

\bibitem[{Dang et~al.(2024)Dang, Singh, D'souza, Ahmadian, Salamanca, Smith, Peppin, Hong, Govindassamy, Zhao, Kublik, Amer, Aryabumi, Campos, Tan, Kocmi, Strub, Grinsztajn, Flet-Berliac, Locatelli, Lin, Talupuru, Venkitesh, Cairuz, Yang, Chung, Ko, Shi, Shukayev, Bae, Piktus, Castagné, Cruz-Salinas, Kim, Crawhall-Stein, Morisot, Roy, Blunsom, Zhang, Gomez, Frosst, Fadaee, Ermis, Üstün, and Hooker}]{dang2024ayaexpansecombiningresearch}
John Dang, Shivalika Singh, Daniel D'souza, Arash Ahmadian, Alejandro Salamanca, Madeline Smith, Aidan Peppin, Sungjin Hong, Manoj Govindassamy, Terrence Zhao, Sandra Kublik, Meor Amer, Viraat Aryabumi, Jon~Ander Campos, Yi-Chern Tan, Tom Kocmi, Florian Strub, Nathan Grinsztajn, Yannis Flet-Berliac, Acyr Locatelli, Hangyu Lin, Dwarak Talupuru, Bharat Venkitesh, David Cairuz, Bowen Yang, Tim Chung, Wei-Yin Ko, Sylvie~Shang Shi, Amir Shukayev, Sammie Bae, Aleksandra Piktus, Roman Castagné, Felipe Cruz-Salinas, Eddie Kim, Lucas Crawhall-Stein, Adrien Morisot, Sudip Roy, Phil Blunsom, Ivan Zhang, Aidan Gomez, Nick Frosst, Marzieh Fadaee, Beyza Ermis, Ahmet Üstün, and Sara Hooker. 2024.
\newblock \href {https://arxiv.org/abs/2412.04261} {Aya expanse: Combining research breakthroughs for a new multilingual frontier}.
\newblock \emph{ArXiv preprint}, abs/2412.04261.

\bibitem[{Devlin et~al.(2019)Devlin, Chang, Lee, and Toutanova}]{devlin-etal-2019-bert}
Jacob Devlin, Ming-Wei Chang, Kenton Lee, and Kristina Toutanova. 2019.
\newblock \href {https://doi.org/10.18653/v1/N19-1423} {{BERT}: Pre-training of deep bidirectional transformers for language understanding}.
\newblock In \emph{Proceedings of the 2019 Conference of the North {A}merican Chapter of the Association for Computational Linguistics: Human Language Technologies, Volume 1 (Long and Short Papers)}, pages 4171--4186, Minneapolis, Minnesota. Association for Computational Linguistics.

\bibitem[{Dong et~al.(2024)Dong, Ruan, Cai, Lai, Xu, Zhao, and Chen}]{dong2024xgrammar}
Yixin Dong, Charlie~F. Ruan, Yaxing Cai, Ruihang Lai, Ziyi Xu, Yilong Zhao, and Tianqi Chen. 2024.
\newblock \href {https://arxiv.org/abs/2411.15100} {Xgrammar: Flexible and efficient structured generation engine for large language models}.
\newblock \emph{Preprint}, arXiv:2411.15100.

\bibitem[{Dubey et~al.(2024)Dubey, Jauhri, Pandey, Kadian, Al-Dahle, Letman, Mathur, Schelten, Yang, Fan et~al.}]{dubey2024llama}
Abhimanyu Dubey, Abhinav Jauhri, Abhinav Pandey, Abhishek Kadian, Ahmad Al-Dahle, Aiesha Letman, Akhil Mathur, Alan Schelten, Amy Yang, Angela Fan, et~al. 2024.
\newblock \href {https://arxiv.org/abs/2407.21783} {The llama 3 herd of models}.
\newblock \emph{ArXiv preprint}, abs/2407.21783.

\bibitem[{Ebner et~al.(2020)Ebner, Xia, Culkin, Rawlins, and Van~Durme}]{ebner-etal-2020-multi}
Seth Ebner, Patrick Xia, Ryan Culkin, Kyle Rawlins, and Benjamin Van~Durme. 2020.
\newblock \href {https://doi.org/10.18653/v1/2020.acl-main.718} {Multi-sentence argument linking}.
\newblock In \emph{Proceedings of the 58th Annual Meeting of the Association for Computational Linguistics}, pages 8057--8077, Online. Association for Computational Linguistics.

\bibitem[{Eirew et~al.(2022)Eirew, Caciularu, and Dagan}]{eirew-etal-2022-cross}
Alon Eirew, Avi Caciularu, and Ido Dagan. 2022.
\newblock \href {https://doi.org/10.18653/v1/2022.emnlp-main.58} {Cross-document event coreference search: Task, dataset and modeling}.
\newblock In \emph{Proceedings of the 2022 Conference on Empirical Methods in Natural Language Processing}, pages 900--913, Abu Dhabi, United Arab Emirates. Association for Computational Linguistics.

\bibitem[{Gantt et~al.(2024)Gantt, Behzad, An, Chen, White, Van~Durme, and Yarmohammadi}]{gantt-etal-2024-multimuc}
William Gantt, Shabnam Behzad, Hannah An, Yunmo Chen, Aaron White, Benjamin Van~Durme, and Mahsa Yarmohammadi. 2024.
\newblock \href {https://aclanthology.org/2024.eacl-long.21/} {{M}ulti{MUC}: Multilingual template filling on {MUC}-4}.
\newblock In \emph{Proceedings of the 18th Conference of the European Chapter of the Association for Computational Linguistics (Volume 1: Long Papers)}, pages 349--368, St. Julian{'}s, Malta. Association for Computational Linguistics.

\bibitem[{Gao et~al.(2024)Gao, Li, Meng, Li, Zhou, Li, Teng, and Ji}]{gao-etal-2024-enhancing}
Qiang Gao, Bobo Li, Zixiang Meng, Yunlong Li, Jun Zhou, Fei Li, Chong Teng, and Donghong Ji. 2024.
\newblock \href {https://aclanthology.org/2024.lrec-main.523/} {Enhancing cross-document event coreference resolution by discourse structure and semantic information}.
\newblock In \emph{Proceedings of the 2024 Joint International Conference on Computational Linguistics, Language Resources and Evaluation (LREC-COLING 2024)}, pages 5907--5921, Torino, Italia. ELRA and ICCL.

\bibitem[{Grishman and Sundheim(1996)}]{grishman-sundheim-1996-message}
Ralph Grishman and Beth Sundheim. 1996.
\newblock \href {https://aclanthology.org/C96-1079/} {{M}essage {U}nderstanding {C}onference- 6: A brief history}.
\newblock In \emph{{COLING} 1996 Volume 1: The 16th International Conference on Computational Linguistics}.

\bibitem[{Hill(2010)}]{hill-number-1973}
M.~O. Hill. 2010.
\newblock \href {https://arxiv.org/abs/10.2307} {Diversity and evenness: A unifying notation and its consequences}.
\newblock \emph{ArXiv preprint}, abs/10.2307.

\bibitem[{Hsu et~al.(2022)Hsu, Huang, Boschee, Miller, Natarajan, Chang, and Peng}]{hsu-etal-2022-degree}
I-Hung Hsu, Kuan-Hao Huang, Elizabeth Boschee, Scott Miller, Prem Natarajan, Kai-Wei Chang, and Nanyun Peng. 2022.
\newblock \href {https://doi.org/10.18653/v1/2022.naacl-main.138} {{DEGREE}: A data-efficient generation-based event extraction model}.
\newblock In \emph{Proceedings of the 2022 Conference of the North American Chapter of the Association for Computational Linguistics: Human Language Technologies}, pages 1890--1908, Seattle, United States. Association for Computational Linguistics.

\bibitem[{Hsu et~al.(2023)Hsu, Huang, Zhang, Cheng, Natarajan, Chang, and Peng}]{hsu-etal-2023-tagprime}
I-Hung Hsu, Kuan-Hao Huang, Shuning Zhang, Wenxin Cheng, Prem Natarajan, Kai-Wei Chang, and Nanyun Peng. 2023.
\newblock \href {https://doi.org/10.18653/v1/2023.acl-long.723} {{TAGPRIME}: A unified framework for relational structure extraction}.
\newblock In \emph{Proceedings of the 61st Annual Meeting of the Association for Computational Linguistics (Volume 1: Long Papers)}, pages 12917--12932, Toronto, Canada. Association for Computational Linguistics.

\bibitem[{Huang et~al.(2022)Huang, Hsu, Natarajan, Chang, and Peng}]{huang-etal-2022-multilingual-generative}
Kuan-Hao Huang, I-Hung Hsu, Prem Natarajan, Kai-Wei Chang, and Nanyun Peng. 2022.
\newblock \href {https://doi.org/10.18653/v1/2022.acl-long.317} {Multilingual generative language models for zero-shot cross-lingual event argument extraction}.
\newblock In \emph{Proceedings of the 60th Annual Meeting of the Association for Computational Linguistics (Volume 1: Long Papers)}, pages 4633--4646, Dublin, Ireland. Association for Computational Linguistics.

\bibitem[{Huang et~al.(2024)Huang, Hsu, Parekh, Xie, Zhang, Natarajan, Chang, Peng, and Ji}]{huang-etal-2024-textee}
Kuan-Hao Huang, I-Hung Hsu, Tanmay Parekh, Zhiyu Xie, Zixuan Zhang, Prem Natarajan, Kai-Wei Chang, Nanyun Peng, and Heng Ji. 2024.
\newblock \href {https://doi.org/10.18653/v1/2024.findings-acl.760} {{T}ext{EE}: Benchmark, reevaluation, reflections, and future challenges in event extraction}.
\newblock In \emph{Findings of the Association for Computational Linguistics: ACL 2024}, pages 12804--12825, Bangkok, Thailand. Association for Computational Linguistics.

\bibitem[{Ilievski et~al.(2018)Ilievski, Vossen, and Schlobach}]{ilievski-etal-2018-systematic}
Filip Ilievski, Piek Vossen, and Stefan Schlobach. 2018.
\newblock \href {https://aclanthology.org/C18-1056/} {Systematic study of long tail phenomena in entity linking}.
\newblock In \emph{Proceedings of the 27th International Conference on Computational Linguistics}, pages 664--674, Santa Fe, New Mexico, USA. Association for Computational Linguistics.

\bibitem[{Jiang et~al.(2023)Jiang, Sablayrolles, Mensch, Bamford, Chaplot, de~las Casas, Bressand, Lengyel, Lample, Saulnier, Lavaud, Lachaux, Stock, Scao, Lavril, Wang, Lacroix, and Sayed}]{jiang2023mistral}
Albert~Q. Jiang, Alexandre Sablayrolles, Arthur Mensch, Chris Bamford, Devendra~Singh Chaplot, Diego de~las Casas, Florian Bressand, Gianna Lengyel, Guillaume Lample, Lucile Saulnier, Lélio~Renard Lavaud, Marie-Anne Lachaux, Pierre Stock, Teven~Le Scao, Thibaut Lavril, Thomas Wang, Timothée Lacroix, and William~El Sayed. 2023.
\newblock \href {https://arxiv.org/abs/2310.06825} {Mistral 7b}.
\newblock \emph{Preprint}, arXiv:2310.06825.

\bibitem[{Lewis et~al.(2020)Lewis, Liu, Goyal, Ghazvininejad, Mohamed, Levy, Stoyanov, and Zettlemoyer}]{lewis-etal-2020-bart}
Mike Lewis, Yinhan Liu, Naman Goyal, Marjan Ghazvininejad, Abdelrahman Mohamed, Omer Levy, Veselin Stoyanov, and Luke Zettlemoyer. 2020.
\newblock \href {https://doi.org/10.18653/v1/2020.acl-main.703} {{BART}: Denoising sequence-to-sequence pre-training for natural language generation, translation, and comprehension}.
\newblock In \emph{Proceedings of the 58th Annual Meeting of the Association for Computational Linguistics}, pages 7871--7880, Online. Association for Computational Linguistics.

\bibitem[{Li et~al.(2020{\natexlab{a}})Li, Peng, Chen, Wang, Pan, Lyu, and Zhu}]{li-etal-2020-event}
Fayuan Li, Weihua Peng, Yuguang Chen, Quan Wang, Lu~Pan, Yajuan Lyu, and Yong Zhu. 2020{\natexlab{a}}.
\newblock \href {https://doi.org/10.18653/v1/2020.findings-emnlp.73} {Event extraction as multi-turn question answering}.
\newblock In \emph{Findings of the Association for Computational Linguistics: EMNLP 2020}, pages 829--838, Online. Association for Computational Linguistics.

\bibitem[{Li et~al.(2021{\natexlab{a}})Li, Li, Wang, Huang, Cho, Ji, Han, and Voss}]{li2021future}
Manling Li, Sha Li, Zhenhailong Wang, Lifu Huang, Kyunghyun Cho, Heng Ji, Jiawei Han, and Clare Voss. 2021{\natexlab{a}}.
\newblock The future is not one-dimensional: Complex event schema induction by graph modeling for event prediction.
\newblock In \emph{Proceedings of the 2021 Conference on Empirical Methods in Natural Language Processing}, pages 5203--5215.

\bibitem[{Li et~al.(2019)Li, Lin, Hoover, Whitehead, Voss, Dehghani, and Ji}]{li-etal-2019-multilingual}
Manling Li, Ying Lin, Joseph Hoover, Spencer Whitehead, Clare Voss, Morteza Dehghani, and Heng Ji. 2019.
\newblock \href {https://doi.org/10.18653/v1/N19-4019} {Multilingual entity, relation, event and human value extraction}.
\newblock In \emph{Proceedings of the 2019 Conference of the North {A}merican Chapter of the Association for Computational Linguistics (Demonstrations)}, pages 110--115, Minneapolis, Minnesota. Association for Computational Linguistics.

\bibitem[{Li et~al.(2020{\natexlab{b}})Li, Zareian, Lin, Pan, Whitehead, Chen, Wu, Ji, Chang, Voss, Napierski, and Freedman}]{li-etal-2020-gaia}
Manling Li, Alireza Zareian, Ying Lin, Xiaoman Pan, Spencer Whitehead, Brian Chen, Bo~Wu, Heng Ji, Shih-Fu Chang, Clare Voss, Daniel Napierski, and Marjorie Freedman. 2020{\natexlab{b}}.
\newblock \href {https://doi.org/10.18653/v1/2020.acl-demos.11} {{GAIA}: A fine-grained multimedia knowledge extraction system}.
\newblock In \emph{Proceedings of the 58th Annual Meeting of the Association for Computational Linguistics: System Demonstrations}, pages 77--86, Online. Association for Computational Linguistics.

\bibitem[{Li et~al.(2020{\natexlab{c}})Li, Zeng, Lin, Cho, Ji, May, Chambers, and Voss}]{li-etal-2020-connecting}
Manling Li, Qi~Zeng, Ying Lin, Kyunghyun Cho, Heng Ji, Jonathan May, Nathanael Chambers, and Clare Voss. 2020{\natexlab{c}}.
\newblock \href {https://doi.org/10.18653/v1/2020.emnlp-main.50} {Connecting the dots: Event graph schema induction with path language modeling}.
\newblock In \emph{Proceedings of the 2020 Conference on Empirical Methods in Natural Language Processing (EMNLP)}, pages 684--695, Online. Association for Computational Linguistics.

\bibitem[{Li et~al.(2021{\natexlab{b}})Li, Ji, and Han}]{li-etal-2021-document}
Sha Li, Heng Ji, and Jiawei Han. 2021{\natexlab{b}}.
\newblock \href {https://doi.org/10.18653/v1/2021.naacl-main.69} {Document-level event argument extraction by conditional generation}.
\newblock In \emph{Proceedings of the 2021 Conference of the North American Chapter of the Association for Computational Linguistics: Human Language Technologies}, pages 894--908, Online. Association for Computational Linguistics.

\bibitem[{Li et~al.(2023)Li, Zhan, Conger, Palmer, Ji, and Han}]{li-etal-2023-glen}
Sha Li, Qiusi Zhan, Kathryn Conger, Martha Palmer, Heng Ji, and Jiawei Han. 2023.
\newblock \href {https://doi.org/10.18653/v1/2023.emnlp-main.170} {{GLEN}: General-purpose event detection for thousands of types}.
\newblock In \emph{Proceedings of the 2023 Conference on Empirical Methods in Natural Language Processing}, pages 2823--2838, Singapore. Association for Computational Linguistics.

\bibitem[{Liu et~al.(2020{\natexlab{a}})Liu, Chen, Liu, Bi, and Liu}]{liu-etal-2020-event}
Jian Liu, Yubo Chen, Kang Liu, Wei Bi, and Xiaojiang Liu. 2020{\natexlab{a}}.
\newblock \href {https://doi.org/10.18653/v1/2020.emnlp-main.128} {Event extraction as machine reading comprehension}.
\newblock In \emph{Proceedings of the 2020 Conference on Empirical Methods in Natural Language Processing (EMNLP)}, pages 1641--1651, Online. Association for Computational Linguistics.

\bibitem[{Liu et~al.(2024{\natexlab{a}})Liu, Tong, Peng, Hou, Li, and Xu}]{liu-etal-2024-docee}
Minghui Liu, MeiHan Tong, Yangda Peng, Lei Hou, Juanzi Li, and Bin Xu. 2024{\natexlab{a}}.
\newblock \href {https://doi.org/10.18653/v1/2024.findings-emnlp.35} {{D}oc{EE}-zh: A fine-grained benchmark for {C}hinese document-level event extraction}.
\newblock In \emph{Findings of the Association for Computational Linguistics: EMNLP 2024}, pages 637--649, Miami, Florida, USA. Association for Computational Linguistics.

\bibitem[{Liu et~al.(2024{\natexlab{b}})Liu, Semnani, Triedman, Xu, Zhao, and Lam}]{liu-etal-2024-spinach}
Shicheng Liu, Sina Semnani, Harold Triedman, Jialiang Xu, Isaac~Dan Zhao, and Monica Lam. 2024{\natexlab{b}}.
\newblock \href {https://doi.org/10.18653/v1/2024.findings-emnlp.938} {{SPINACH}: {SPARQL}-based information navigation for challenging real-world questions}.
\newblock In \emph{Findings of the Association for Computational Linguistics: EMNLP 2024}, pages 15977--16001, Miami, Florida, USA. Association for Computational Linguistics.

\bibitem[{Liu et~al.(2024{\natexlab{c}})Liu, Liu, Zhang, Wang, Liu, and Chen}]{liu-etal-2024-onenet}
Xukai Liu, Ye~Liu, Kai Zhang, Kehang Wang, Qi~Liu, and Enhong Chen. 2024{\natexlab{c}}.
\newblock \href {https://doi.org/10.18653/v1/2024.emnlp-main.756} {{O}ne{N}et: A fine-tuning free framework for few-shot entity linking via large language model prompting}.
\newblock In \emph{Proceedings of the 2024 Conference on Empirical Methods in Natural Language Processing}, pages 13634--13651, Miami, Florida, USA. Association for Computational Linguistics.

\bibitem[{Liu et~al.(2020{\natexlab{b}})Liu, Gu, Goyal, Li, Edunov, Ghazvininejad, Lewis, and Zettlemoyer}]{liu-etal-2020-multilingual-denoising}
Yinhan Liu, Jiatao Gu, Naman Goyal, Xian Li, Sergey Edunov, Marjan Ghazvininejad, Mike Lewis, and Luke Zettlemoyer. 2020{\natexlab{b}}.
\newblock \href {https://doi.org/10.1162/tacl_a_00343} {Multilingual denoising pre-training for neural machine translation}.
\newblock \emph{Transactions of the Association for Computational Linguistics}, 8:726--742.

\bibitem[{Logeswaran et~al.(2019{\natexlab{a}})Logeswaran, Chang, Lee, Toutanova, Devlin, and Lee}]{logeswaran2019zero}
Lajanugen Logeswaran, Ming-Wei Chang, Kenton Lee, Kristina Toutanova, Jacob Devlin, and Honglak Lee. 2019{\natexlab{a}}.
\newblock \href {https://doi.org/10.18653/v1/P19-1335} {Zero-shot entity linking by reading entity descriptions}.
\newblock In \emph{Proceedings of the 57th Annual Meeting of the Association for Computational Linguistics}, pages 3449--3460, Florence, Italy. Association for Computational Linguistics.

\bibitem[{Logeswaran et~al.(2019{\natexlab{b}})Logeswaran, Chang, Lee, Toutanova, Devlin, and Lee}]{logeswaran-etal-2019-zero}
Lajanugen Logeswaran, Ming-Wei Chang, Kenton Lee, Kristina Toutanova, Jacob Devlin, and Honglak Lee. 2019{\natexlab{b}}.
\newblock \href {https://doi.org/10.18653/v1/P19-1335} {Zero-shot entity linking by reading entity descriptions}.
\newblock In \emph{Proceedings of the 57th Annual Meeting of the Association for Computational Linguistics}, pages 3449--3460, Florence, Italy. Association for Computational Linguistics.

\bibitem[{Loshchilov and Hutter(2019)}]{loshchilov2017decoupled}
Ilya Loshchilov and Frank Hutter. 2019.
\newblock \href {https://openreview.net/forum?id=Bkg6RiCqY7} {Decoupled weight decay regularization}.
\newblock In \emph{7th International Conference on Learning Representations, {ICLR} 2019, New Orleans, LA, USA, May 6-9, 2019}. OpenReview.net.

\bibitem[{Lu et~al.(2023)Lu, Ran, Tetreault, and Jaimes}]{lu-etal-2023-event}
Di~Lu, Shihao Ran, Joel Tetreault, and Alejandro Jaimes. 2023.
\newblock \href {https://doi.org/10.18653/v1/2023.acl-short.143} {Event extraction as question generation and answering}.
\newblock In \emph{Proceedings of the 61st Annual Meeting of the Association for Computational Linguistics (Volume 2: Short Papers)}, pages 1666--1688, Toronto, Canada. Association for Computational Linguistics.

\bibitem[{Lu et~al.(2024)Lu, Mao, Lan, Xu, and Huang}]{lu2024exact}
Yi-Fan Lu, Xian-Ling Mao, Tian Lan, Chen Xu, and Heyan Huang. 2024.
\newblock \href {https://arxiv.org/abs/2410.09418} {Beyond exact match: Semantically reassessing event extraction by large language models}.
\newblock \emph{Preprint}, arXiv:2410.09418.

\bibitem[{Maheshwari et~al.(2019)Maheshwari, Patel, Rathod, Kumar, Ramakrishnan, and Bhattacharyya}]{maheshwari2019tale}
Ayush Maheshwari, Hrishikesh Patel, Nandan Rathod, Ritesh Kumar, Ganesh Ramakrishnan, and Pushpak Bhattacharyya. 2019.
\newblock \href {https://arxiv.org/abs/1908.07018} {Tale of tails using rule augmented sequence labeling for event extraction}.
\newblock \emph{Preprint}, arXiv:1908.07018.

\bibitem[{Mallen et~al.(2023)Mallen, Asai, Zhong, Das, Khashabi, and Hajishirzi}]{mallen-etal-2023-trust}
Alex Mallen, Akari Asai, Victor Zhong, Rajarshi Das, Daniel Khashabi, and Hannaneh Hajishirzi. 2023.
\newblock \href {https://doi.org/10.18653/v1/2023.acl-long.546} {When not to trust language models: Investigating effectiveness of parametric and non-parametric memories}.
\newblock In \emph{Proceedings of the 61st Annual Meeting of the Association for Computational Linguistics (Volume 1: Long Papers)}, pages 9802--9822, Toronto, Canada. Association for Computational Linguistics.

\bibitem[{Manning et~al.(2008)Manning, Raghavan, and Sch\"{u}tze}]{intro-to-ir-2008}
Christopher~D. Manning, Prabhakar Raghavan, and Hinrich Sch\"{u}tze. 2008.
\newblock \emph{Introduction to Information Retrieval}.
\newblock Cambridge University Press, USA.

\bibitem[{Milne and Witten(2008)}]{learningtolink}
David Milne and Ian~H. Witten. 2008.
\newblock \href {https://doi.org/10.1145/1458082.1458150} {Learning to link with wikipedia}.
\newblock In \emph{Proceedings of the 17th ACM Conference on Information and Knowledge Management}, CIKM '08, page 509–518, New York, NY, USA. Association for Computing Machinery.

\bibitem[{Mohammad et~al.(2016)Mohammad, Kiritchenko, Sobhani, Zhu, and Cherry}]{mohammad-etal-2016-semeval}
Saif Mohammad, Svetlana Kiritchenko, Parinaz Sobhani, Xiaodan Zhu, and Colin Cherry. 2016.
\newblock \href {https://doi.org/10.18653/v1/S16-1003} {{S}em{E}val-2016 task 6: Detecting stance in tweets}.
\newblock In \emph{Proceedings of the 10th International Workshop on Semantic Evaluation ({S}em{E}val-2016)}, pages 31--41, San Diego, California. Association for Computational Linguistics.

\bibitem[{Nguyen et~al.(2024)Nguyen, Tran, Luu, Nguyen, and Nguyen}]{nguyen-etal-2024-bkee}
Thi-Nhung Nguyen, Bang~Tien Tran, Trong-Nghia Luu, Thien~Huu Nguyen, and Kiem-Hieu Nguyen. 2024.
\newblock \href {https://aclanthology.org/2024.lrec-main.217/} {{BKEE}: Pioneering event extraction in the {V}ietnamese language}.
\newblock In \emph{Proceedings of the 2024 Joint International Conference on Computational Linguistics, Language Resources and Evaluation (LREC-COLING 2024)}, pages 2421--2427, Torino, Italia. ELRA and ICCL.

\bibitem[{{OpenStreetMap contributors}(2017)}]{OpenStreetMap}
{OpenStreetMap contributors}. 2017.
\newblock {Planet dump retrieved from https://planet.osm.org }.
\newblock https://www.openstreetmap.org/copyright.

\bibitem[{Paolini et~al.(2021)Paolini, Athiwaratkun, Krone, Ma, Achille, Anubhai, dos Santos, Xiang, and Soatto}]{tanl-2021}
Giovanni Paolini, Ben Athiwaratkun, Jason Krone, Jie Ma, Alessandro Achille, Rishita Anubhai, Cicero~Nogueira dos Santos, Bing Xiang, and Stefano Soatto. 2021.
\newblock \href {https://arxiv.org/abs/2101.05779} {Structured prediction as translation between augmented natural languages}.
\newblock \emph{Preprint}, arXiv:2101.05779.

\bibitem[{Piantadosi(2014)}]{piantadosi2014zipf}
Steven~T Piantadosi. 2014.
\newblock Zipf’s word frequency law in natural language: A critical review and future directions.
\newblock \emph{Psychonomic bulletin \& review}, 21:1112--1130.

\bibitem[{Pouran Ben~Veyseh et~al.(2022)Pouran Ben~Veyseh, Ebrahimi, Dernoncourt, and Nguyen}]{pouran-ben-veyseh-etal-2022-mee}
Amir Pouran Ben~Veyseh, Javid Ebrahimi, Franck Dernoncourt, and Thien Nguyen. 2022.
\newblock \href {https://doi.org/10.18653/v1/2022.emnlp-main.652} {{MEE}: A novel multilingual event extraction dataset}.
\newblock In \emph{Proceedings of the 2022 Conference on Empirical Methods in Natural Language Processing}, pages 9603--9613, Abu Dhabi, United Arab Emirates. Association for Computational Linguistics.

\bibitem[{Prabhu et~al.(2019)Prabhu, Goel, Debnath, and Shrivastava}]{prabhu-etal-2019-incorporating}
Suhan Prabhu, Pranav Goel, Alok Debnath, and Manish Shrivastava. 2019.
\newblock \href {https://aclanthology.org/2019.icon-1.5/} {Incorporating sub-word level information in language invariant neural event detection}.
\newblock In \emph{Proceedings of the 16th International Conference on Natural Language Processing}, pages 36--44, International Institute of Information Technology, Hyderabad, India. NLP Association of India.

\bibitem[{Rabinovich et~al.(2017)Rabinovich, Stern, and Klein}]{rabinovich-etal-2017-abstract}
Maxim Rabinovich, Mitchell Stern, and Dan Klein. 2017.
\newblock \href {https://doi.org/10.18653/v1/P17-1105} {Abstract syntax networks for code generation and semantic parsing}.
\newblock In \emph{Proceedings of the 55th Annual Meeting of the Association for Computational Linguistics (Volume 1: Long Papers)}, pages 1139--1149, Vancouver, Canada. Association for Computational Linguistics.

\bibitem[{Raleigh et~al.(2010)Raleigh, Linke, Hegre, and Karlsen}]{raleigh2010introducing}
Clionadh Raleigh, Rew Linke, H{\aa}vard Hegre, and Joakim Karlsen. 2010.
\newblock Introducing acled: An armed conflict location and event dataset.
\newblock \emph{Journal of peace research}, 47(5):651--660.

\bibitem[{Rasley et~al.(2020)Rasley, Rajbhandari, Ruwase, and He}]{deepspeed2020}
Jeff Rasley, Samyam Rajbhandari, Olatunji Ruwase, and Yuxiong He. 2020.
\newblock \href {https://dl.acm.org/doi/10.1145/3394486.3406703} {Deepspeed: System optimizations enable training deep learning models with over 100 billion parameters}.
\newblock In \emph{{KDD} '20: The 26th {ACM} {SIGKDD} Conference on Knowledge Discovery and Data Mining, Virtual Event, CA, USA, August 23-27, 2020}, pages 3505--3506. {ACM}.

\bibitem[{Reddy et~al.(2023)Reddy, Fung, Zeng, Li, Wang, Sullivan, and Ji}]{reddy2023smartbook}
Revanth~Gangi Reddy, Yi~R Fung, Qi~Zeng, Manling Li, Ziqi Wang, Paul Sullivan, and Heng Ji. 2023.
\newblock \href {https://arxiv.org/abs/2303.14337} {Smartbook: Ai-assisted situation report generation}.
\newblock \emph{ArXiv preprint}, abs/2303.14337.

\bibitem[{Ren et~al.(2024)Ren, Cao, Li, Li, Ma, Fang, Guo, and Ma}]{ren-etal-2024-deie}
Yubing Ren, Yanan Cao, Hao Li, Yingjie Li, Zixuan~ZM Ma, Fang Fang, Ping Guo, and Wei Ma. 2024.
\newblock \href {https://aclanthology.org/2024.lrec-main.410/} {{DEIE}: Benchmarking document-level event information extraction with a large-scale {C}hinese news dataset}.
\newblock In \emph{Proceedings of the 2024 Joint International Conference on Computational Linguistics, Language Resources and Evaluation (LREC-COLING 2024)}, pages 4592--4604, Torino, Italia. ELRA and ICCL.

\bibitem[{Rozière et~al.(2023)Rozière, Gehring, Gloeckle, Sootla, Gat, Tan, Adi, Liu, Sauvestre, Remez, Rapin, Kozhevnikov, Evtimov, Bitton, Bhatt, Ferrer, Grattafiori, Xiong, Défossez, Copet, Azhar, Touvron, Martin, Usunier, Scialom, and Synnaeve}]{rozire2023code}
Baptiste Rozière, Jonas Gehring, Fabian Gloeckle, Sten Sootla, Itai Gat, Xiaoqing~Ellen Tan, Yossi Adi, Jingyu Liu, Romain Sauvestre, Tal Remez, Jérémy Rapin, Artyom Kozhevnikov, Ivan Evtimov, Joanna Bitton, Manish Bhatt, Cristian~Canton Ferrer, Aaron Grattafiori, Wenhan Xiong, Alexandre Défossez, Jade Copet, Faisal Azhar, Hugo Touvron, Louis Martin, Nicolas Usunier, Thomas Scialom, and Gabriel Synnaeve. 2023.
\newblock \href {https://arxiv.org/abs/2308.12950} {Code llama: Open foundation models for code}.
\newblock \emph{Preprint}, arXiv:2308.12950.

\bibitem[{Saetia et~al.(2024)Saetia, Thonglong, Amornchaiteera, Chalothorn, Taerungruang, and Buabthong}]{thai-ee}
Chanatip Saetia, Areeya Thonglong, Thanpitcha Amornchaiteera, Tawunrat Chalothorn, Supawat Taerungruang, and Pakpoom Buabthong. 2024.
\newblock \href {https://doi.org/10.3389/frai.2024.1361483} {Streamlining event extraction with a simplified annotation framework}.
\newblock \emph{Frontiers in Artificial Intelligence}, 7:1361483.
\newblock ECollection 2024.

\bibitem[{Sainz et~al.(2024)Sainz, Garc{\'\i}a-Ferrero, Agerri, de~Lacalle, Rigau, and Agirre}]{sainz2024gollie}
Oscar Sainz, Iker Garc{\'\i}a-Ferrero, Rodrigo Agerri, Oier~Lopez de~Lacalle, German Rigau, and Eneko Agirre. 2024.
\newblock \href {https://openreview.net/forum?id=Y3wpuxd7u9} {Go{LLIE}: Annotation guidelines improve zero-shot information-extraction}.
\newblock In \emph{The Twelfth International Conference on Learning Representations}.

\bibitem[{{Sam Jones}(2022)}]{acled2022}
{Sam Jones}. 2022.
\newblock \href {https://acleddata.com/2022/02/02/new-expansion-brings-acled-to-full-global-coverage/} {New expansion brings {ACLED} to full global coverage}.
\newblock Accessed: 2024-09-28.

\bibitem[{Shin and Van~Durme(2022)}]{shin-van-durme-2022-shot}
Richard Shin and Benjamin Van~Durme. 2022.
\newblock \href {https://doi.org/10.18653/v1/2022.naacl-main.396} {Few-shot semantic parsing with language models trained on code}.
\newblock In \emph{Proceedings of the 2022 Conference of the North American Chapter of the Association for Computational Linguistics: Human Language Technologies}, pages 5417--5425, Seattle, United States. Association for Computational Linguistics.

\bibitem[{Sundheim(1992)}]{sundheim-1992-overview}
Beth~M. Sundheim. 1992.
\newblock \href {https://aclanthology.org/M92-1001/} {Overview of the fourth {M}essage {U}nderstanding {E}valuation and {C}onference}.
\newblock In \emph{{F}ourth {M}essage {U}nderstanding {C}onference ({MUC}-4): Proceedings of a Conference Held in {M}c{L}ean, {V}irginia, {J}une 16-18, 1992}.

\bibitem[{Tong et~al.(2022)Tong, Xu, Wang, Han, Cao, Zhu, Chen, Hou, and Li}]{tong-etal-2022-docee}
MeiHan Tong, Bin Xu, Shuai Wang, Meihuan Han, Yixin Cao, Jiangqi Zhu, Siyu Chen, Lei Hou, and Juanzi Li. 2022.
\newblock \href {https://doi.org/10.18653/v1/2022.naacl-main.291} {{D}oc{EE}: A large-scale and fine-grained benchmark for document-level event extraction}.
\newblock In \emph{Proceedings of the 2022 Conference of the North American Chapter of the Association for Computational Linguistics: Human Language Technologies}, pages 3970--3982, Seattle, United States. Association for Computational Linguistics.

\bibitem[{Tunstall et~al.(2023)Tunstall, Beeching, Lambert, Rajani, Rasul, Belkada, Huang, von Werra, Fourrier, Habib, Sarrazin, Sanseviero, Rush, and Wolf}]{tunstall2023zephyr}
Lewis Tunstall, Edward Beeching, Nathan Lambert, Nazneen Rajani, Kashif Rasul, Younes Belkada, Shengyi Huang, Leandro von Werra, Clémentine Fourrier, Nathan Habib, Nathan Sarrazin, Omar Sanseviero, Alexander~M. Rush, and Thomas Wolf. 2023.
\newblock \href {https://arxiv.org/abs/2310.16944} {Zephyr: Direct distillation of lm alignment}.
\newblock \emph{Preprint}, arXiv:2310.16944.

\bibitem[{Uddin et~al.(2024)Uddin, George, Blanco, and Corman}]{uddin-etal-2024-generating}
Md~Nayem Uddin, Enfa George, Eduardo Blanco, and Steven Corman. 2024.
\newblock \href {https://doi.org/10.18653/v1/2024.naacl-long.312} {Generating uncontextualized and contextualized questions for document-level event argument extraction}.
\newblock In \emph{Proceedings of the 2024 Conference of the North American Chapter of the Association for Computational Linguistics: Human Language Technologies (Volume 1: Long Papers)}, pages 5612--5627, Mexico City, Mexico. Association for Computational Linguistics.

\bibitem[{Wadden et~al.(2019)Wadden, Wennberg, Luan, and Hajishirzi}]{wadden-etal-2019-entity}
David Wadden, Ulme Wennberg, Yi~Luan, and Hannaneh Hajishirzi. 2019.
\newblock \href {https://doi.org/10.18653/v1/D19-1585} {Entity, relation, and event extraction with contextualized span representations}.
\newblock In \emph{Proceedings of the 2019 Conference on Empirical Methods in Natural Language Processing and the 9th International Joint Conference on Natural Language Processing (EMNLP-IJCNLP)}, pages 5784--5789, Hong Kong, China. Association for Computational Linguistics.

\bibitem[{Walker et~al.(2006)Walker, Strassel, Medero, and Maeda}]{ace05}
Christopher Walker, Stephanie Strassel, Julie Medero, and Kazuaki Maeda. 2006.
\newblock \href {https://doi.org/10.35111/mwxc-vh88} {Ace 2005 multilingual training corpus ldc2006t06}.

\bibitem[{Wang et~al.(2020)Wang, Wang, Han, Jiang, Han, Liu, Li, Li, Lin, and Zhou}]{wang-etal-2020-maven}
Xiaozhi Wang, Ziqi Wang, Xu~Han, Wangyi Jiang, Rong Han, Zhiyuan Liu, Juanzi Li, Peng Li, Yankai Lin, and Jie Zhou. 2020.
\newblock \href {https://doi.org/10.18653/v1/2020.emnlp-main.129} {{MAVEN}: {A} {M}assive {G}eneral {D}omain {E}vent {D}etection {D}ataset}.
\newblock In \emph{Proceedings of the 2020 Conference on Empirical Methods in Natural Language Processing (EMNLP)}, pages 1652--1671, Online. Association for Computational Linguistics.

\bibitem[{Wang et~al.(2023)Wang, Li, and Ji}]{wang-etal-2023-code4struct}
Xingyao Wang, Sha Li, and Heng Ji. 2023.
\newblock \href {https://doi.org/10.18653/v1/2023.acl-long.202} {{C}ode4{S}truct: Code generation for few-shot event structure prediction}.
\newblock In \emph{Proceedings of the 61st Annual Meeting of the Association for Computational Linguistics (Volume 1: Long Papers)}, pages 3640--3663, Toronto, Canada. Association for Computational Linguistics.

\bibitem[{Wen et~al.(2021)Wen, Lin, Lai, Pan, Li, Lin, Zhou, Li, Wang, Zhang, Yu, Dong, Wang, Fung, Mishra, Lyu, Sur{\'i}s, Chen, Brown, Palmer, Callison-Burch, Vondrick, Han, Roth, Chang, and Ji}]{wen-etal-2021-resin}
Haoyang Wen, Ying Lin, Tuan Lai, Xiaoman Pan, Sha Li, Xudong Lin, Ben Zhou, Manling Li, Haoyu Wang, Hongming Zhang, Xiaodong Yu, Alexander Dong, Zhenhailong Wang, Yi~Fung, Piyush Mishra, Qing Lyu, D{\'i}dac Sur{\'i}s, Brian Chen, Susan~Windisch Brown, Martha Palmer, Chris Callison-Burch, Carl Vondrick, Jiawei Han, Dan Roth, Shih-Fu Chang, and Heng Ji. 2021.
\newblock \href {https://doi.org/10.18653/v1/2021.naacl-demos.16} {{RESIN}: A dockerized schema-guided cross-document cross-lingual cross-media information extraction and event tracking system}.
\newblock In \emph{Proceedings of the 2021 Conference of the North American Chapter of the Association for Computational Linguistics: Human Language Technologies: Demonstrations}, pages 133--143, Online. Association for Computational Linguistics.

\bibitem[{Willard and Louf(2023)}]{br2023efficient}
Brandon~T. Willard and Rémi Louf. 2023.
\newblock \href {https://arxiv.org/abs/2307.09702} {Efficient guided generation for large language models}.
\newblock \emph{Preprint}, arXiv:2307.09702.

\bibitem[{Wolf et~al.(2019)Wolf, Debut, Sanh, Chaumond, Delangue, Moi, Cistac, Rault, Louf, Funtowicz, Davison, Shleifer, von Platen, Ma, Jernite, Plu, Xu, Scao, Gugger, Drame, Lhoest, and Rush}]{wolf2019huggingfaces}
Thomas Wolf, Lysandre Debut, Victor Sanh, Julien Chaumond, Clement Delangue, Anthony Moi, Pierric Cistac, Tim Rault, Rémi Louf, Morgan Funtowicz, Joe Davison, Sam Shleifer, Patrick von Platen, Clara Ma, Yacine Jernite, Julien Plu, Canwen Xu, Teven~Le Scao, Sylvain Gugger, Mariama Drame, Quentin Lhoest, and Alexander~M. Rush. 2019.
\newblock \href {https://arxiv.org/abs/1910.03771} {Huggingface's transformers: State-of-the-art natural language processing}.
\newblock \emph{Preprint}, arXiv:1910.03771.

\bibitem[{Xiao et~al.(2022)Xiao, Liu, Shao, and Cao}]{xiao-etal-2022-retromae}
Shitao Xiao, Zheng Liu, Yingxia Shao, and Zhao Cao. 2022.
\newblock \href {https://doi.org/10.18653/v1/2022.emnlp-main.35} {{R}etro{MAE}: Pre-training retrieval-oriented language models via masked auto-encoder}.
\newblock In \emph{Proceedings of the 2022 Conference on Empirical Methods in Natural Language Processing}, pages 538--548, Abu Dhabi, United Arab Emirates. Association for Computational Linguistics.

\bibitem[{Xu et~al.(2023)Xu, Chen, Hu, and Zhang}]{xu2023read}
Zhenran Xu, Yulin Chen, Baotian Hu, and Min Zhang. 2023.
\newblock \href {https://arxiv.org/abs/2310.12450} {A read-and-select framework for zero-shot entity linking}.
\newblock \emph{ArXiv preprint}, abs/2310.12450.

\bibitem[{Xue et~al.(2021)Xue, Constant, Roberts, Kale, Al-Rfou, Siddhant, Barua, and Raffel}]{xue-etal-2021-mt5}
Linting Xue, Noah Constant, Adam Roberts, Mihir Kale, Rami Al-Rfou, Aditya Siddhant, Aditya Barua, and Colin Raffel. 2021.
\newblock \href {https://doi.org/10.18653/v1/2021.naacl-main.41} {m{T}5: A massively multilingual pre-trained text-to-text transformer}.
\newblock In \emph{Proceedings of the 2021 Conference of the North American Chapter of the Association for Computational Linguistics: Human Language Technologies}, pages 483--498, Online. Association for Computational Linguistics.

\bibitem[{Yang et~al.(2021)Yang, Sui, Chen, Liu, Zhao, and Wang}]{yang-etal-2021-document}
Hang Yang, Dianbo Sui, Yubo Chen, Kang Liu, Jun Zhao, and Taifeng Wang. 2021.
\newblock \href {https://doi.org/10.18653/v1/2021.acl-long.492} {Document-level event extraction via parallel prediction networks}.
\newblock In \emph{Proceedings of the 59th Annual Meeting of the Association for Computational Linguistics and the 11th International Joint Conference on Natural Language Processing (Volume 1: Long Papers)}, pages 6298--6308, Online. Association for Computational Linguistics.

\bibitem[{Zavarella et~al.(2014)Zavarella, K{\"u}{\c{c}}{\"u}k, Tanev, and H{\"u}rriyeto{\u{g}}lu}]{zavarella-etal-2014-event}
Vanni Zavarella, Dilek K{\"u}{\c{c}}{\"u}k, Hristo Tanev, and Ali H{\"u}rriyeto{\u{g}}lu. 2014.
\newblock \href {https://doi.org/10.3115/v1/E14-2017} {Event extraction for {B}alkan languages}.
\newblock In \emph{Proceedings of the Demonstrations at the 14th Conference of the {E}uropean Chapter of the Association for Computational Linguistics}, pages 65--68, Gothenburg, Sweden. Association for Computational Linguistics.

\bibitem[{Zhang et~al.(2024{\natexlab{a}})Zhang, Zhang, Long, Xie, Dai, Tang, Lin, Yang, Xie, Huang, Zhang, Li, and Zhang}]{zhang-etal-2024-mgte}
Xin Zhang, Yanzhao Zhang, Dingkun Long, Wen Xie, Ziqi Dai, Jialong Tang, Huan Lin, Baosong Yang, Pengjun Xie, Fei Huang, Meishan Zhang, Wenjie Li, and Min Zhang. 2024{\natexlab{a}}.
\newblock \href {https://doi.org/10.18653/v1/2024.emnlp-industry.103} {{mGTE}: Generalized long-context text representation and reranking models for multilingual text retrieval}.
\newblock In \emph{Proceedings of the 2024 Conference on Empirical Methods in Natural Language Processing: Industry Track}, pages 1393--1412, Miami, Florida, US. Association for Computational Linguistics.

\bibitem[{Zhang et~al.(2024{\natexlab{b}})Zhang, Zhang, Long, Xie, Dai, Tang, Lin, Yang, Xie, Huang, Zhang, Li, and Zhang}]{zhang2024mgte}
Xin Zhang, Yanzhao Zhang, Dingkun Long, Wen Xie, Ziqi Dai, Jialong Tang, Huan Lin, Baosong Yang, Pengjun Xie, Fei Huang, Meishan Zhang, Wenjie Li, and Min Zhang. 2024{\natexlab{b}}.
\newblock \href {https://arxiv.org/abs/2407.19669} {mgte: Generalized long-context text representation and reranking models for multilingual text retrieval}.
\newblock \emph{Preprint}, arXiv:2407.19669.

\bibitem[{Zhu et~al.(2024)Zhu, Xu, Zeng, Xiao, Wang, Ke, and Huang}]{zhu-etal-2024-cmnee}
Mengna Zhu, Zijie Xu, Kaisheng Zeng, Kaiming Xiao, Mao Wang, Wenjun Ke, and Hongbin Huang. 2024.
\newblock \href {https://aclanthology.org/2024.lrec-main.299/} {{CMNEE}:a large-scale document-level event extraction dataset based on open-source {C}hinese military news}.
\newblock In \emph{Proceedings of the 2024 Joint International Conference on Computational Linguistics, Language Resources and Evaluation (LREC-COLING 2024)}, pages 3367--3379, Torino, Italia. ELRA and ICCL.

\end{thebibliography}

\appendix
\clearpage

\section{Creation of \dataset}
\label{appendix:dataset-creation}
In this section, we describe the detailed steps involved in creating \dataset, including data cleaning and the reannotation of specific event arguments. The overall process combined domain expert spot-checks, iterative improvements by authors of this paper, and assistance from a large language model (GPT-4o) for straightforward yet labor-intensive tasks.

\subsection{Original ACLED Annotation Process}
The Armed Conflict Location and Event Data (ACLED) project~\citep{raleigh2010introducing}, first published in 2010, provides comprehensive annotations of civil wars, subnational and transnational violent events, political violence, and civil unrest across 243 countries and territories. The dataset covers events reported in approximately 100 languages and is updated in near real-time~\citep{acled2022, acled2023impact}.

ACLED annotations are produced by a team of around 200 domain experts and updated weekly. The data sources include news media, reports from international organizations, NGOs, security agencies, local partner organizations, and social media channels.

Annotations are conducted by researchers familiar with the specific region and language of the events they annotate. These researchers also provide an English-language ``summary and note,'' explaining the event, its context, and any uncertainties regarding its labeling. Annotators utilize a dedicated annotation tool that maintains an up-to-date list of entities and locations, and they regularly communicate with each other to resolve challenging annotation decisions. Finally, annotations undergo a rigorous multi-step review and quality assurance process~\citep{acled2020coding}, consisting of three distinct review stages:

\begin{enumerate}
    \item \textbf{Initial Review:} Annotations are first reviewed by another researcher familiar with the same region.
    \item \textbf{Regional Manager Review:} A region-specific research manager, familiar with some (but not necessarily all) languages of the region, conducts a second review, primarily relying on the provided English summaries and notes.
    \item \textbf{Centralized Review:} A central team performs a final review, again using the English summaries and notes, to ensure methodological consistency across different regions.
\end{enumerate}

In addition to these review rounds, the ACLED quality assurance team conducted a separate review of 1,265 randomly selected events against the ACLED codebook, reporting the following findings:

\begin{itemize}
    \item 5 events had incorrect event types.
    \item 32 events had missing entities.
    \item 12 events had inaccurate locations.
    \item 30 events exhibited other miscellaneous issues.
\end{itemize}
When analyzed at the level of individual data points (e.g., event type, entities, location, casualties), fewer than 1\% of researcher-coded labels contained errors.

\subsection{Converting ACLED to \dataset}
\paragraph{Data Filtering and Cleaning}

We obtained all ACLED events from January 2024 to January 2025 (13 months) totaling 344,116 events. Each event was associated with one or more URLs linking to relevant online sources. Upon analysis, we found that many social media posts included images (e.g., protest flyers), making text alone insufficient for accurate annotation. Consequently, we excluded all social media posts from our dataset.

We also removed the shortest and longest 1\% of documents. Very short documents, often sourced from local partner organizations, lacked sufficient context for accurate annotation, while very long documents were frequently concatenations of multiple news articles included erroneously.

We used GPT-4o to detect the language of each document. Additionally, we retrieved the full text from the provided URLs and cleaned the documents by removing advertisements and irrelevant content using LLM prompts.

\paragraph{Balancing the Dataset}
Since ACLED data reflects real-world distributions, the frequency of event types within each language heavily depends on the political stability of countries where the language is spoken. For instance, most events in English and Korean are categorized as ``Peaceful Protest,'' whereas Burmese events (from Myanmar) predominantly fall under ``Armed Clash.'' Such imbalance can negatively impact AI system performance. To mitigate this, we downsampled the most frequent event types within each language, resulting in a balanced dataset of 114,743 events. Furthermore, we restricted our dataset to languages with at least 500 events each, ensuring sufficient data for robust model evaluation.

\paragraph{Converting ACLED to a Document-Level Dataset}
As mentioned earlier, each ACLED event is associated with one or more documents. To facilitate document-level event extraction research, we converted the dataset to a one-event-per-document format by pairing each document with its corresponding event. However, a document might contain only partial information about an event, for example, mentioning only the ``attacker'' in a Mob Violence event, while the ``victim'' is described in another document).
To address this, we processed each event-document pair independently, ensuring that each event annotation contained only information explicitly mentioned in its paired document. After processing, we deduplicated event-document pairs by retaining only the most complete annotation for each event.

\paragraph{Location Reannotation}
For location reannotation, we leveraged the original ACLED location annotations to query the OpenStreetMap geographic database~\citep{OpenStreetMap}, retrieving the full hierarchical location structure above the neighborhood level. Starting from the lowest location level, we removed any location components not explicitly supported by the document, continuing upward until we identified a supported location. We retained this location and all higher-level locations. This final step was performed using a carefully designed LLM prompt. The authors conducted spot-checks on the final location annotations, confirming that 97\% were accurate according to the above criteria.
Below is an example illustrating the location annotation before and after our reannotation process:

\textbf{Example Event (Armed Clash):}  
\textit{``Balochistan Liberation Front Claims Responsibility for... the attack on the Pakistani army occupying \textbf{Mand} in a press release issued to the media. The spokesperson stated that the Sarmachars attacked the main camp of the enemy army in the Mand area of \textbf{Kech}. At nine o'clock last night, the Sarmachars launched an attack on the main camp of the occupying Pakistani army in Mand Soro with rockets, resulting in...''}

\begin{itemize}
    \item \textbf{Location Before Reannotation:} Bolan Mach, Kachhi, Balochistan, Pakistan
    \item \textbf{Location After Reannotation:} \textbf{Mand} Tehsil, \textbf{Kech} District, Balochistan, Pakistan
\end{itemize}

The authors carried out all annotation work for this paper, communicating with the original ACLED team to clarify their annotation processes and ensure compatibility with their codebook.

\paragraph{Schematization}
ACLED employs uniform event argument roles across all event types, resulting in some roles consistently remaining empty or overly generic. To address this, we defined distinct event argument roles tailored specifically to each event type. For example, we removed the ``fatalities'' argument from the ``Peaceful Protest'' event type and renamed ``actor 1'' to ``Abductor'' for the ``Abduction or Forced Disappearance'' event type. Additionally, we provided concise descriptions for each event type and each event argument, facilitating the development of zero-shot models.

Following recent trends in event extraction, we represented annotations using Python code. This approach has been shown to improve the performance of supervised~\citep{sainz2024gollie} and few-shot~\citep{wang-etal-2023-code4struct} models by aligning labels more closely with the code data on which many language models are pre-trained. Moreover, this representation enables the use of constrained decoding algorithms~\citep{rabinovich-etal-2017-abstract, br2023efficient}, effectively eliminating malformed outputs. The complete schema for \dataset is provided in Appendix~\ref{appendix:dataset-schema}.

\paragraph{Entity Descriptions}
As discussed earlier, we provide a short description for each entity in the database to facilitate entity linking. These descriptions are generated by GPT-4o using the news articles that are annotated to have involved each entity.

\section{Properties of \dataset}

\subsection{One Event per Document}
In \dataset, only the \emph{main} event described in each document is annotated, excluding background or historical events typically mentioned to provide context. The rationale behind this approach is that news articles generally focus on a single new event. For example, an article titled ``Anti-war protests spurred by recent missile strikes'' likely covers protests as the main event, while the missile strikes themselves would have been reported separately in earlier articles. Thus, annotating one event per document can achieve comprehensive event coverage while minimizing redundancy, making it suitable for real-world applications such as automated news monitoring systems.

This formulation also simplifies the task for event extraction systems, leading to higher accuracy compared to multi-event scenarios. For instance, \citet{yang-etal-2021-document} demonstrated that single-event formulations yield superior model accuracy compared to multi-event formulations.

\subsection{Entity Distribution}
\begin{figure}[htbp]
    \centering
    \includegraphics[width=0.95\linewidth]{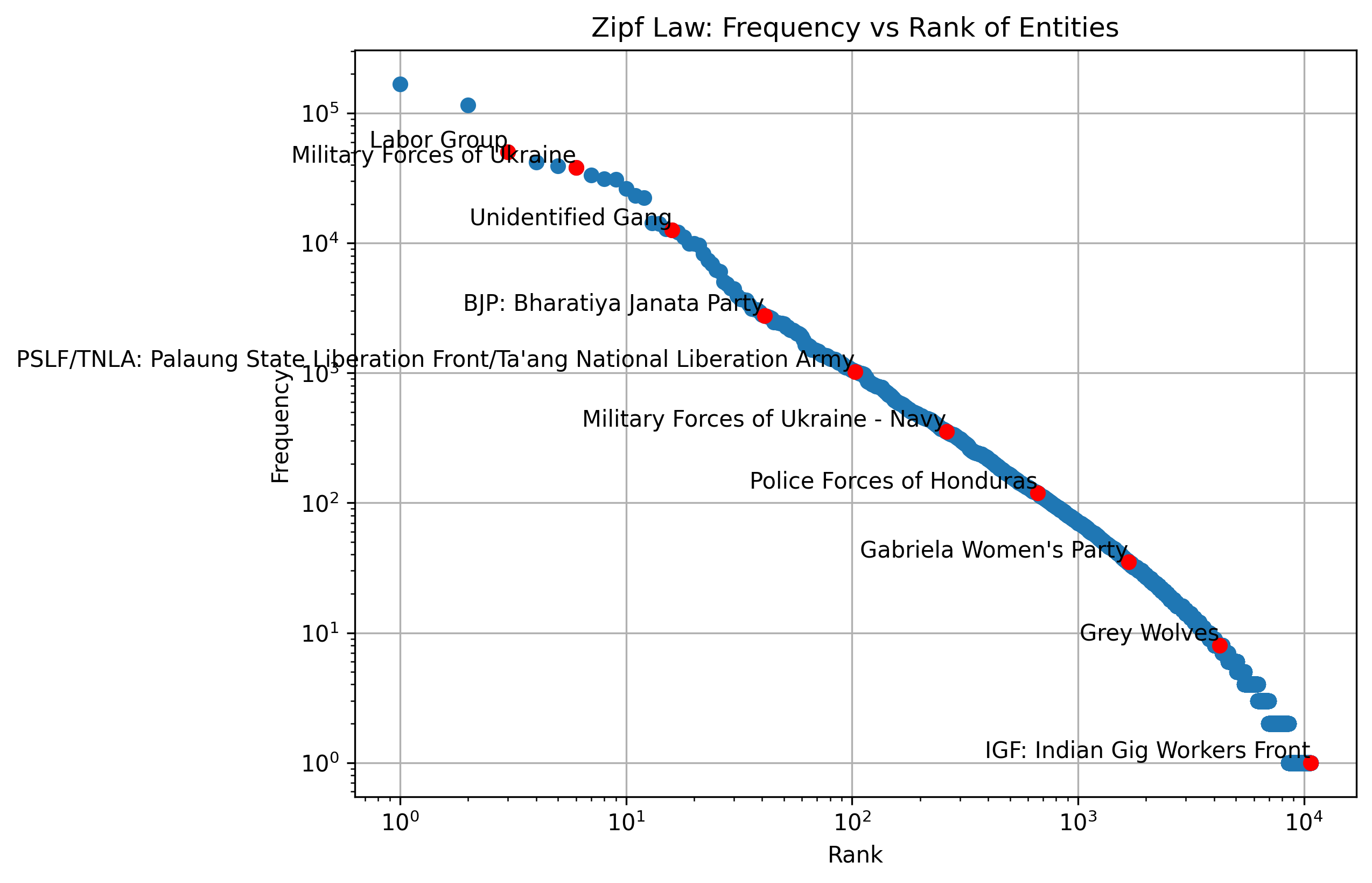}
    \caption{Entity frequency distribution in the original ACLED data, plotted on a log-log scale from the most frequent (rank 1) to the least frequent (rank 10,707). Examples are provided at every 10\% interval.}
    \label{figure:lemonade-entity-dist}
\end{figure}

Figure~\ref{figure:lemonade-entity-dist} illustrates the frequency distribution of entities in the original ACLED dataset. The entity distribution follows Zipf's law~\cite{piantadosi2014zipf}, a phenomenon previously studied in entity distributions by~\citet{ilievski-etal-2018-systematic}.

\subsection{\dataset Statistics}
\label{appendix:dataset-stats}

Table~\ref{table:dataset-stats} presents the number of events per language in each data split of \dataset. It also includes the full language names along with the abbreviations used throughout this paper.

\begin{table*}[htbp]
\begin{center}
\begin{tabular}{lccc}
\hline
\rowcolor{gray!10}
Language (language code) & Train & Dev & Test \\
\hline
English (en) & 4593 & 500 & 500 \\
\rowcolor{gray!10}
Spanish (es) & 1528 & 500 & 500 \\
Arabic (ar) & 3171 & 500 & 500 \\
\rowcolor{gray!10}
French (fr) & 805 & 500 & 500 \\
Italian (it) & 773 & 500 & 500 \\
\rowcolor{gray!10}
Russian (ru) & 482 & 500 & 500 \\
German (de) & 1422 & 500 & 500 \\
\rowcolor{gray!10}
Turkish (tr) & 925 & 500 & 500 \\
Burmese (my) & 932 & 500 & 500 \\
\rowcolor{gray!10}
Indonesian (id) & 754 & 500 & 500 \\
Ukrainian (uk) & 1157 & 500 & 500 \\
\rowcolor{gray!10}
Korean (ko) & 1167 & 500 & 500 \\
Portuguese (pt) & 1759 & 500 & 500 \\
\rowcolor{gray!10}
Dutch (nl) & 256 & 284 & 284 \\
Somali (so) & 251 & 358 & 358 \\
\rowcolor{gray!10}
Nepali (ne) & 389 & 439 & 439 \\
Chinese (zh) & 332 & 500 & 500 \\
\rowcolor{gray!10}
Persian/Farsi (fa) & 368 & 500 & 500 \\
Hebrew (he) & 177 & 332 & 332 \\
\rowcolor{gray!10}
Japanese (ja) & 175 & 272 & 272 \\
\hline
Total & 21,416 & 9,185 & 9,185 \\
\hline
\end{tabular}
\end{center}

\caption{\dataset statistics per language and split.}

\label{table:dataset-stats}
\end{table*}

Table~\ref{table:dataset-diversity} shows the Hill number~\cite{hill-number-1973}, or the effective number of event types, in \dataset. The Hill number is a diversity metric originating from ecology. We calculate it using ($q = 1$), which corresponds to the exponential of the Shannon entropy computed with natural logarithms.

\begin{table*}[htbp]
\begin{center}
\begin{tabular}{lcccc}
\hline
\rowcolor{gray!10}
Language (language code) & Train & Dev & Test & Total \\
\hline
English (en) & 11.1 & 11 & 10.8 & 11.2 \\
\rowcolor{gray!10}
Spanish (es) & 8 & 9 & 8.2 & 8.3 \\
Arabic (ar) & 12.5 &13.5 & 14 & 13  \\
\rowcolor{gray!10}
French (fr) & 10.2 &11.3 & 10.8 & 10.9 \\
Italian (it) & 1.4 & 1.5 &1.6 &1.5 \\
\rowcolor{gray!10}
Russian (ru) & 7.1 & 8.5 &8.8 &8.5 \\
German (de) & 1.3 & 1.8 &2 &1.5 \\
\rowcolor{gray!10}
Turkish (tr) & 3.7 & 4 &3.7 &3.8 \\
Burmese (my) & 11.2 &11.6 & 11.5 & 11.7 \\
\rowcolor{gray!10}
Indonesian (id) & 2.3 & 1.9 &2 &2.1 \\
Ukrainian (uk) &4.6 & 4.4 &4.4 &4.5 \\
\rowcolor{gray!10}
Korean (ko) & 1.1 & 1.1 &1.2 &1.1 \\
Portuguese (pt) & 4.2 & 4.8 &4.7 &4.4 \\
\rowcolor{gray!10}
Dutch (nl) & 1.7 & 1.8 &2 &1.9 \\
Somali (so) & 7.5 & 6.7 &6.8 &7.2 \\
\rowcolor{gray!10}
Nepali (ne) & 2.4 & 2.5 &2.8 &2.6 \\
Chinese (zh) & 1.5 & 1.6 &1.5 &1.6 \\
\rowcolor{gray!10}
Persian/Farsi (fa) & 4 & 4 &3.5 &3.9 \\
Hebrew (he) & 5.3 & 5.2 &5.4 &5.6 \\
\rowcolor{gray!10}
Japanese (ja) & 1.1 & 1.1 &1.1 &1.1 \\
\hline
Total & 8.6 & 7.2 &7.1 &8 \\
\hline
\end{tabular}
\end{center}

\caption{Hill number (effective number of event types) calculated for each language in \dataset.}
\label{table:dataset-diversity}
\end{table*}

\begin{figure*}[ht]
    \centering
    
    \begin{subfigure}{0.3\linewidth}
        \centering
        \includegraphics[width=\linewidth]{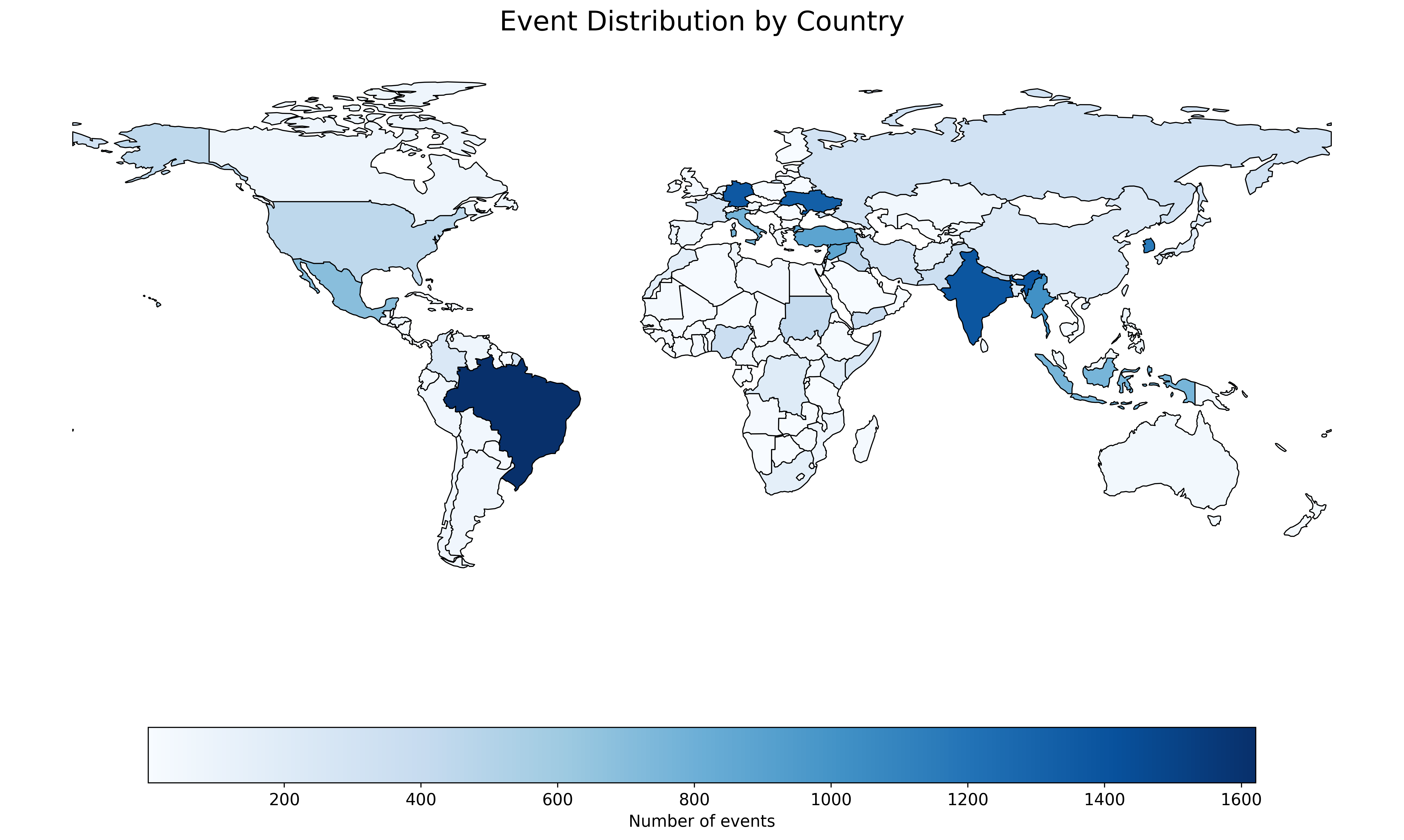}
        \caption{Train set}
    \end{subfigure}
    \hfill
    \begin{subfigure}{0.3\linewidth}
        \centering
        \includegraphics[width=\linewidth]{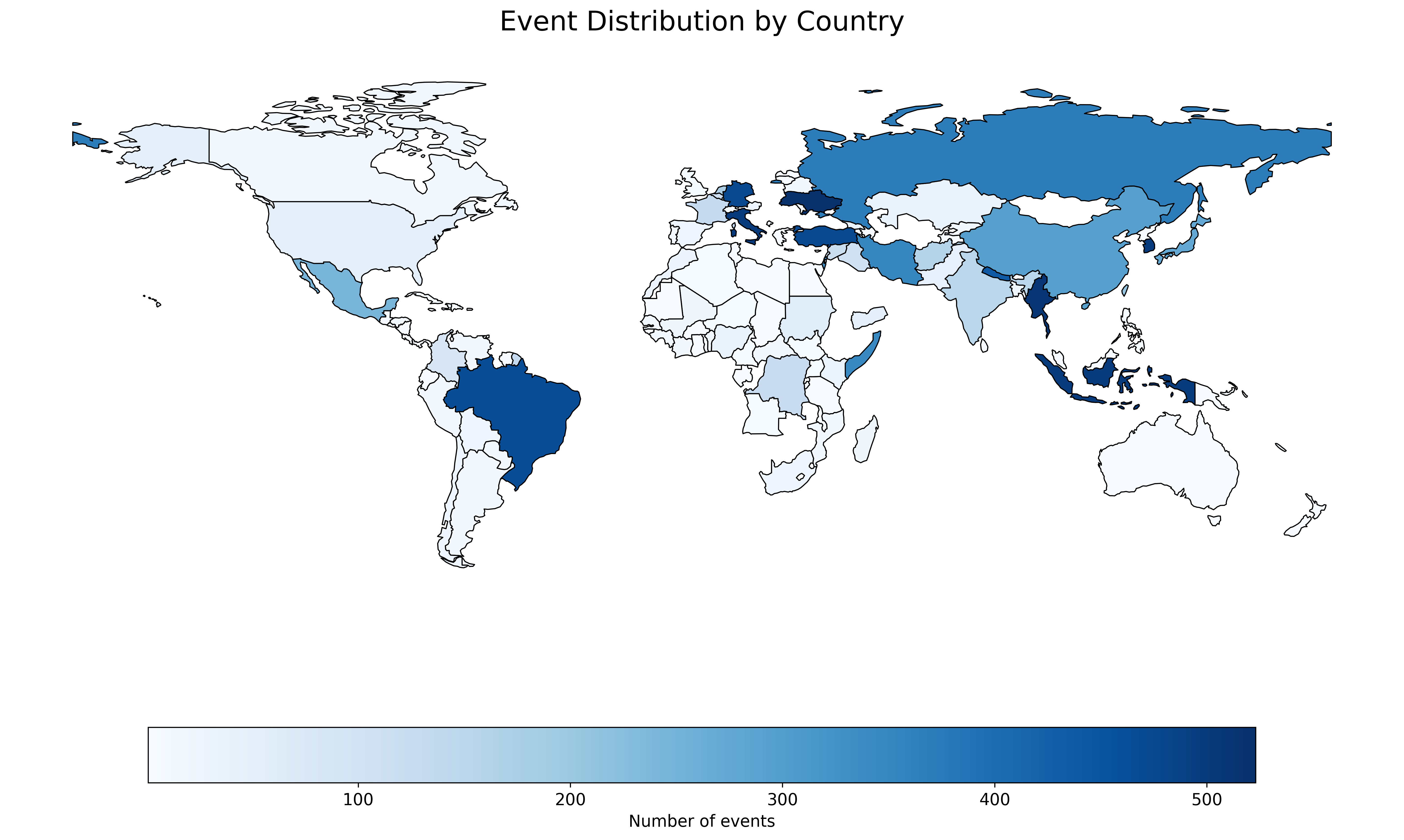}
        \caption{Validation set}
    \end{subfigure}
    \hfill
    \begin{subfigure}{0.3\linewidth}
        \centering
        \includegraphics[width=\linewidth]{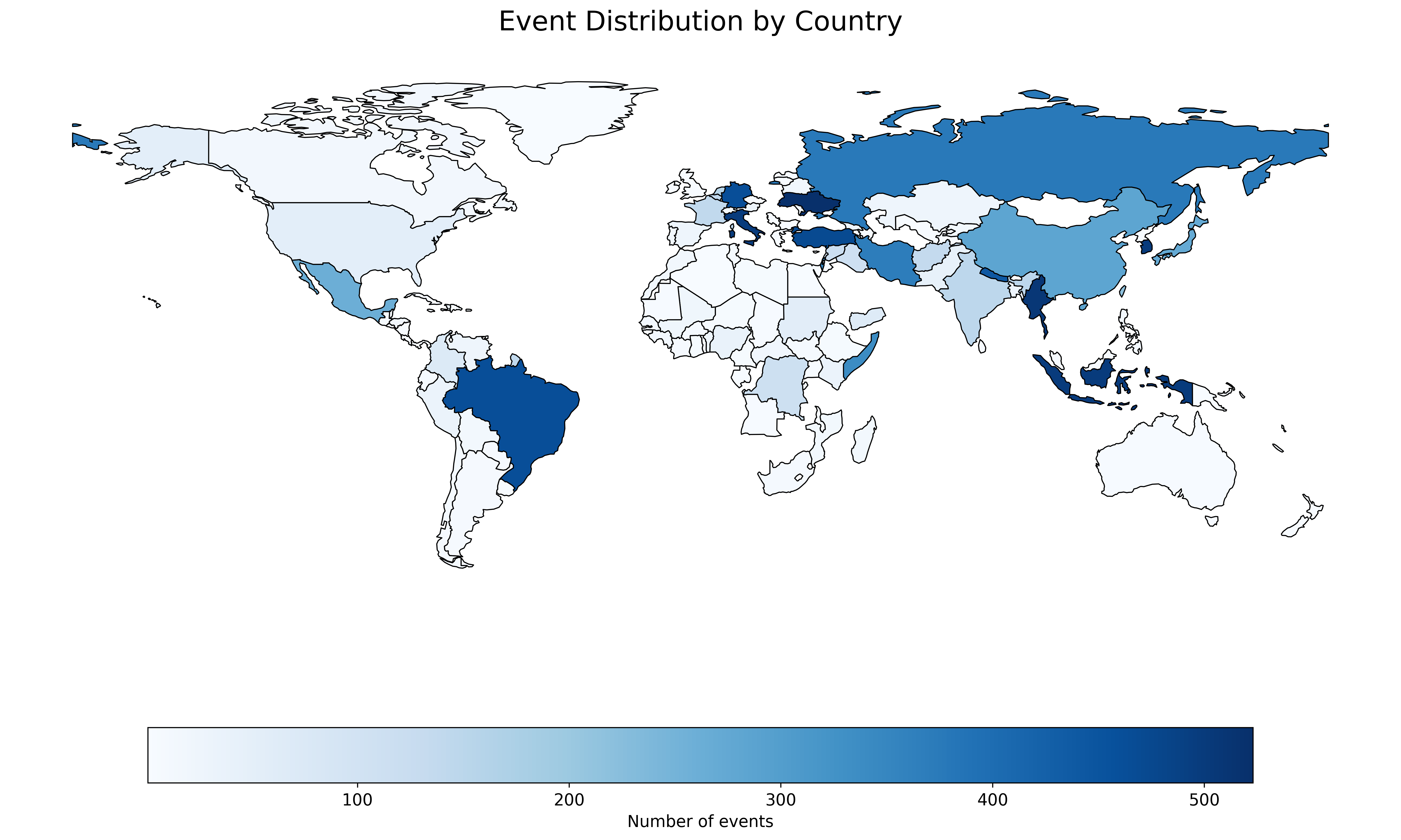}
        \caption{Test set}
    \end{subfigure}
    
    \caption{Distribution of event locations in the \dataset dataset. Although the dataset contains more specific location information, only country-level data are shown here. In addition to linguistic diversity, the dataset also exhibits substantial geographical diversity.}
    \label{figure:lemonade-country-stats}
\end{figure*}

Table~\ref{table:lemonade-type-stats} and Figure~\ref{figure:lemonade-country-stats} illustrate the distribution of event types and the geographical distribution of events at the country level in the \dataset dataset, respectively.

\begin{table}

\begin{center}
\begin{adjustbox}{center, width=0.95\linewidth}
    \begin{tabular}{>{\raggedright\arraybackslash}p{0.4\textwidth}|r}
        \hline
        \rowcolor{gray!20}
        \textbf{Event Type} & \textbf{Count} \\
        \hline
        GovernmentRegainsTerritory & 50 \\
        \rowcolor{gray!10}
        NonStateActorOvertakesTerritory & 130 \\
        ArmedClash & 3,473 \\
        \rowcolor{gray!10}
        ExcessiveForceAgainstProtestors & 49 \\
        ProtestWithIntervention & 1,001 \\
        \rowcolor{gray!10}
        PeacefulProtest & 18,481 \\
        ViolentDemonstration & 1,050 \\
        \rowcolor{gray!10}
        MobViolence & 1,398 \\
        AirOrDroneStrike & 2,074 \\
        \rowcolor{gray!10}
        SuicideBomb & 13 \\
        ShellingOrArtilleryOrMissileAttack & 2,226 \\
        \rowcolor{gray!10}
        RemoteExplosiveOrLandmineOrIED & 783 \\
        Grenade & 145 \\
        \rowcolor{gray!10}
        SexualViolence & 79 \\
        Attack & 3,418 \\
        \rowcolor{gray!10}
        AbductionOrForcedDisappearance & 674 \\
        Agreement & 87 \\
        \rowcolor{gray!10}
        Arrest & 910 \\
        ChangeToArmedGroup & 667 \\
        \rowcolor{gray!10}
        DisruptedWeaponsUse & 1,126 \\
        BaseEstablished & 16 \\
        \rowcolor{gray!10}
        LootingOrPropertyDestruction & 1,204 \\
        NonViolentTransferOfTerritory & 42 \\
        \rowcolor{gray!10}
        OtherStrategicDevelopment & 690 \\
        \hline
        Total & 39,786 \\
        \hline
    \end{tabular}
\end{adjustbox}
\end{center}

\caption{Distribution of event types across all splits of the \dataset dataset. Although the distribution is imbalanced, it accurately reflects real-world occurrences. For instance, peaceful protests constitute the majority of events.}
\label{table:lemonade-type-stats}

\end{table}

\begin{figure*}[ht]
    \centering
    \includegraphics[width=\linewidth]{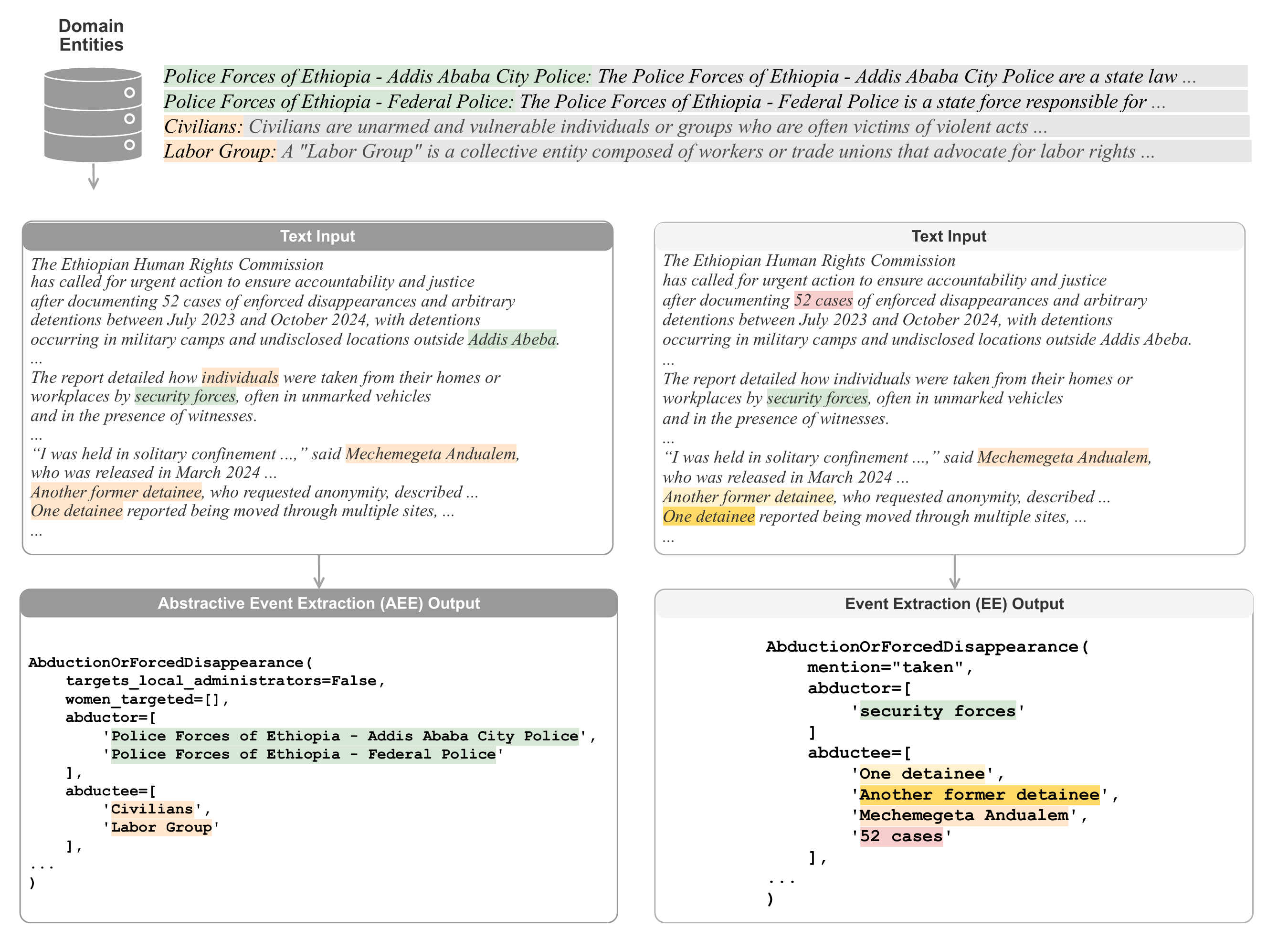}
    \caption{Another example from the \dataset dataset, shown with its abstractive event annotation. The input text and annotations have been summarized for clarity. A hypothetical extractive annotation for the same event is also provided for comparison. Note that identifying the abductors as ``Ethiopian Police Forces'' requires inference based on the event location (Addis Ababa) and contextual information.}    
    \label{figure:lemonade-example-2}
\end{figure*}

We show another example from \dataset with abstractive event annotation in Figure~\ref{figure:lemonade-example-2}

\section{Comparison of \dataset with Other Document-Level Event Datasets}
While most EE datasets primarily focus on English and Chinese~\citep{zhu-etal-2024-cmnee, ren-etal-2024-deie, ace05}, several datasets have been developed for other languages. These datasets vary significantly in annotation quality and often focus solely on the simpler event detection subtask. Notable examples include:

BKEE~\citep{nguyen-etal-2024-bkee} for Vietnamese,
InDEE-2019~\citep{maheshwari2019tale} for five Indic languages,
MEE~\citep{pouran-ben-veyseh-etal-2022-mee} for Portuguese, Spanish, Polish, Turkish, Hindi, Japanese, and Korean, \citet{zavarella-etal-2014-event} for Bulgarian, Romanian, and Turkish, \citet{Balali_2022} for Farsi,
\citet{li-etal-2019-multilingual} for Russian and Ukrainian, \citet{prabhu-etal-2019-incorporating} for English, Spanish, Italian, and French,
\citet{thai-ee} for Thai, \citet{dutch-ee} for Dutch, and \citet{ace05-pt} for Portuguese.

The \AEE definition unifies several traditionally separate subtasks. Traditional document-level EE (illustrated on the right side of the figure) typically involves the following sequential steps~\cite{huang-etal-2024-textee}:

\begin{enumerate}
    \item \textbf{Event Detection (ED):} Identifying trigger spans and their corresponding event types (e.g., a \texttt{MobViolence} event).
    \item \textbf{Event Argument Extraction (EAE):} Identifying argument spans and their roles for each event.
    \item \textbf{Entity Detection:} Finding text spans (mentions) that refer to entities.
    \item \textbf{Entity Coreference Resolution and/or Linking:} Resolving entity coreferences and linking them to corresponding entries in an entity database.
\end{enumerate}

It is worth noting that conventional EE systems often limit event arguments exclusively to entities~\cite{wadden-etal-2019-entity}.

\begin{table*}[htbp]
\centering
\small
\setlength\tabcolsep{5pt}
\renewcommand{\arraystretch}{1.2}
\begin{tabular}{@{}p{0.12\textwidth}%
                p{0.12\textwidth}%
                p{0.12\textwidth}%
                p{0.12\textwidth}%
                p{0.12\textwidth}%
                p{0.12\textwidth}%
                p{0.12\textwidth}@{}}
\toprule
 & \textbf{ACE05} & \textbf{DocEE} & \textbf{\dataset} & \textbf{WikiEvents} & \textbf{RAMS} & \textbf{Maven-Arg} \\
\midrule
Languages            & English, Chinese, Arabic, Portuguese & English, Chinese$^*$       & 20 Languages   & English   & English   & English \\

\midrule

Event Argument Types & string                              & string                   & string, numerical, categorical, boolean & string  & string    & string \\

\midrule

Entity Database      & Wikipedia$^*$                       & –                        & Domain Expert Curated & –     & –        & – \\
\midrule

Avg. Doc len         & 2,410                               & 2,052                  & 6,428  & 3,919   & 591      & 1,589 \\

\midrule

Annotators           & LDC Annotation Group                & Crowdworkers             & ACLED Domain Experts   & Graduate Students & Crowdworkers & Crowdworkers \\

\midrule

Num. Docs            & 1,635                               & 64,214$^*$               & 39,786 & 246     & 3,993    & 4,480 \\

\midrule

Num. Events          & 8,878                               & 64,214$^*$               & 39,786 & 3,951   & 9,124    & 98,591 \\

\midrule

Source               & News                                & Wikipedia                & News      & Wikipedia & News     & Wikipedia \\
\bottomrule
\end{tabular}
\caption{Comparison of \dataset with several other document-level event datasets. \dataset covers the largest number of languages and has the longest average document length. As an \AEE dataset, it also includes a broader range of event argument types. An en-dash (--) indicates that the dataset does not include an entity linking subtask.\\
\smallskip
\emph{Note:} $^*$ Includes aggregate statistics from multiple datasets. Entity linking was added to ACE05 by~\citet{bentivogli-etal-2010-extending}. \citet{ace05-pt} translated ACE05 into Portuguese using automatic translation. \citet{liu-etal-2024-docee} created another dataset in Chinese using the same ontology as DocEE.}
\label{table:dataset-comparison}
\end{table*}

\section{Examples of Entities from \dataset}
\label{appendix:entity-examples}
Tables~\ref{table:entity-examples-1}, ~\ref{table:entity-examples-2}, ~\ref{table:entity-examples-3}, ~\ref{table:entity-examples-4} and ~\ref{table:entity-examples-5} contain examples of \dataset entities and their description.

\begin{table*}[htbp]
\begin{center}
\begin{tabular}{p{5cm}p{10cm}}
\hline
\rowcolor{gray!10}
Entity Name & Entity Description \\
\hline
Women & Women are adult human females who can play diverse roles in society, ranging from caregivers and economic participants to political and social activists. They may be involved in a variety of social, economic, and political events, sometimes facing unique challenges such as discrimination or violence. Women's roles and their societal impact can be profound, as seen in their involvement in protests, advocacy for rights, and even in conflict situations where they may be victims or participants. Globally, women continue to strive for gender equality and empowerment, often organizing and mobilizing to address issues affecting their communities and themselves. \\
\rowcolor{gray!10}
Men & Men are adult human males who may be involved in a variety of societal roles and activities. As an entity, men can be participants in diverse events ranging from everyday community interactions to more extreme scenarios such as protests, violence against civilians, and riots. Men, as a group, can be both perpetrators and victims of violence, including sexual violence, as evidenced in various global incidents. Their involvement in these events can be influenced by cultural, social, and political contexts. This entity operates globally across all countries and societies. \\
Police Forces of the United States & The Police Forces of the United States are a collective entity composed of various local, state, and federal law enforcement agencies tasked with maintaining public order, enforcing laws, and ensuring public safety across the nation. These forces include municipal police departments, sheriff's offices, and specialized agencies such as the Federal Bureau of Investigation. They are recognized for their involvement in a wide range of activities, from managing public protests and investigating crimes to ensuring security during emergencies. While they play a crucial role in law enforcement, they have also faced scrutiny and legal challenges related to incidents of misconduct and use of force. Their operations are governed by state and federal laws, and they are accountable to governmental oversight bodies. They have been active since at least 1993 and continue to operate across the United States. \\
\rowcolor{gray!10}
Military Forces of Russia & The Military Forces of Russia are the armed forces of the Russian Federation, responsible for national defense and military operations both within and outside Russia. Established in 2000, they operate under the command of the President of Russia, who is the supreme commander-in-chief. These forces are composed of various branches, including the Ground Forces, Navy, and Air Force, along with strategic missile troops and airborne troops. They have been involved in international military operations, peacekeeping missions, and domestic security tasks. The Russian military is recognized for its significant involvement in various conflicts, including actions in Ukraine, Syria, and other regions, often collaborating with or opposing other nations' forces. The Military Forces of Russia are known for their extensive use of armored vehicles, aerial support, and advanced military technology. \\
\hline
\end{tabular}
\end{center}

\caption{20 entities in \dataset entity database -- Part 1}
\label{table:entity-examples-1}
\end{table*}

\begin{table*}[htbp]
\begin{center}
\begin{tabular}{p{5cm}p{10cm}}
\hline
\rowcolor{gray!10}
Entity Name & Entity Description \\
\hline
Protestors & Protestors are individuals or groups who actively participate in demonstrations to express opposition or demand action on specific issues. They can be found globally and may engage in peaceful protests or civil disobedience to draw attention to their causes. Protestors often advocate for political, social, or economic changes and can be associated with various movements, including anti-corruption, electoral fairness, and human rights. While they primarily aim for peaceful expression, their activities can sometimes lead to confrontations with authorities or opposing groups. Protestors play a crucial role in civil society by challenging perceived injustices and influencing public discourse and policy. \\
\rowcolor{gray!10}
Civilians & Civilians are unarmed, non-combatant individuals who are often vulnerable to violence and conflict, particularly in areas of political or social unrest. They can be affected by or involved in a wide range of events, including riots, protests, and violence, as seen in various global contexts. Civilians may participate in social movements, such as protests at educational institutions, or be subject to violence and negotiation processes in conflict zones, like settlements in Syria or mass violence in Colombia. Their involvement can manifest in active participation in civic actions or as victims of political and criminal violence, highlighting their diverse roles and the threats they face in unstable environments. \\
Rioters & Rioters are loosely assembled groups or mobs that engage in violent and disruptive behavior during demonstrations or spontaneously, often in response to perceived injustices or grievances. They may be civilians acting without inherent organization, and their actions typically involve confrontations with law enforcement or other entities. Rioters can be motivated by various social, political, or economic factors and are known to participate in actions such as vandalism, clashes, and other forms of violence. Their activities can occur in any country and are often part of broader social movements or tensions. \\
\rowcolor{gray!10}
Students & Students are individuals enrolled in educational institutions, ranging from primary schools to universities, and are often involved in various social and political activities. They can be a diverse and dynamic group that participates in protests, movements against discrimination, and other forms of activism, sometimes leading to confrontations with law enforcement or political opposition. Students can also be impacted by external conflicts, such as gang violence, which may directly affect their safety and educational environment. While they are typically associated with learning and academic pursuits, students have historically played significant roles in advocating for change and challenging established systems, sometimes at the risk of becoming involved in violent or controversial situations. \\

\hline
\end{tabular}
\end{center}

\caption{20 entities in \dataset entity database -- Part 2}
\label{table:entity-examples-2}
\end{table*}

\begin{table*}[htbp]
\begin{center}
\begin{tabular}{p{5cm}p{10cm}}
\hline
\rowcolor{gray!10}
Entity Name & Entity Description \\
\hline
Farmers & Farmers are individuals or communities engaged in agriculture, responsible for cultivating crops and raising livestock. They operate globally and can be involved in various socio-political and economic events such as land disputes, protests, and negotiations affecting their livelihoods. Farmers often face challenges like resource competition, violence from armed groups, and policy changes impacting their work conditions and income. Their role is crucial in food production and sustainability, and they frequently interact with governments, organizations, and other agricultural stakeholders to address issues like land rights, security, and agricultural policies.\\
\rowcolor{gray!10}
Labor Group & A ``Labor Group" is a collective of workers united to advocate for their rights and interests in various sectors of the economy. These groups often engage in activities such as protests, strikes, and negotiations to address issues related to working conditions, wages, and employment security. They may also become involved in larger civil unrest, participating in events like riots or demonstrations to exert pressure on employers or authorities. While not typically associated with violent activities, labor groups can sometimes be connected to broader social movements that may encounter conflicts with law enforcement or political entities. Labor groups operate globally, often organized at local, national, or industry levels, and play a crucial role in labor relations and policy advocacy. \\
Guatemalan Group & The ``Guatemalan Group" refers to a collective of Guatemalan immigrants and workers who engage in activism, particularly around labor rights, in countries like the United States and Australia. This group is involved in protests and rallies advocating for fair labor conditions and justice for immigrant workers. Their activism is exemplified through participation in events such as May Day rallies, where representatives like Eder Juarez highlight issues such as wage theft and lack of employee rights, making them a voice for immigrant labor struggles.\\
\rowcolor{gray!10}

JI: Jamiat-e-Islami & Jamiat-e-Islami (JI) is a significant political and military organization in Afghanistan, primarily composed of ethnic Tajiks. Established in the 1970s, it played a crucial role in the resistance against the Soviet invasion and later in the Afghan civil war. Historically aligned with prominent leaders such as Ahmad Shah Masoud, JI has maintained influence in Afghan politics, often representing non-Pashtun interests. Despite the Taliban's dominance, JI continues to be active, reflecting ongoing ethnic and political tensions within the country. Its members, including prominent diplomats and officials, have been involved in key governmental roles and resistance efforts against various regimes. \\

\hline
\end{tabular}
\end{center}

\caption{20 entities in \dataset entity database -- Part 3}
\label{table:entity-examples-3}
\end{table*}

\begin{table*}[htbp]
\begin{center}
\begin{tabular}{p{5cm}p{10cm}}
\hline
\rowcolor{gray!10}
Entity Name & Entity Description \\
\hline

Chang Tribal Group & The Chang Tribal Group is an indigenous community in India, primarily located in the state of Nagaland. Represented by the Chang Wedoshi Setshang (CWS), the group is known for advocating for their rights and addressing local grievances, particularly in the educational sector. They have been involved in protests to demand better resources and support from the government, as seen in their actions to secure transportation for Sao Chang College. The group's activities underscore their active role in seeking improved living and educational conditions for their community.\\
\rowcolor{gray!10}

SD: Solidarity Party & The SD: Solidarity Party, also known as Solidariedade, is a political organization based in Brazil that is categorized as a political militia. It is known for its involvement in violent actions against civilians, often linked to political motives. The party has been associated with political figures in vulnerable positions, such as José Erlânio Firmiano, a city councilor who was assassinated in Alagoas. The Solidarity Party remains active in Brazilian politics, highlighting ongoing challenges related to political violence in the region.\\

Los Motonetos Gang & The Los Motonetos Gang is a political militia group operating primarily in Mexico, known for using violence to further their political aims. The gang gained notoriety for its involvement in riots and for the use of high-caliber weapons, which has resulted in significant unrest and necessitated interventions by local and national security forces, including the Municipal Police, State Preventive Police, and the Mexican Army. The group's influence and operational capacity were highlighted following the assassination of their presumed leader, Juan Hernández López, also known as El Fayo, which led to armed protests and heightened security measures in the region of San Cristóbal de las Casas, Chiapas.\\
\rowcolor{gray!10}

Mebri Tribal Group & The Mebri Tribal Group is an indigenous community in Indonesia, specifically located in Papua. They are actively involved in advocating for the recognition and protection of their ancestral land rights. The group is known for organizing protests to demand fair compensation for the use of their land by government projects, such as healthcare infrastructure. They emphasize negotiation and dialogue with government authorities to resolve land disputes, as exemplified by their demands directed at the Indonesian Ministry of Health regarding land claims in areas under development. \\
\hline
\end{tabular}
\end{center}

\caption{20 entities in \dataset entity database -- Part 4}
\label{table:entity-examples-4}
\end{table*}

\begin{table*}[htbp]
\begin{center}
\begin{tabular}{lp{10cm}}
\hline
\rowcolor{gray!10}
Entity Name & Entity Description \\
\hline

Ara Communal Group & The Ara Communal Group is a communal entity based in the town of Ara, located in the western countryside of As-Suwayda, Syria. Formed in 2024, the group is involved in regional socio-political activism and has participated in significant anti-Hayat Tahrir al-Sham protests across Idlib and Aleppo. These protests have called for political changes including the resignation of the group's leader ``al-Jolani," the release of detainees, and the dismantling of the General Security Apparatus. The group has also been linked to incidents of remote violence, such as assassination attempts using explosive devices, amidst a backdrop of security instability and weak law enforcement in areas controlled by regime forces. The Ara Communal Group remains active and continues to influence political dynamics in the region.\\
\rowcolor{gray!10}

Back the Blue & ``Back the Blue'' is a slogan and movement within the United States that expresses support for law enforcement officers. It is often used by individuals and groups, including political supporters, during protests and public demonstrations to show solidarity with police forces. The phrase is commonly associated with conservative and pro-law enforcement sentiments, frequently appearing in contexts where participants oppose policies perceived as critical of the police or supportive of police reform. ``Back the Blue'' can also signify a broader political stance that emphasizes law and order.\\

Nalia Communal Group & The Nalia Communal Group is a factional community group based in Nalia village, located in the Lohagara Upazila of Narail, India. It is characterized by internal conflict, with power struggles between different factions, notably those led by Shaukat Khan and Ravi Khan. The group has been involved in violent clashes, often requiring police intervention to restore order. These conflicts are primarily driven by issues of dominance within the community, and the group remains active in its region. \\
\rowcolor{gray!10}
ZPR: For Justice and Order & ZPR: For Justice and Order, also known as Za Pravdu i Red, is a political militia operating in Bosnia and Herzegovina since 2020. It is involved in political activities and protests, aiming to address issues of governance and electoral integrity. The group is led by Nebojša Vukanović and has been active in organizing demonstrations against political corruption and foreign exploitation of natural resources. ZPR is also linked to political candidates in regional elections, such as Slaviša Pavlović, whose affiliation with the group highlights its engagement in local politics.\\
\rowcolor{gray!10}
\hline
\end{tabular}
\end{center}

\caption{20 entities in \dataset entity database -- Part 5}
\label{table:entity-examples-5}
\end{table*}

\section{Examples of \AEL System Outputs}
\label{appendix:ael-system-outputs}
\begin{figure*}[htbp]
    \centering
    \includegraphics[width=0.9\linewidth]{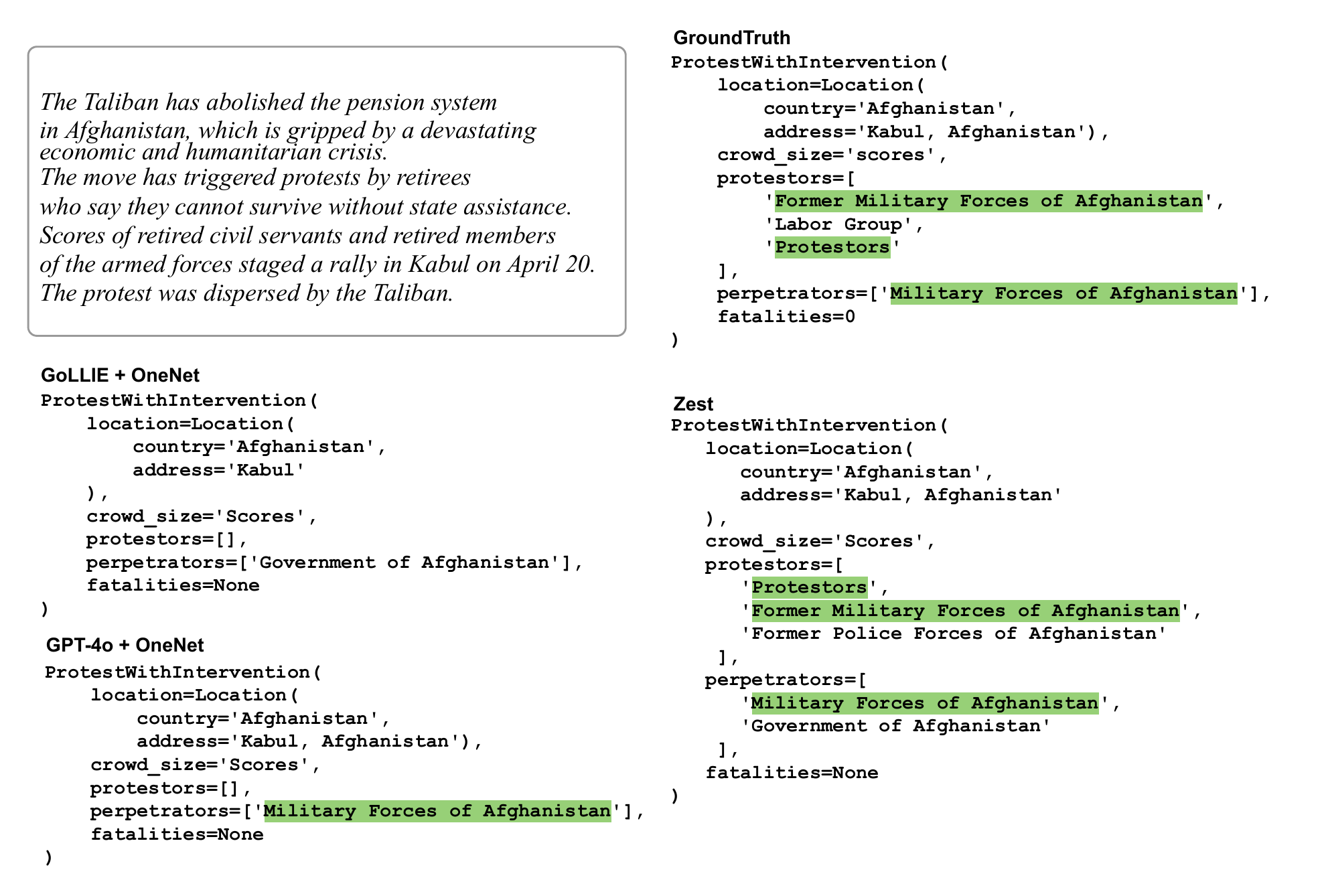}
    \caption{Example 1. The term ``retired'' in the input document indicates the involvement of the ``Former Military Forces of Afghanistan'' in the event. However, OneNet-based systems fail to capture this detail.}
    
    \label{figure:case1}
\end{figure*}

\begin{figure*}[htbp]
    \centering
    \includegraphics[width=0.9\linewidth]{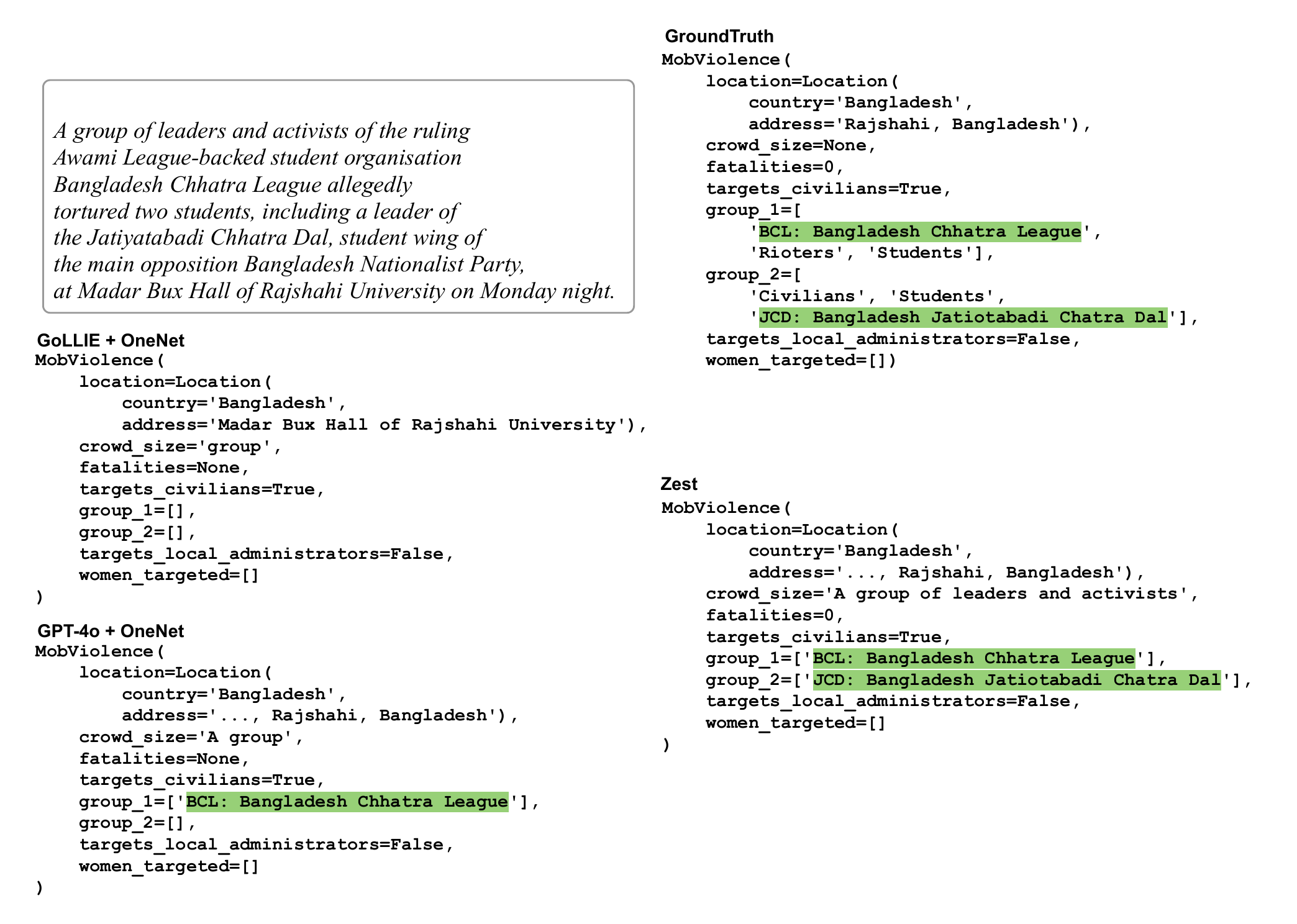}
    \caption{Example 2.}
    
    \label{figure:case2}
\end{figure*}

\begin{figure*}[htbp]
    \centering
    \includegraphics[width=0.9\linewidth]{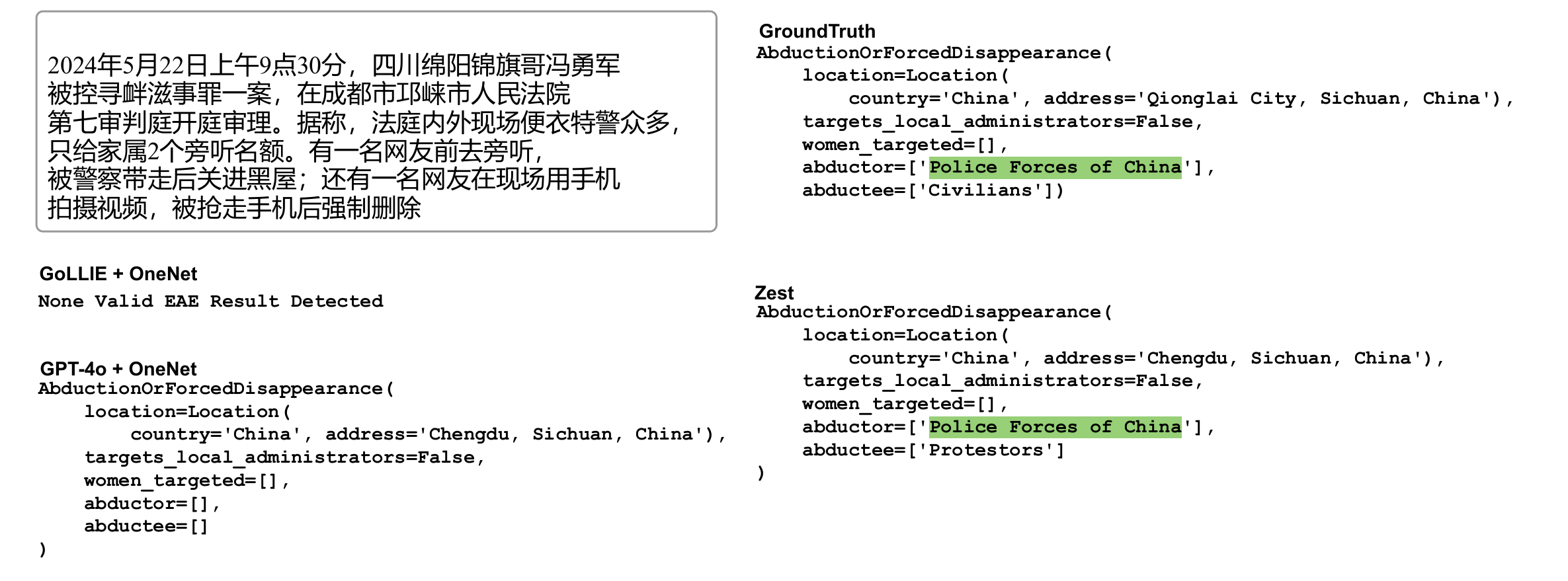}
    \caption{Example 3. The input text is in Chinese and translates as follows: ``On the morning of May 22, 2024, at 9:30 AM, the case of Feng Yongjun, known as the `Banner Brother' from Mianyang, Sichuan, who was charged with the crime of provoking trouble, was heard in the seventh courtroom of the Qionglai City People's Court in Chengdu. It is reported that numerous plainclothes special police officers were present both inside and outside the courtroom, and only two seats were allocated for family members to attend the hearing. One netizen who went to observe the trial was taken away by the police and detained in a dark room. Another netizen who was recording a video on their phone at the scene had their phone confiscated and the video forcibly deleted.''}
    
    \label{figure:case3}
\end{figure*}

\begin{figure*}[htbp]
    \centering
    \includegraphics[width=0.9\linewidth]{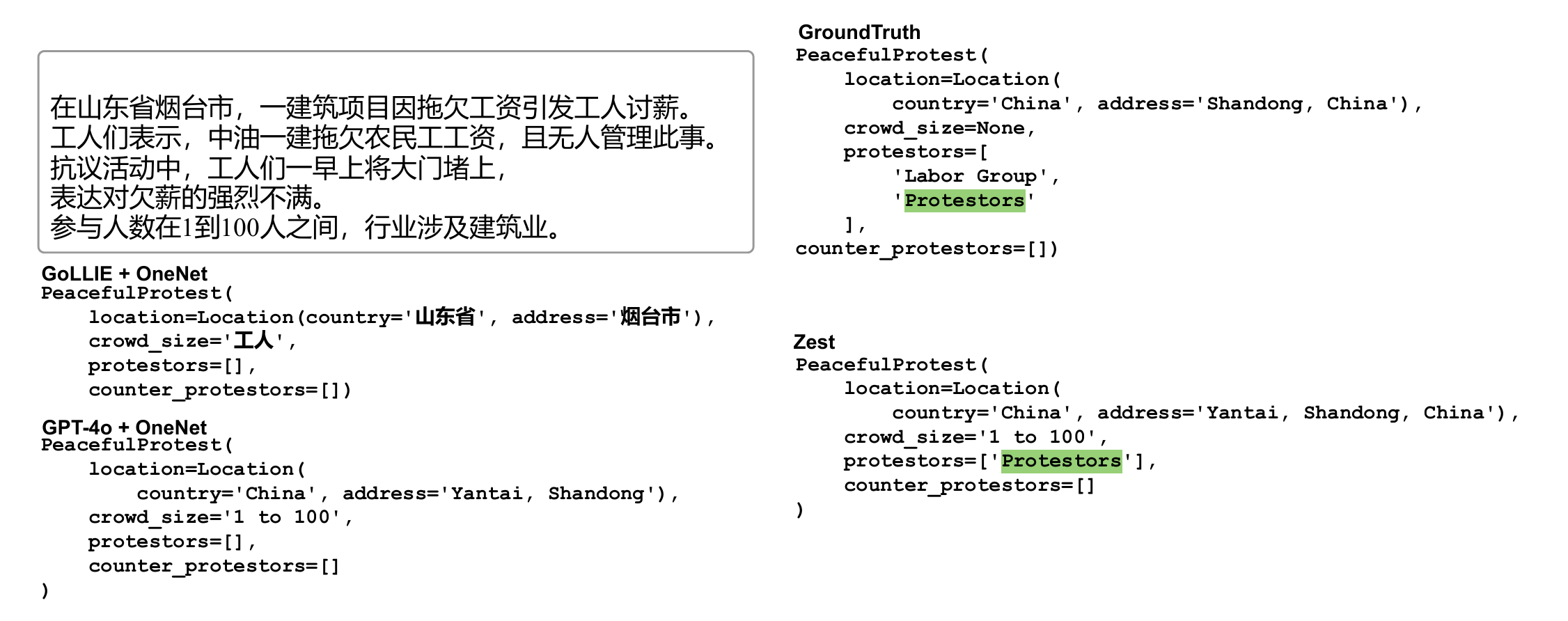}
    \caption{Example 4. The input text is in Chinese and translates as follows: ``In Yantai City, Shandong Province, a construction project has sparked a wage dispute due to unpaid wages. The workers stated that China Petroleum First Construction Corporation has been delaying the payment of wages to migrant workers, and no one is addressing the issue. During the protest, the workers blocked the entrance early in the morning to express their strong dissatisfaction with the unpaid wages. The number of participants ranged from 1 to 100 people, and the industry involved is construction.''}
    
    \label{figure:case4}
\end{figure*}

To illustrate the comparative performance of Span (GoLLIE-7B) + OneNet, Span (GPT-4o) + OneNet, and \system, we present four representative examples in English and Chinese. These examples are shown in Figures~\ref{figure:case1}, \ref{figure:case2}, \ref{figure:case3}, and \ref{figure:case4}.

\section{Prompts Used in Experiments}
\label{appendix:pipeline-prompts}
In this section, we provide the prompts used for various baselines, including \system. The prompts are written using the Jinja2 template language, which supports Python-like loops (\texttt{\{\% for \%\}\{\% endfor \%\}}), conditional statements (\texttt{\{\% if \%\}\{\% endif \%\}}), variables (\texttt{\{\{ var \}\}}), and comments (\texttt{\#}).

After substituting the variables into the prompt templates, the resulting strings under the sections labeled \texttt{\# instruction} and \texttt{\# input} are sent to the LLM as the system prompt and user message, respectively.

\begin{table*}[htbp]
\begin{lstlisting}[language=Jinja2, basicstyle=\ttfamily\small]
# instruction
You are tasked with determining the best matching Event types for a given news article. You will be provided with annotation guidelines and a news article to analyze. Your goal is to identify the most relevant event types and rank them in order of their match to the article content.

# input
Here is the news article you need to analyze:
{{ article }}

Now, carefully review the annotation guidelines for various event types:

{%
[{{ loop.index }}] "{{ ed[0] }}": {{ ed[1] }}

{%


1. For each event type, determine how well it matches the article content. Consider the following factors:
  - How closely the event description aligns with the main focus of the article
  - The presence of key actors or entities mentioned in the event type description
  - The occurrence of specific actions or outcomes associated with the event type

2. Rank the event types based on their relevance to the article content. Only include event types that have a meaningful connection to the article.

3. Output your results using the following format:
   - List the relevant event types in descending order of match quality
   - Use the ">" symbol to separate the event types

Your output should look like this:

[Explain your reasoning for the event types you decide to include, and their order]

event_type_1 > event_type_2 > ... 

Provide only the ranked list of event types in your final answer.
\end{lstlisting}
\caption{Prompt for event detection (ED).}
\label{prompt:ed}
\end{table*}

\begin{table*}[htbp]
\begin{lstlisting}[language=Jinja2, basicstyle=\ttfamily\small]
You will be given a news article about an event. Your task is to identify all potential Entities who are directly or indirectly involved in the event. Then, write a very short Wikipedia paragraph describing each entity in the general sense.

An Entity is defined as an individual, group, collective, or organization involved in an event. This includes:
  * Organized armed groups with political purposes (e.g. "Hezbollah", "ISIS")
  * Organizations, governments, and political parties (e.g. "BJP: Bharatiya Janata Party", "Government of India", "Democratic Party of U.S.")
  * Ethnic, religious, social or occupational groups (e.g. "Jewish Group", "Muslim Group", "Women", "Students", "Farmers", "Journalists", "Teachers", "Lawyers")
  * General terms describing people involved (e.g., "Rioters", "Protestors", "Civilians", "Labor Group")

When identifying Entities, follow these guidelines:

1. Be as thorough as possible. Think about what groups are implicitly or indirectly involved in the event. Ask yourself:
  - Can the identity group (religion, gender, occupation etc.) of the victims or perpetrators be inferred? If so, you should create an entity for that group.
  - Does the event involve workers or unions, or is it a labor issue? If so, you should add "Labor Group" as an entity.
  - Does the event in any way involve students, school or university? If so, you should add "Students" as an entity.
  - Does the event involve women in any way? If so, you should add "Women" as an entity.
  - Does the event involve civilians? If so, you should add "Civilians" as an entity.
  - Is the event a protest or a riot? If so, you should add "Protestors" or "Rioters" as an entity.
  - Does the event involve an unknown or unspecified group? If so, you should add one of "Unidentified Armed Group", "Unidentified Gang", "Unidentified Communal Group" etc. as an entity.
  - Given the country the event is taking place in, what are the major political parties, religious groups, armed groups, or social movements that could be involved? Consider cultural context of the region, like common religions, ethnicities etc.
  - And the like.

2. Include alternative names or spellings of each entity if mentioned in the article

3. For individuals, infer their role, affiliation, or social group as explained above.

4. For each entity you identify, think about its affiliated, parent or member groups. For example, if a politician is mentioned, think about their political party or any other group they are associated with. If a union is mentioned, think about the workers or labor groups it represents.

Use a scratchpad to think through your process:

<scratchpad>
[Your thought process here, including your answer to the above questions]
</scratchpad>

Then, present your output in the following JSON format. Output as many entities as you can possibly think of.
<entity_list>
{
    "entity 1": "Wikipedia paragraph 1",
    "entity 2": "Wikipedia paragraph 2",
    ...
}
</entity_list>

# input
Article: {{ article }}

\end{lstlisting}
\caption{Prompt for the first stage \system, to generate queries.}
\label{prompt:edl1}
\end{table*}

\begin{table*}[htbp]
\begin{lstlisting}[language=Jinja2, basicstyle=\ttfamily\small]
You will be given a news article about an event and potential Entities who are directly or indirectly involved in the event. Your task is to find supporting evidence for each of the specified entities in the given article.

An Entity is defined as an individual, group, collective, or organization involved in an event. This includes:
  * Organized armed groups with political purposes (e.g. "Hezbollah", "ISIS")
  * Organizations, governments, and political parties (e.g. "BJP: Bharatiya Janata Party", "Government of India", "Democratic Party of U.S.")
  * Ethnic, religious, social or occupational groups (e.g. "Jewish Group", "Muslim Group", "Women", "Students", "Farmers", "Journalists", "Teachers", "Lawyers")
  * General terms describing people involved (e.g., "Rioters", "Protestors", "Civilians", "Labor Group")

Follow these steps carefully:

1. First, you will be provided with the full text of the news article. Read the article carefully to understand the context of the event.

2. Next, you will be given a list of entities involved with the event.

3. Identify all supporting evidence of each given entity. Each evidence should be a short span from the article that has one of the following:
   - Contains the entity name, abbreviation or variations of its name
   - Implies the entity indirectly. For example "Madrasa" could be an evidence for "Muslim Group".
   - Mentions an affiliated group or organization of the entity.

4. If there are multiple evidence for the involvement of an entity, output one of them. If no evidence is found for an entity, respond with a mostly empty `EntitySpan` and only fill the `explanation` field.

5. For each evidence you find for an entity, provide your answer in the provided JSON format. Include the original entity name in the `entity_name` field to denote which entities the evidence is for.

6. If unsure, err on the side of including the span as evidence.


# input
<article>
Country of event: {{ country }}
{{ article }}
</article>

<entities>
{%
  - {{ e.name }}
  {{ e.description }}

{%
</entities>
\end{lstlisting}
\caption{Prompt for the second stage of \system.}
\label{prompt:edl2}
\end{table*}

\begin{table*}[htbp]
\begin{lstlisting}[language=Jinja2, basicstyle=\ttfamily\small]

You will be given a news article, and structured information about a {{ event_type }} event.
    A {{ event_type }} {{ event_type_definition }}.
Given a list of Entities that are involved in the event, your task is to assign each entity to the correct field.

An Entity is defined as an individual, group, collective, or organization involved in an event. This includes:
* Organized armed groups with political purposes (e.g. "Hezbollah", "ISIS")
* Organizations, governments, and political parties (e.g. "BJP: Bharatiya Janata Party", "Government of India", "Democratic Party of U.S.")
* Ethnic, religious, social or occupational groups (e.g. "Jewish Group", "Muslim Group", "Women", "Students", "Farmers", "Journalists", "Teachers", "Lawyers")
* General terms describing people involved (e.g., "Rioters", "Protestors", "Civilians", "Labor Group")

Possible fields are:
{{ possible_fields }}

To complete this task, follow these steps:

1. Analyze the news article and the {{ event_type }} event carefully.

2. For each entity in the provided list, determine their appropriate field based on the information in the news article.

3. Assign each entity to the most appropriate field. Try to assign all entities, event if their involvement in the event is very indirect. For example, "Government of India" is still an actor if the Indian congress is involved in the event.

4. If a field doesn't have a corresponding entity, leave it as an empty list.

Output the assignment of entities to fields in the following JSON format. Note that you should always include the full name of the entities without change.

{
    "field_name 1" : ["entity 1", "entity 2", ...],
    "field_name 2" : ["entity 3", "entity 4", ...],
}

# input
<news_article>
    {{ article }}
</news_article>

<event>
    {{ event_with_empty_entities }}
</event>

Here is the list of entities and their definitions.
<entities>
    {%
        - {{ e.name }}: {{ e.description }}

    {%
</entities>

\end{lstlisting}
\caption{Prompt used in the third stage of \system for assigning entities to their correct event argument. A Pydantic schema is also passed to the model to follow.}
\label{prompt:edl3}
\end{table*}

\begin{table*}
\begin{lstlisting}[language=Jinja2, basicstyle=\ttfamily\small]

You are an AI assistant tasked with extracting event arguments from a given news article. You will be provided with annotation guidelines for an event type and a news article to analyze.
Extract the arguments of the main event in the article, which is of type {{ event_type }}.
    
{{ event_type }}: {{ event_type_definition }}

When extracting event arguments, only pay attention to the main event in the article. Do not include any background information or other previous events that may be mentioned in the article.
            
# input
{{ article }}

\end{lstlisting}
\caption{Prompt used for Abstract Code4Struct}
\label{prompt:eae}
\end{table*}

\section{Experiment Details}
\label{appendix:experiment-details}

\subsection{Hyperparameters}
\label{appendix:hyperparameters}
All models were fine-tuned for 3 epochs with a batch size of 64. The final model checkpoint was selected for evaluation. We used a learning rate of \(2 \times 10^{-5}\), a cosine learning rate scheduler, and the AdamW optimizer~\citep{loshchilov2017decoupled}.

Training was conducted on a machine equipped with four NVIDIA A100 GPUs (80GB each), using DeepSpeed~\citep{deepspeed2020} and the Transformers library~\citep{wolf2019huggingfaces}. The fine-tuning process took approximately 3 hours in total.

For GPT-4o, we accessed the model through the OpenAI API. The total API usage cost was approximately \$2,500.

For geolocation information, we utilized the publicly hosted OpenStreetMap service via Nominatim (\url{https://nominatim.openstreetmap.org/}).

\subsection{Model Versions}
We use the following models:
\begin{itemize}
    \item \textbf{XLM-RRM}: \url{https://huggingface.co/BAAI/bge-m3-retromae}
    \item \textbf{mGTE (for entity retrieval)}: \url{https://huggingface.co/Alibaba-NLP/gte-multilingual-base}
    \item \textbf{Aya Expanse 8B}: \url{https://huggingface.co/CohereForAI/aya-expanse-8b}
    \item \textbf{GPT-4o-mini}: \texttt{gpt-4o-mini-2024-07-18}
    \item \textbf{GPT-4o}: \texttt{gpt-4o-2024-11-20}
\end{itemize}

\subsection{Baselines}
\label{appendix:baselines}

\paragraph{Selection of EE Baselines}
Many existing EE models rely on pre-trained language models with custom architectural modifications specifically designed for extractive EE tasks. Notable examples include DEGREE~\cite{hsu-etal-2022-degree}, TANL~\cite{tanl-2021}, X-GEAR~\cite{huang-etal-2022-multilingual-generative}, and TagPrime~\cite{hsu-etal-2023-tagprime}. However, these models are not suitable for evaluation on \dataset. They typically use pre-trained models such as BART~\cite{lewis-etal-2020-bart}, mT5~\cite{xue-etal-2021-mt5}, mBART~\cite{liu-etal-2020-multilingual-denoising}, and BERT~\cite{devlin-etal-2019-bert}, which have been pre-trained on sequences limited to 512 tokens, while approximately 39\% of the inputs in \dataset exceed this length.
 Furthermore, these models have primarily been evaluated on sentence-level EE datasets.
 Additionally, these models rely on explicit event triggers and argument spans, which are not provided in the \AEE formulation or in \dataset.

\paragraph{GoLLIE}
GoLLIE achieved state-of-the-art zero-shot generalization results on several event extraction datasets, including document-level EE datasets~\cite{li-etal-2021-document}. Notably, GoLLIE has been trained on RAMS~\cite{ebner-etal-2020-multi} and ACE05 document-level event datasets, making it a strong candidate for evaluating the \ED and \AEAE subtasks in a zero-shot setting. We use the GoLLIE-7B model, which is based on CodeLLaMA~\cite{rozire2023code}.
For zero-shot experiments, we provide GoLLIE with event descriptions formatted similarly to its original instruction-tuning data. Specifically, we define each event type as a Python class, including the event type description in the docstring and typical trigger words in class comments.

For the Event Detection (ED) subtask, GoLLIE predicts both the event type and its trigger span. For Event Argument Extraction (EAE), GoLLIE predicts the event type, trigger, and associated arguments. We discard the trigger span predictions.

To more closely match GoLLIE's instruction-tuning data, and to keep the instructions similarly short, we implement a two-stage event detection approach using GoLLIE. Initially, we predict one of six general event categories: \texttt{Battle}, \texttt{Protest}, \texttt{Riot}, \texttt{ExplosionOrRemoteViolence}, \texttt{ViolenceAgainstCivilians}, and \texttt{StrategicDevelopment}. After predicting the general event category, we further use GoLLIE to predict the corresponding subtype. Subsequently, Event Argument Extraction (EAE) is performed based on the predicted subtype.

\paragraph{OneNet}
OneNet is originally based on the 7B-parameter Zephyr model~\cite{tunstall2023zephyr}, an instruction-tuned version of Mistral~\cite{jiang2023mistral}. In our preliminary experiments, OneNet performed poorly on \dataset. Therefore, we replaced Zephyr with a stronger LLM (GPT-4o). We refer to this improved version as OneNet (GPT-4o).

Since OneNet expects entity spans as input, we first perform EAE using GoLLIE to obtain entity argument spans. Following the original OneNet setting, we retrieve 64 candidate entities for each entity span and then use GPT-4o instead of Zephyr for improved performance.

Both OneNet and \system include an entity retrieval component. We use the \texttt{gte-multilingual-base} model from~\citet{zhang2024mgte} to generate dense embeddings for entities based on their names and descriptions.

Following~\citet{logeswaran2019zero}, we initially reduce the candidate entities by selecting the top 64 most relevant entities for each argument mention. We then use the LLM to evaluate each candidate entity individually, given the contextual information, resulting in a refined set of potential entities.

In the dual-perspective entity linking stage of OneNet, we leverage the LLM to perform entity linking from two complementary perspectives: contextual analysis and prior knowledge. For each perspective, the LLM selects the most appropriate entity from the previously filtered set. In the contextual linking approach, the LLM is provided with both the context and the argument mention, enabling context-aware predictions. Conversely, in the prior knowledge approach, the LLM receives only the argument mention, relying solely on its inherent knowledge.
The final merging stage involves using the LLM to select the final entity from the two candidates identified in the previous stage. 

After EAE, we adopt a similar framework to OneNet for linking arguments to entities in the database. OneNet introduces an innovative approach using a fixed LLM to perform entity linking through few-shot prompting. The original framework comprises three distinct stages: entity reduction, dual-perspective entity linking, and merging linked entities. We closely follow this three-stage method, with minor modifications. Specifically, during the entity reduction stage, we first generate concise summaries for each entity description.

\section{Supplementary Experiments}
\begin{figure}[htbp]
    \centering
    \includegraphics[width=\linewidth]{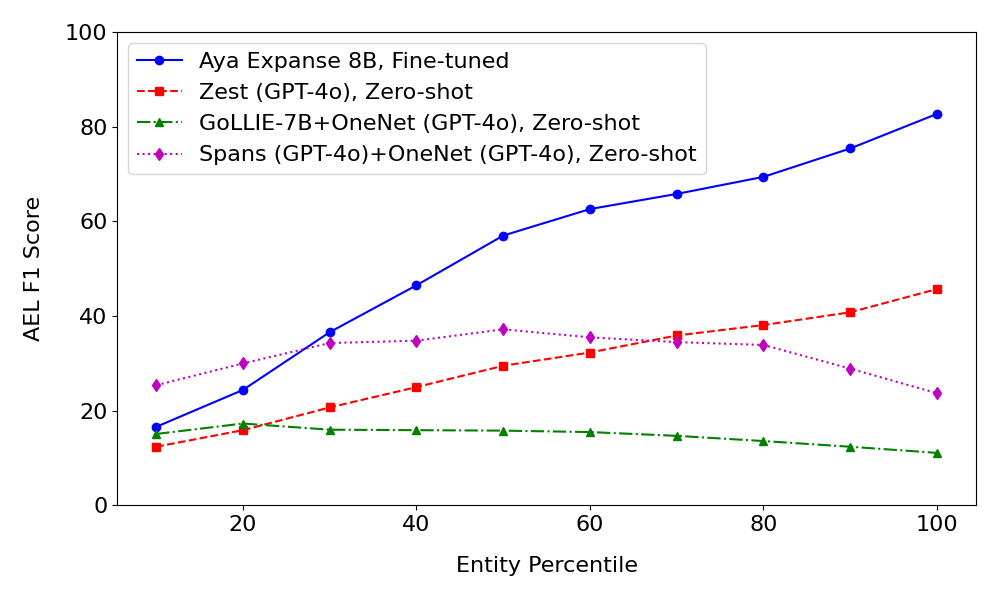}
    \caption{\AEL F1 as a function of how common entities are in the ACLED data.}
    \label{fig:lemonade-entity-percentile}
\end{figure}

Figure~\ref{fig:lemonade-entity-percentile} shows the performance of the best fine-tuned model (Aya Expanse) compared to \system and OneNet. \system and Aya Expanse, perform better on more common entities. OneNet (GPT-4o) models slightly outperform \system on very rare entities (less common than 20th and 60th percentiles), but \system outperforms them on more frequent entities, and on average.

\subsection{The Effect of Input Length}
\dataset has the longest average input text among document-level event datasets (Table~\ref{table:dataset-comparison}. Here, we analyze whether tackling this dataset requires using the entire documents.
We use the best supervised model, and run truncated documents from the test set through the model. The performance of this model in the end-to-end setting drops to 55.9, 60.9, 69.2, 72.6 for truncations to 32, 64, 128, 256 tokens respectively. This indicates that many examples in \dataset require the information from all parts of the document, not just the beginning.

\section{Full Schema of \dataset}
\label{appendix:dataset-schema}
The following is the full schema of \dataset, after conversion to Python code, in Pydantic~\citep{pydantic} format. Abstract classes (denoted by \texttt{ABC} are only meant to group event types together and store common event arguments, are not counted as an event type, and are not used by \system. Docstrings are adapted from the ACLED codebook~\citep{acled_codebook_2023}. \texttt{WomenTargetedCategory} and \texttt{Location} are two additional classes.

\pagebreak

\begin{lstlisting}[language=python]
class Battle(ACLEDEvent, ABC):
    """
    A "Battle" event is defined as a violent interaction between two organized armed groups at a particular time and location. "Battle" can occur between armed and organized state, non-state, and external groups, and in any combination therein. There is no fatality minimum necessary for inclusion. Civilians can be harmed in the course of larger "Battle" events if they are caught in the crossfire, for example, or affected by strikes on military targets, which is commonly referred to as "collateral damage" (for more, see Indirect Killing of Civilians). When civilians are harmed in a "Battle" event, they are not recorded as an "Actor", nor is a separate civilian-specific event recorded. If any civilian fatalities are reported as part of a battle, they are aggregated in the "Fatalities" field for the "Battle" event.
    The specific elements of the definition of a "Battle" event are as follows:
    Violent interaction: the exchange of armed force, or the use of armed force at close distance, between armed groups capable of inflicting harm upon the opposing side.
    Organized armed groups: collective actors assumed to be operating cohesively around an agenda, identity, or political purpose, using weapons to inflict harm. These groups frequently have a designated name and stated agenda.
    The "Battle" event type may include: ground clashes between different armed groups, ground clashes between armed groups supported by artillery fire or airstrikes, ambushes of on-duty soldiers or armed militants, exchanges of artillery fire, ground attacks against military or militant positions, air attacks where ground forces are able to effectively fire on the aircraft, and air-to-air combat.
    Cases where territory is regained or overtaken without resistance or armed interaction are not recorded as "Battle" events. Instead, they are recorded as "NonStateActorOvertakesTerritory" under the "StrategicDevelopment" event type
    "Battle" event type has the following subtypes:
    - GovernmentRegainsTerritory: Government forces or their affiliates regain control of a location from competing state forces or non-state groups through armed interaction.
    - NonStateActorOvertakesTerritory: A non-state actor or foreign state actor captures territory from an opposing government or non-state actor through armed interaction, establishing a monopoly of force within that territory.
    - ArmedClash: Armed, organized groups engage in a battle without significant changes in territorial control.

    """

    location: Location = Field(..., description="Location where the event takes place")
    fatalities: Optional[int] = Field(
        ...,
        description="Total number of fatalities, if known",
    )


class GovernmentRegainsTerritory(Battle):
    """
    Is a type of "Battle" event. This event type is used when government forces or their affiliates that are fighting against competing state forces or against a non-state group regain control of a location through armed interaction. This event type is only recorded for the re-establishment of government control and not for cases where competing non-state actors exchange control. Short-lived and/or small-scale territorial exchanges that do not last for more than one day are recorded as "ArmedClash".
    """

    government_force: List[str] = Field(
        ...,
        description="The government forces or their affiliates that regain control of the territory",
        is_entity_field=True,
    )
    adversary: List[str] = Field(
        ...,
        description="The competing state forces or non-state group that lose control of the territory. Can be State Forces, Rebel Groups, Political Militias, Identity Militias or External Forces",
        is_entity_field=True,
    )


class NonStateActorOvertakesTerritory(Battle):
    """
    Is a type of "Battle" event. This event type is used when a non-state actor (excluding those operating directly on behalf of the government) or a foreign state actor, through armed interaction, captures territory from an opposing government or non-state actor; as a result, they are regarded as having a monopoly of force within that territory. Short-lived and/or small-scale territorial exchanges that do not last for more than one day are recorded as "ArmedClash" events. In cases where non-state forces fight with opposing actors in a location many times before gaining control, only the final territorial acquisition is recorded as "Non-state actor overtakes territory". All other battles in that location are recorded as "ArmedClash".
    """

    non_state_actor: List[str] = Field(
        ...,
        description="The non-state actor overtaking territory. Can be Rebel Groups, Political Militias, Identity Militias or External Forces",
        is_entity_field=True,
    )
    adversary: List[str] = Field(
        ...,
        description="The opposing government or non-state actor from whom the territory was taken. Can be State Forces, Rebel Groups, Political Militias, Identity Militias or External Forces",
        is_entity_field=True,
    )


class ArmedClash(Battle):
    """
    Is a type of "Battle" event. This event type is used when two organized groups like State Forces, Rebel Groups, Political Militias, Identity Militias or External Forces engage in a battle, and no reports indicate a significant change in territorial control.
    `side_1` and `side_2` denote the two sides of the armed clash.
    Excludes demonstrations that turn violent, riots, and other forms of violence that are not organized armed clashes.
    """

    side_1: List[str] = Field(
        ...,
        description="Groups involved in the clash. Can be State Forces, Rebel Groups, Political Militias, Identity Militias or External Forces",
        is_entity_field=True,
    )
    side_2: List[str] = Field(
        ...,
        description="Groups involved in the clash. Can be State Forces, Rebel Groups, Political Militias, Identity Militias or External Forces",
        is_entity_field=True,
    )
    targets_local_administrators: bool = Field(
        ...,
        description="Whether this violent event is affecting current local government officials and administrators - including governors, mayors, councilors, and other civil servants.",
    )
    women_targeted: List[WomenTargetedCategory] = Field(
        ...,
        description="The category of violence against women, if any. If this violence is not targeting women, this should be an empty list.",
    )



class Protest(ACLEDEvent, ABC):
    """
    A "Protest" event is defined as an in-person public demonstration of three or more participants in which the participants do not engage in violence, though violence may be used against them. Events include individuals and groups who peacefully demonstrate against a political entity, government institution, policy, group, tradition, business, or other private institution. The following are not recorded as "Protest" events: symbolic public acts such as displays of flags or public prayers (unless they are accompanied by a demonstration); legislative protests, such as parliamentary walkouts or members of parliaments staying silent; strikes (unless they are accompanied by a demonstration); and individual acts such as self-harm actions like individual immolations or hunger strikes.
    Protestor are noted by generic actor name "Protestor". If they are representing a group, the name of that group is also recorded in the field.
    "Protest" event type has the following subtypes:
    - ExcessiveForceAgainstProtestors: Peaceful protestor are targeted with lethal violence or violence resulting in serious injuries by state or non-state actors.
    - ProtestWithIntervention: A peaceful protest is physically dispersed or suppressed without serious injuries, or protestor interact with armed groups or rioters without serious harm, or protestors are arrested.
    - PeacefulProtest: Demonstrators gather for a protest without engaging in violence or rioting and are not met with force or intervention.

    """

    location: Location = Field(..., description="Location where the event takes place")
    crowd_size: Optional[str] = Field(
        ...,
        description="Estimated size of the crowd. It can be an exact number, a range, or a qualitative description like 'small'.",
    )
    protestors: List[str] = Field(
        ...,
        description="List of protestor groups or individuals involved in the protest",
        is_entity_field=True,
    )


class ExcessiveForceAgainstProtestors(Protest):
    """
    Is a type of "Protest" event (Protest events include individuals and groups who peacefully demonstrate against a political entity, government institution, policy, group, tradition, business, or other private institution.) This event type is used when individuals are engaged in a peaceful protest and are targeted with lethal violence or violence resulting in serious injuries (e.g. requiring hospitalization). This includes situations where remote explosives, such as improvised explosive devices, are used to target protestors, as well as situations where non-state actors, such as rebel groups, target protestors.
    """

    perpetrators: List[str] = Field(
        ...,
        description="Entities perpetrating the violence. Can be State Forces, Rebel Groups, Political Militias, Identity Militias, External Forces",
        is_entity_field=True,
    )
    targets_civilians: bool = Field(
        ...,
        description="Indicates if the 'ExcessiveForceAgainstProtestors' event is mainly or only targeting civilians. E.g. state forces using lethal force to disperse peaceful protestors.",
    )

    fatalities: Optional[int] = Field(
        ...,
        description="Total number of fatalities, if known",
    )


class ProtestWithIntervention(Protest):
    """
    Is a type of "Protest" event. This event type is used when individuals are engaged in a peaceful protest during which there is a physically violent attempt to disperse or suppress the protest, which resulted in arrests, or minor injuries . If there is intervention, but not violent, the event is recorded as "PeacefulProtest" event type.
    """

    perpetrators: List[str] = Field(
        ...,
        description="Group(s) or entities attempting to disperse or suppress the protest",
        is_entity_field=True,
    )
    fatalities: Optional[int] = Field(
        ...,
        description="Total number of fatalities, if known",
    )


class PeacefulProtest(Protest):
    """
    Is a type of "Protest" event (Protest events include individuals and groups who peacefully demonstrate against a political entity, government institution, policy, group, tradition, business, or other private institution.) This event type is used when demonstrators gather for a protest and do not engage in violence or other forms of rioting activity, such as property destruction, and are not met with any sort of violent intervention.
    """

    counter_protestors: List[str] = Field(
        ...,
        description="Groups or entities engaged in counter protest, if any",
        is_entity_field=True,
    )




class Riot(ACLEDEvent, ABC):
    """
    "Riot" are violent events where demonstrators or mobs of three or more engage in violent or destructive acts, including but not limited to physical fights, rock throwing, property destruction, etc. They may engage individuals, property, businesses, other rioting groups, or armed actors. Rioters are noted by generic actor name "Rioters". If rioters are affiliated with a specific group - which may or may not be armed - or identity group, that group is recorded in the respective "Actor" field. Riots may begin as peaceful protests, or a mob may have the intention to engage in violence from the outset.
    "Riot" event type has the following subtypes:
    - ViolentDemonstration: Demonstrators engage in violence or destructive activities, such as physical clashes, vandalism, or road-blocking, regardless of who initiated the violence.
    - MobViolence: Rioters violently interact with other rioters, civilians, property, or armed groups outside of demonstration contexts, often involving disorderly crowds with the intention to cause harm or disruption.

    """

    location: Location = Field(..., description="Location where the event takes place")
    crowd_size: Optional[str] = Field(
        ...,
        description="Estimated size of the crowd. It can be an exact number, a range, or a qualitative description like 'small'.",
    )
    fatalities: Optional[int] = Field(
        ...,
        description="Total number of fatalities, if known",
    )
    targets_civilians: bool = Field(
        ...,
        description="Indicates if the 'Riot' event is mainly or only targeting civilians. E.g. a village mob assaulting another villager over a land dispute.",
    )
    group_1: List[str] = Field(
        ...,
        description="Group or individual involved in the violence",
        is_entity_field=True,
    )
    group_2: List[str] = Field(
        ...,
        description="The other group or individual involved in the violence, if any",
        is_entity_field=True,
    )
    targets_local_administrators: bool = Field(
        ...,
        description="Whether this violent event is affecting current local government officials and administrators - including governors, mayors, councilors, and other civil servants.",
    )
    women_targeted: List[WomenTargetedCategory] = Field(
        ...,
        description="The category of violence against women, if any. If this violence is not targeting women, this should be an empty list.",
    )


class ViolentDemonstration(Riot):
    """
    Is a type of "Riot" event. This event type is used when demonstrators engage in violence and/or destructive activity. Examples include physical clashes with other demonstrators or government forces; vandalism; and road-blocking using barricades, burning tires, or other material. The coding of an event as a "Violent demonstration" does not necessarily indicate that demonstrators initiated the violence and/or destructive actions.
    Excludes events where a weapon is drawn but not used, or when the situation is de-escalated before violence occurs.
    """


class MobViolence(Riot):
    """
    Is a type of "Riot" event. A mob is considered a crowd of people that is disorderly and has the intention to cause harm or disruption through violence or property destruction. Note that this type of violence can also include spontaneous vigilante mobs clashing with other armed groups or attacking civilians. While a "Mob violence" event often involves unarmed or crudely armed rioters, on rare occasions, it can involve violence by people associated with organized groups and/or using more sophisticated weapons, such as firearms.
    """




class ExplosionOrRemoteViolence(ACLEDEvent, ABC):
    """
    "ExplosionOrRemoteViolence" is defined as events as incidents in which one side uses weapon types that, by their nature, are at range and widely destructive. The weapons used in "ExplosionOrRemoteViolence" events are explosive devices, including but not limited to: bombs, grenades, improvised explosive devices (IEDs), artillery fire or shelling, missile attacks, air or drone strikes, and other widely destructive heavy weapons or chemical weapons. Suicide attacks using explosives also fall under this category. When an "ExplosionOrRemoteViolence" event is reported in the context of an ongoing battle, it is merged and recorded as a single "Battles" event. "ExplosionOrRemoteViolence" can be used against armed agents as well as civilians.
    "ExplosionOrRemoteViolence" event type has the following subtypes:
    - ChemicalWeapon: The use of chemical weapons in warfare without any other engagement.
    - AirOrDroneStrike: Air or drone strikes occurring without any other engagement, including attacks by helicopters.
    - SuicideBomb: A suicide bombing or suicide vehicle-borne improvised explosive device (SVBIED) attack without an armed clash.
    - ShellingOrArtilleryOrMissileAttack: The use of long-range artillery, missile systems, or other heavy weapons platforms without any other engagement.
    - RemoteExplosiveOrLandmineOrIED: Detonation of remotely- or victim-activated devices, including landmines and IEDs, without any other engagement.
    - Grenade: The use of a grenade or similar hand-thrown explosive without any other engagement.
    """

    location: Location = Field(..., description="Location where the event takes place")
    targets_civilians: bool = Field(
        ...,
        description="Indicates if the 'ExplosionOrRemoteViolence' event is mainly or only targeting civilians. E.g. a landmine killing a farmer.",
    )
    fatalities: Optional[int] = Field(
        ...,
        description="Total number of fatalities, if known",
    )
    attackers: List[str] = Field(
        ...,
        description="Entities conducting the violence",
        is_entity_field=True,
    )
    targeted_entities: List[str] = Field(
        ...,
        description="Entities or actors being targeted",
        is_entity_field=True,
    )
    targets_local_administrators: bool = Field(
        ...,
        description="Whether this violent event is affecting current local government officials and administrators - including governors, mayors, councilors, and other civil servants.",
    )
    women_targeted: List[WomenTargetedCategory] = Field(
        ...,
        description="The category of violence against women, if any. If this violence is not targeting women, this should be an empty list.",
    )


class ChemicalWeapon(ExplosionOrRemoteViolence):
    """
    Is a type of "ExplosionOrRemoteViolence" event. This event type captures the use of chemical weapons in warfare in the absence of any other engagement. ACLED considers chemical weapons as all substances listed as Schedule 1 of the Chemical Weapons Convention, including sarin gas, mustard gas, chlorine gas, and anthrax. Napalm and white phosphorus, as well as less-lethal crowd control substances - such as tear gas - are not considered chemical weapons within this event type.
    """


class AirOrDroneStrike(ExplosionOrRemoteViolence):
    """
    Is a type of "ExplosionOrRemoteViolence" event. This event type is used when air or drone strikes take place in the absence of any other engagement. Please note that any air-to-ground attacks fall under this event type, including attacks by helicopters that do not involve exchanges of fire with forces on the ground.
    """


class SuicideBomb(ExplosionOrRemoteViolence):
    """
    Is a type of "ExplosionOrRemoteViolence" event. This event type is used when a suicide bombing occurs in the absence of an armed clash, such as an exchange of small arms fire with other armed groups. It also includes suicide vehicle-borne improvised explosive device (SVBIED) attacks. Note that the suicide bomber is included in the total number of reported fatalities coded for such events.
    """


class ShellingOrArtilleryOrMissileAttack(ExplosionOrRemoteViolence):
    """
    Is a type of "ExplosionOrRemoteViolence" event. This event type captures the use of long-range artillery, missile systems, or other heavy weapons platforms in the absence of any other engagement. When two armed groups exchange long-range fire, it is recorded as an "ArmedClash". "ShellingOrArtilleryOrMissileAttack" events include attacks described as shelling, the use of artillery and cannons, mortars, guided missiles, rockets, grenade launchers, and other heavy weapons platforms. Crewed aircraft shot down by long-range systems fall under this event type.  Uncrewed armed drones that are shot down, however, are recorded as interceptions under "DisruptedWeaponsUse" because people are not targeted (see below). Similarly, an interception of a missile strike itself (such as by the Iron Dome in Israel) is also recorded as "DisruptedWeaponsUse".
    """


class RemoteExplosiveOrLandmineOrIED(ExplosionOrRemoteViolence):
    """
    Is a type of "ExplosionOrRemoteViolence" event. This event type is used when remotely- or victim-activated devices are detonated in the absence of any other engagement. Examples include landmines, IEDs - whether alone or attached to a vehicle, or any other sort of remotely detonated or triggered explosive. Unexploded ordnances (UXO) also fall under this category.
    SVBIEDs are recorded as "Suicide bomb" events, while the safe defusal of an explosive or its accidental detonation by the actor who planted it (with no other casualties reported) is recorded under "DisruptedWeaponsUse".
    """


class Grenade(ExplosionOrRemoteViolence):
    """
    Is a type of "ExplosionOrRemoteViolence" event. This event type captures the use of a grenade or any other similarly hand-thrown explosive, such as an IED that is thrown, in the absence of any other engagement. Events involving so-called "crude bombs" (such as Molotov cocktails, firecrackers, cherry bombs, petrol bombs, etc.) as well as "stun grenades" are not recorded in this category, but are included under either "Riot" or "StrategicDevelopment" depending on the context in which they occurred.
    """



class ViolenceAgainstCivilians(ACLEDEvent, ABC):
    """
    ACLED defines "ViolenceAgainstCivilians" as violent events where an organized armed group inflicts violence upon unarmed non-combatants. By definition, civilians are unarmed and cannot engage in political violence. Therefore, the violence is understood to be asymmetric as the perpetrator is assumed to be the only actor capable of using violence in the event. The perpetrators of such acts include state forces and their affiliates, rebels, militias, and external/other forces.
    In cases where the identity and actions of the targets are in question (e.g. the target may be employed as a police officer), ACLED determines that if a person is harmed or killed while unarmed and unable to either act defensively or counter-attack, this is an act of "ViolenceAgainstCivilians". This includes extrajudicial killings of detained combatants or unarmed prisoners of war.
    "ViolenceAgainstCivilians" also includes attempts at inflicting harm (e.g. beating, shooting, torture, rape, mutilation, etc.) or forcibly disappearing (e.g. kidnapping and disappearances) civilian actors. Note that the "ViolenceAgainstCivilians" event type exclusively captures violence targeting civilians that does not occur concurrently with other forms of violence - such as rioting - that are coded higher in the ACLED event type hierarchy. To get a full list of events in the ACLED dataset where civilians were the main or only target of violence, users can filter on the "Civilian targeting" field.
    "ViolenceAgainstCivilians" event type has the following subtypes:
    - SexualViolence: Any event where an individual is targeted with sexual violence, including but not limited to rape, public stripping, and sexual torture, with the gender identities of victims recorded when reported.
    - Attack: An event where civilians are targeted with violence by an organized armed actor outside the context of other forms of violence, including severe government overreach by law enforcement.
    - AbductionOrForcedDisappearance: An event involving the abduction or forced disappearance of civilians without reports of further violence, including arrests by non-state groups and extrajudicial detentions by state forces, but excluding standard judicial arrests by state forces.
    """

    location: Location = Field(..., description="Location where the event takes place")
    targets_local_administrators: bool = Field(
        ...,
        description="Whether this violent event is affecting current local government officials and administrators - including governors, mayors, councilors, and other civil servants.",
    )
    women_targeted: List[WomenTargetedCategory] = Field(
        ...,
        description="The category of violence against women, if any. If this violence is not targeting women, this should be an empty list.",
    )


class SexualViolence(ViolenceAgainstCivilians):
    """
    Is a type of "ViolenceAgainstCivilians" event. This event type is used when any individual is targeted with sexual violence. SexualViolence is defined largely as an action that inflicts harm of a sexual nature. This means that it is not limited to solely penetrative rape, but also includes actions like public stripping, sexual torture, etc. Given the gendered nature of sexual violence, the gender identities of the victims - i.e. "Women", "Men", and "LGBTQ+", or a combination thereof - are recorded in the "Associated Actor" field for these events when reported. Note that it is possible for sexual violence to occur within other event types such as "Battle" and "Riot".
    """

    fatalities: Optional[int] = Field(
        ...,
        description="Total number of fatalities, if known",
    )  # Is very very rare, only 7 events in English for 2024
    perpetrators: List[str] = Field(
        ...,
        description="The attacker(s) entity or actor",
        is_entity_field=True,
    )
    victims: List[str] = Field(
        ...,
        description="The entity or actor(s) that is the target or victim of the SexualViolence event",
        is_entity_field=True,
    )


class Attack(ViolenceAgainstCivilians):
    """
    Is a type of "ViolenceAgainstCivilians" event. This event type is used when civilians are targeted with violence by an organized armed actor outside the context of other forms of violence like ArmedClash, Protests, Riots, or ExplosionOrRemoteViolence. Violence by law enforcement that constitutes severe government overreach is also recorded as an "Attack" event.
    Attacks of a sexual nature are recorded as SexualViolence.
    If only property is attacked and not people, the event should be recorded as LootingOrPropertyDestruction event type.
    Excludes discovery of mass graves, which are recorded as "OtherStrategicDevelopment" events.
    """

    fatalities: Optional[int] = Field(
        ...,
        description="Total number of fatalities, if known",
    )
    attackers: List[str] = Field(
        ...,
        description="The attacker entity or actor(s)",
        is_entity_field=True,
    )
    targeted_entities: List[str] = Field(
        ...,
        description="The entity or actor(s) that is the target of the attack",
        is_entity_field=True,
    )


class AbductionOrForcedDisappearance(ViolenceAgainstCivilians):
    """
    Is a type of "ViolenceAgainstCivilians" event. This event type is used when an actor engages in the abduction or forced disappearance of civilians, without reports of further violence. If fatalities or serious injuries are reported during the abduction or forced disappearance, the event is recorded as an "Attack" event instead. If such violence is reported in later periods during captivity, this is recorded as an additional "Attack" event. Note that multiple people can be abducted in a single "Abduction/forced disappearance" event.
    Arrests by non-state groups and extrajudicial detentions by state forces are considered "Abduction/forced disappearance". Arrests conducted by state forces within the standard judicial process are, however, considered "Arrest".
    """

    abductor: List[str] = Field(
        ...,
        description="The abductor person or group(s)",
        is_entity_field=True,
    )
    abductee: List[str] = Field(
        ...,
        description="People or group(s) that were abducted or disappeared. Note that multiple people can be abducted in a single AbductionOrForcedDisappearance event",
        is_entity_field=True,
    )



class StrategicDevelopment(ACLEDEvent, ABC):
    """
    This event type captures contextually important information regarding incidents and activities of groups that are not recorded as "Political violence" or "Demonstration" events, yet may trigger future events or contribute to political dynamics within and across states. The inclusion of such events is limited, as their purpose is to capture pivotal events within the broader political landscape. They typically include a disparate range of events, such as recruitment drives, looting, and incursions, as well as the location and date of peace talks and the arrests of high-ranking officials or large groups. While it is rare for fatalities to be reported as a result of such events, they can occur in certain cases - e.g. the suspicious death of a high-ranking official, the accidental detonation of a bomb resulting in the bomber being killed, etc.
    Due to their context-specific nature, "StrategicDevelopment" are not collected and recorded in the same cross-comparable fashion as "Political violence" and "Demonstration" events. As such, the "StrategicDevelopment" event type is primarily a tool for understanding particular contexts.
    "StrategicDevelopment" event type has the following subtypes:
    - Agreement: Records any agreement between different actors, such as peace talks, ceasefires, or prisoner exchanges.
    - Arrest: Used when state forces or controlling actors detain a significant individual or conduct politically important mass arrests.
    - ChangeToArmedGroup: Records significant changes in the activity or structure of armed groups, including creation, recruitment, movement, or absorption of forces.
    - DisruptedWeaponsUse: Captures instances where an explosion or remote violence event is prevented, or when significant weapons caches are seized.
    - BaseEstablished: Used when an organized armed group establishes a permanent or semi-permanent base or headquarters.
    - LootingOrPropertyDestruction: Records incidents of looting or seizing goods/property outside the context of other forms of violence or destruction.
    - NonViolentTransferOfTerritory: Used when actors acquire control of a location without engaging in violent interaction with another group.
    - OtherStrategicDevelopment: Covers significant developments that don't fall into other Strategic Development event types, such as coups or population displacements.
    """

    location: Location = Field(..., description="Location where the event takes place")


class Agreement(StrategicDevelopment):
    """
    Is a type of "StrategicDevelopment" event. This event type is used to record any sort of agreement between different armed actors (such as governments and rebel groups). Examples include peace agreements/talks, ceasefires, evacuation deals, prisoner exchanges, negotiated territorial transfers, prisoner releases, surrenders, repatriations, etc.
    Excludes agreements between political parties, trade unions, or other non-armed actors like protestors.
    """

    group_1: List[str] = Field(
        ...,
        description="Group or individual involved in the agreement",
        is_entity_field=True,
    )
    group_2: List[str] = Field(
        ...,
        description="The other group or individual involved in the agreement",
        is_entity_field=True,
    )


class Arrest(StrategicDevelopment):
    """
    Is a type of "StrategicDevelopment" event. This event type is used when state forces or other actors exercising de facto control over a territory either detain a particularly significant individual or engage in politically significant mass arrests. This excludes arrests of individuals for common crimes, such as theft or assault, unless the individual is a high-ranking official or the arrest is politically significant.
    """

    detainers: List[str] = Field(
        ...,
        description="The person or group(s) who detains or jails the detainee(s)",
        is_entity_field=True,
    )
    detainees: List[str] = Field(
        ...,
        description="The person or group(s) being detained or jailed",
        is_entity_field=True,
    )


class ChangeToArmedGroup(StrategicDevelopment):
    """
    Is a type of "StrategicDevelopment" event. This event type is used to record significant changes in the activity or structure of armed groups. It can cover anything from the creation of a new rebel group or a paramilitary wing of the security forces, "voluntary" recruitment drives, movement of forces, or any other non-violent security measures enacted by armed actors. This event type can also be used if one armed group is absorbed into a different armed group or to track large-scale defections.
    """

    armed_group: List[str] = Field(
        ...,
        description="The name of armed group that underwent change",
        is_entity_field=True,
    )
    other_actors: List[str] = Field(
        ...,
        description="Other actors or groups involved. E.g. the government that ordered a change to its army.",
        is_entity_field=True,
    )


class DisruptedWeaponsUse(StrategicDevelopment):
    """
    Is a type of "StrategicDevelopment" event. This event type is used to capture all instances in which an event of "ExplosionOrRemoteViolence" is prevented from occurring, or when armed actors seize significant caches of weapons. It includes the safe defusal of an explosive, the accidental detonation of explosives by those allegedly responsible for planting it, the interception of explosives in the air, as well as the seizure of weapons or weapons platforms such as jets, helicopters, tanks, etc. Note that in cases where a group other than the one that planted an explosive is attempting to render an explosive harmless and it goes off, this is recorded under the "ExplosionOrRemoteViolence" event type, as the explosive has harmed an actor other than the one that planted it.
    """

    attackers: List[str] = Field(
        ...,
        description="The entity or actor(s) responsible for the remote violence",
        is_entity_field=True,
    )
    disruptors: List[str] = Field(
        ...,
        description="The entity or actor(s) disrupting the explosion or remote violence",
        is_entity_field=True,
    )
    targets_local_administrators: bool = Field(
        ...,
        description="Whether this violent event is affecting current local government officials and administrators - including governors, mayors, councilors, and other civil servants.",
    )
    women_targeted: List[WomenTargetedCategory] = Field(
        ...,
        description="The category of violence against women, if any. If this violence is not targeting women, this should be an empty list.",
    )


class BaseEstablished(StrategicDevelopment):
    """
    Is a type of "StrategicDevelopment" event. This event type is used when an organized armed group establishes a permanent or semi-permanent base or headquarters. There are few cases where opposition groups other than rebels can also establish a headquarters or base (e.g. AMISOM forces in Somalia).
    """

    group: List[str] = Field(
        ...,
        description="Entity or group(s) establishing the base",
        is_entity_field=True,
    )


class LootingOrPropertyDestruction(StrategicDevelopment):
    """
    Is a type of "StrategicDevelopment" event. This event type is used when actors engage in looting or seizing goods or property outside the context of other forms of violence or destruction, such as rioting or armed clashes. This excludes the seizure or destruction of weapons or weapons systems, which are captured under the "DisruptedWeaponsUse" event type. This can occur during raiding or after the capture of villages or other populated places by armed groups that occur without reported violence.
    """

    perpetrators: List[str] = Field(
        ...,
        description="The group or entity that does the looting or seizure",
        is_entity_field=True,
    )
    victims: List[str] = Field(
        ...,
        description="The group or entity that was the target of looting or seizure",
        is_entity_field=True,
    )
    targets_local_administrators: bool = Field(
        ...,
        description="Whether this violent event is affecting current local government officials and administrators - including governors, mayors, councilors, and other civil servants.",
    )
    women_targeted: List[WomenTargetedCategory] = Field(
        ...,
        description="The category of violence against women, if any. If this violence is not targeting women, this should be an empty list.",
    )


class NonViolentTransferOfTerritory(StrategicDevelopment):
    """
    Is a type of "StrategicDevelopment" event. This event type is used in situations in which rebels, governments, or their affiliates acquire control of a location without engaging in a violent interaction with another group. Rebels establishing control of a location without any resistance is an example of this event.
    """

    actors_taking_over: List[str] = Field(
        ...,
        description="The entity or actor(s) establishing control.",
        is_entity_field=True,
    )
    actors_giving_up: List[str] = Field(
        ...,
        description="The entity or actor(s) giving up territory, if known.",
        is_entity_field=True,
    )


class OtherStrategicDevelopment(StrategicDevelopment):
    """
    Is a type of "StrategicDevelopment" event. This event type is used to cover any significant development that does not fall into any of the other "StrategicDevelopment" event types. Includes the occurrence of a coup, the displacement of a civilian population as a result of fighting, and the discovery of mass graves.
    """

    group_1: List[str] = Field(
        ...,
        description="Group or individual involved in the StrategicDevelopment",
        is_entity_field=True,
    )
    group_2: List[str] = Field(
        ...,
        description="The other group or individual involved in the violence, if any",
        is_entity_field=True,
    )




class WomenTargetedCategory(str, Enum):
    CANDIDATES_FOR_OFFICE = "Women who are running in an election to hold a publicly elected government position"
    POLITICIANS = "Women who currently serve in an elected position in government"
    POLITICAL_PARTY_SUPPORTERS = "political party supporters"
    VOTERS = "Women who are registering to vote or are casting a ballot in an election"
    GOVERNMENT_OFFICIALS = "Women who work for the local, regional, or national government in a non-partisan capacity"
    ACTIVISTS_HRD_SOCIAL_LEADERS = (
        "Women who are activists/human rights defenders/social leaders"
    )
    RELATIVES_OF_TARGETED_GROUPS = "Women who are subject to violence as a result of who they are married to, the daughter of, related to, or are otherwise personally connected to (e.g. candidates, politicians, social leaders, armed actors, voters, party supporters, etc.)"
    ACCUSED_OF_WITCHCRAFT = "Women accused of witchcraft or sorcery, or other mystical or spiritual practices that are typically considered taboo or dangerous within some societies (excluding women who serve as religious leaders in religious structures that are typically not viewed as taboo or dangerous, such as nuns, female priests, or shamans)"
    GIRLS = "Girls who are under the age of 18; they may be specifically referred to by age or explicitly referred to as a child/girl"




class Location(BaseModel):
    """
    The most specific location for an event. Locations can be named populated places, geostrategic locations, natural locations, or neighborhoods of larger cities.
    In selected large cities with activity dispersed over many neighborhoods, locations are further specified to predefined subsections within a city. In such cases, City Name - District name (e.g. Mosul - Old City) is recorded in "specific_location". If information about the specific neighborhood/district is not known, the location is recorded at the city level (e.g. Mosul).
    """

    country: str = Field(
        ...,
        description="Name of the country in English. Example: United States",
    )
    address: str = Field(
        ...,
        description="Comma-separated address in order from neighborhood level to village/city, district, county, province, region, and country, if available. Excludes street names, buildings, and other specific landmarks. Example: Mosul, Old City, Nineveh, Nineveh, Iraq",
    )

\end{lstlisting}

\end{document}